\documentclass[11pt, a4paper]{article}
\usepackage{amsmath, amssymb, amsthm}
\usepackage{graphicx}
\graphicspath{{figures/}}
\usepackage{booktabs}
\usepackage{tabularx}
\usepackage[margin=1in]{geometry}
\usepackage{hyperref}
\hypersetup{bookmarksdepth=1}
\usepackage{bm}
\usepackage{empheq}
\usepackage{float}
\usepackage[section]{placeins}  % auto-FloatBarrier at each \section{}
\usepackage[font=small,labelfont=bf]{subcaption}

\makeatletter
\renewcommand\paragraph{\@startsection{paragraph}{4}{\z@}%
  {3.25ex \@plus1ex \@minus.2ex}%
  {-1em}%
  {\normalfont\normalsize\itshape}}
\makeatother

\title{\textbf{Learning as Observable Matrix Dynamics: \\
Diffusive Relaxations versus Phase Transitions}}

\author{Igor Halperin\thanks{All
calculations, numerical analysis, and manuscript
preparation were performed by Claude Code with Opus 4.7
working as an AI assistant under author's supervision. I
would like to thank Charles Martin for numerous discussions
and comments on the manuscript. I thank Eric Berger,
Andrey Itkin and Alejandro Rodriguez Dominguez for helpful
conversations on topics related to this research. All
remaining errors are my
own. All Python code, analysis scripts, figures, and animated
visualisations supporting this paper are available at
\url{https://github.com/ighalp/observable-matrix-dynamics-ml}.}}

\date{\today}

\begin{document}

\emergencystretch=2em

\maketitle

\begin{center}
Email: ighalp@gmail.com
\end{center}

\begin{abstract}
\emph{Observable Matrix Dynamics} (OMD) is a diagnostic
framework that probes the dynamics of high-dimensional
internal representations of inputs by a neural network via a
fixed-size $N \times N$ distance matrix $M(t)$ on a held set
of $N$ inputs. OMD uses methods of random matrix theory and
particle dynamics to explore spectral reorganisations that
are missed by scalar loss functions, but are informative of
the training process. We read $M(t)$ against a perturbative
ambient-versus-latent decomposition extending the
Bogomolny--Bohigas--Schmit (BBS) theory of random distance
matrices, with per-snapshot diagnostics for the
top-of-spectrum band structure and ambient noise,
trajectory-level observables linking snapshots, and a 3D MDS
embedding (bottom-three eigenvectors) rendering training as
a moving particle cloud. Across seven experiments, diffusive
regimes lack stable top-of-spectrum band structure, while
sharp endogenous or externally driven reorganisations
produce stable fingerprints: consistent with smooth or
product latent geometries in BBS-adjacent cases, and with
finite-cluster or Fourier-soliton structures otherwise. OMD
thus reads the geometric regime of a representation rather
than reporting a single intrinsic dimension.
\end{abstract}

\setcounter{tocdepth}{1}
\tableofcontents

%==============================================================================
\section{Introduction}
\label{sec:intro}
%==============================================================================

This paper develops a quantitative approach for online
monitoring of training or inference processes in neural
networks that we call \emph{Observable Matrix Dynamics}
(OMD). The main idea is to view these dynamics through
the lens of random matrix theory and physics-based
approaches by analysing the related dynamics of distance
matrices of $N$ i.i.d.\ task inputs across selected
layers of the network that carry internal
representations of its input. We use $N = 1000$
throughout. In modern networks the
parameter vector $\theta(t) \in \mathbb{R}^{P}$ has
$P$ ranging from millions to billions, which makes it
\emph{de facto unobservable} for any practical purpose:
no monitoring tool can follow such a process in full,
and the standard Langevin reading of training collapses
it externally into the single scalar loss curve. We
call our framework \emph{observable} matrix dynamics to
emphasise the contrast: we treat internal-layer
activations as observables computable from a forward
pass on a fixed evaluation sample at every training
step, and read training dynamics through them rather
than through the unobservable $\theta(t)$.
\par
Instead of following $\theta(t)$, OMD follows the dynamics
of a fixed-size $N \times N$ \textbf{distance matrix} of
$N$ test inputs in an internal representation space:
\[
M_{ij}(t) := \arccos\bigl(\hat h_i(t) \cdot \hat h_j(t)\bigr),
\qquad i, j = 1, \ldots, N,
\]
where $\hat h_i(t)$ stands for a unit-normalised
$D$-dimensional vector representation of the $i$-th input
at time $t$. The training trajectory is then mapped onto a
trajectory of the $M$-matrix whose
$N(N-1)/2$ coupled components evolve under the unobserved
action of $\theta(t)$. While the vector representation is used to compute the
distance matrix, all analyses and computations are
performed \textit{directly on the matrix, without
involving the original vector space}.
\par
The matrix observable opens two lines of analysis. The
first is random matrix theory: an arccos distance
matrix on $N$ points sampled from a $d$-dimensional
sub-manifold of the unit sphere is a Euclidean Random
Matrix \cite{mezard1999, goetschy2013}, and its
spectrum in the large-$N$ limit was analysed by Bogomolny,
Bohigas, and Schmit (BBS)
\cite{bogomolny2003, bogomolny2007}. The companion paper
\cite{halperin2026IBBS} extends BBS to an inference
setting, and proposes the Inference-BBS (I-BBS) method for
the ambient-versus-latent manifold setting relevant for
high-dimensional representations. Within I-BBS, the ambient matrix
decomposes as $M^{(D)} = M^{(d)} + \epsilon\,M^{(\mathrm{noise})} + O(\epsilon^2)$,
with $M^{(d)}$ being the latent BBS matrix on a sub-manifold
$\mathcal M_d \subset S^{D-1}$. Here $M^{(\mathrm{noise})}$
is a noise term describing deviations of the ambient
distance matrix $M^{(D)}$ from a low-dimensional distance
matrix $M^{(d)}$ computed on an embedded manifold in
$\mathbb{R}^D$, controlled by the noise strength parameter
$\epsilon$. The analysis
delivers a four-diagnostic per-snapshot toolkit,
as presented in Sections~\ref{sec:background}
and \ref{subsec:bbs_inference}. The Frustrated Distance
Matrix (FDM) framework of \cite{halperin2026FDM} adds
dynamic extensions and trajectory-level diagnostics on
top of the per-snapshot I-BBS toolkit. The present
paper applies a combined static
(I-BBS) plus dynamic (FDM) machinery to neural network
training.
\par
The second line is the physics of Langevin systems of
interacting particles. Our OMD framework transforms the
learning (or inference) dynamics of a neural network to
the dynamics of distance matrices on hyper-spheres that
can be equivalently interpreted in terms of non-equilibrium Langevin
dynamics of interacting particles. Furthermore, the
It\^o SDE for $M_{ij}(t)$ derived in
Section~\ref{subsec:ito_sde} expresses the observed
dynamics as a Langevin relaxation driven by a
loss-gradient drift plus noise-induced It\^o corrections. This
theoretical link of the observable $M$-matrix with the
unobserved parameter vector $\theta(t)$ enables inference
of the latter by observing the former, though we do not
pursue this direction here, and rather focus on other use
cases in this paper.

The OMD approach enables three distinct classes of use, all derived
from the same observable $M(t)$.
\emph{(i) Per-snapshot I-BBS plus trajectory-level FDM
diagnostics.} The empirical spectrum of $M(t)$ at every
training step is read against the I-BBS static-spectral
toolkit \cite{halperin2026IBBS} (multiplet multiplicity,
corrected delocalised and localised slopes, residual-RMT
test), stacked with the trajectory-level FDM observables
of \cite{halperin2026FDM} (level statistics, projector
drift, commutator norm, joint train--test gap). The
applicability of the multiplet and residual tests
depends on whether the representation cloud admits a
band structure (Sections~\ref{subsec:bbs_inference} and
\ref{sec:results}).
\emph{(ii) 3D visualisation as a physical multi-particle
process.} The bottom-three eigenvectors of $M(t)$
furnish an MDS embedding in $\mathbb{R}^{3}$ that
renders training as a moving particle cloud.
\emph{(iii) Quantitative description of phase transitions.}
Sharp structural reorganisations of the training
trajectory are captured by two complementary readings
(scalar spectral diagnostics localise the transition step,
the 3D MDS embedding shows the post-transition geometry),
with the same machinery for endogenous (grokking, sparse
parity) and externally driven (task switch,
input-topology) cases.

The band-structure regime is distinguished from diffusive relaxation:
the full four-diagnostic I-BBS toolkit fires only in
experiments where the spectrum of $M(t)$ admits a clean
band structure separating the leading eigenvalues from
the bulk. The experiments split into a
\emph{Group~A} (smooth or diffusive: MNIST + MLP at two
weight-decay levels, multi-output regression, and the
8-Gaussian GAN mode-coverage trajectory) where the
spectrum is either featureless or evolves smoothly with
no sharp jump, and a \emph{Group~B} (sharp phase
transitions: modular-arithmetic grokking and
sparse-parity learning as endogenous cases, synthetic
task switch and input-topology bifurcation as externally
driven quenches) where the post-event cloud condenses
onto a stable low-dimensional spectral fingerprint of
$M(t)$, which OMD detects directly. The interpretation
of that fingerprint is regime-dependent: BBS-adjacent in
the modular-arithmetic transformer (smooth/product
geometry), but finite-cluster or Fourier-soliton in the
others, outside the literal smooth-manifold BBS
asymptotic. OMD therefore acts as a \emph{regime
classifier} for representation geometry rather than as
a universal smooth-manifold validator.
Section~\ref{subsec:results-cross-experiment} of the
Results section unpacks the central manifold-formation
diagnostic (across-seed/through-time stability of
$\hat h_1$) and the regime classification (smooth /
product / Fourier-soliton / atomic-cluster / diffusive)
that the per-experiment readouts populate.

By contrast with this matrix-valued diagnostics, the conventional supervised-learning paradigm collapses
the joint behaviour of the entire dataset, viewed
through the model, into a single scalar per training step
(train loss and test loss), hiding the
\emph{collective geometry} of how $f_\theta$ acts on the
data. Two models with the same test accuracy can produce
very different output configurations on the same
evaluation set. Two stages of the same training run can
have indistinguishable loss values yet correspond to
qualitatively different structural states (feature
learning vs kernel, pre- vs post-grokking
\cite{power2022grokking, halperin2026grokking}, pre- vs
post-collapse in a generative model, pre- vs post-rank
collapse in a deep linear network). A scalar metric
cannot, by construction, see these distinctions. The joint train--test
object $(M^{\rm train}(t), M^{\rm test}(t))$ with the
spectral-synchronisation scalar
$\Delta_\beta(t) = \beta^{\rm train}(t) -
\beta^{\rm test}(t)$ (Section~\ref{subsec:joint}) is the
matrix-valued analogue of the standard
train-loss\,/\,test-loss pair, strictly finer.

We note that the spectrum of $M(t)$ is a
direct test of whether the network's internal
representation respects the symmetries of the supervised
task. The random choice of which $N$ inputs to follow
is similar to a quenched disorder in spin glasses \cite{mezard1987beyond}. The
$\binom{N_{\text{test}}}{N}$ possible $N$-subsets index
a replica space. The specific subset is fixed at the
start of analysis and held throughout training, and the
spectral observables self-average over the replica
space. The standard ML assumption that train and test
data are i.i.d.\ is a property of the data, not of the
representation: an arbitrary $h(\cdot; t)$ can map
i.i.d.\ data into representations whose pairwise
spectral statistics differ between train and test even
when the scalar losses agree. The joint diagnostics of
Section~\ref{subsec:joint}
($\Delta_\beta(t)$, off-diagonal block asymmetry,
MDS--Procrustes residual) test whether the train and
test representation geometries are exchangeable under
selected spectral summaries. Empirically, these
diagnostics separate from zero precisely when the
representation has not yet developed a structure that
fits the task's input or output symmetries, and contract
back toward zero when it does.

All experiments in this paper use neural networks (MLPs, a one-layer
transformer, a small CNN), but
the construction does not require neural architecture:
any parameterised $f_\theta$ producing an internal
$h(x; t) \in \mathbb{R}^D$ admits the same $M(t)$
(random forests via leaf-embedding, kernels via the RKHS
feature map, gradient-boosted trees via leaf-index
encodings), and the matched-space construction applies to
any regression-style learner. Extending to non-neural
architectures is left for future work.

The paper is organised as follows.
Section~\ref{sec:background} recaps the I-BBS, FDM, and
grokking-probe background \cite{halperin2026IBBS,
halperin2026FDM, halperin2026grokking}.
Section~\ref{sec:framework} develops the OMD framework
(ensemble construction, the It\^o SDE for $M_{ij}(t)$,
the I-BBS toolkit applied per snapshot, and the joint
train--test object \eqref{eq:M_joint}).
Section~\ref{sec:experiments} introduces the
experimental programme and the Group~A / Group~B split.
Section~\ref{sec:results} opens with the
cross-experiment observations
(Section~\ref{subsec:results-cross-experiment}: central
manifold-formation diagnostic, three-regime
classification, RSM ambient-noise finding) and
then reports the per-experiment analyses across the four
Group~A experiments (two diffusive runs plus the
8-Gaussian GAN smooth cluster-coverage trajectory) and
the four Group~B sharp-transition experiments. Section~\ref{sec:discussion}
synthesises the findings. Section~\ref{sec:summary}
summarises.
Appendix~\ref{app:ito-derivation} gives the It\^o-SDE
derivation,
Appendix~\ref{app:loss-symmetries} the equivariance
condition on $L(\theta)$, and
Appendix~\ref{app:robustness} the sensitivity and
calibration analyses.

\subsection{Related work}
\label{subsec:related_diagnostics}

Several existing tools extract geometric information from
learned representations on a fixed evaluation sample. OMD
relates to each as a continuous-time extension or a
matrix-valued generalisation.
\emph{RSA and CKA}
\cite{kriegeskorte2008rsa, kornblith2019cka} and
empirical-kernel-matrix methods (empirical NTK,
kernel-target alignment) compare two representation
matrices by a single scalar similarity score at fixed
checkpoints. OMD instead treats $M(t)$ as a continuous-time
stochastic dynamical object via the It\^o SDE
\eqref{eq:M_sde}, with the spectrum read along the full
training trajectory.
\emph{Neural-collapse diagnostics}
\cite{papyan2020neural,fang2021layerpeeled,zhu2021geometric,mixon2020neural,he2023law,rangamani2023intermediate}
scalar-contract the matrix observable to within- and
between-class moments. OMD generalises this to the full
spectrum of $M(t)$ and the I-BBS multiplet readout, with
the neural-collapse statistics recovered as particular
contractions of the top of the spectrum.
\emph{Intrinsic-dimension estimators}
(TwoNN \cite{facco2017twonn} and related local-density
approaches \cite{ansuini2019intrinsic, li2018measuring})
read $d_{\rm intrinsic}$ from nearest-neighbour statistics
of the representation cloud. The BBS-inferred dimension
$d_\beta(t)$ instead reads the rank-decay of the
arccos-distance-matrix spectrum against the random-matrix
prediction $d_\beta = \beta/(\beta - 1)$ for a smooth
manifold, with finite-$N$ correction from \cite{halperin2026IBBS}.
\emph{Manifold-learning visualisation}
(MDS, t-SNE, UMAP, PCA, Isomap) gives static embeddings at
fixed checkpoints. The 3D MDS-based visualisation of OMD
is the continuous-time, particle-identity-preserving
extension, with the same $N$ evaluation points tracked
across training steps.
\emph{Weight-spectrum analyses} (slingshot
\cite{thilak2022slingshot}, weight-space spectral analyses
adjacent to \cite{nanda2023progress}) track $\theta(t)$ or
derived spectral quantities. These are dual to OMD, which
looks at the observable representation side from
forward-pass observations alone.
\emph{Constructive symmetry approaches} (geometric deep
learning \cite{bronstein2021geometric}, equivariant
architectures \cite{cohen2016group, esteves2018learning},
self-supervised invariance-trained features) encode
symmetries into the architecture or data. OMD takes the
opposite stance, reading off whichever symmetries the
gradient flow has actually induced and, when a target
symmetry is known, promoting the same construction to an
active regulariser via a target matrix $T_{ij}$ (online
appendix).
The novel axes that OMD adds to this picture are the
continuous-time matrix dynamics identified by
\eqref{eq:M_sde}, the BBS-and-FDM spectral toolkit
imported from random-matrix theory, and the joint
train--test gap $\Delta_\beta(t)$.

\section{Background: I-BBS, FDM, and the grokking probe}
\label{sec:background}
%==============================================================================

For the per-snapshot analysis of the distance matrix we
inherit the I-BBS static-spectral toolkit \cite{halperin2026IBBS}. I-BBS
treats $N$ sample points on the ambient unit sphere
$S^{D - 1}$ close to a latent $d$-dimensional
sub-manifold $\mathcal M_d \subset S^{D-1}$, with the
ambient distance matrix decomposing perturbatively as
$M^{(D)} = M^{(d)} + \epsilon\,M^{(\mathrm{noise})} + O(\epsilon^2)$, where $M^{(d)}$
is the latent BBS matrix on $\mathcal M_d$ and
$\epsilon\,M^{(\mathrm{noise})}$ is the off-manifold correction
produced by one of two generative noise models, the
Residual Sphere Mixture (RSM) or the Free Spectral
Mixture (FSM). The I-BBS Algorithm~1 reads several
diagnostics from $M^{(D)}$ that together recover the
latent geometry and identify the ambient noise model.
We follow the dimension convention of
\cite{halperin2026IBBS}: the latent sphere is
$S^{d-1} \subset \mathbb R^d$, so $d$ is the embedding
dimension and the manifold has intrinsic dimension $d-1$.
The primary, gap-protected handle is the integer-valued
multiplet multiplicity $h(1, d) = d$ on the lowest
non-Perron BBS multiplet, which fixes the embedding
dimension $\hat d_{\rm mult} = \hat h_1$ (intrinsic
dimension $\hat h_1 - 1$). The multiplet positions give a second
handle through the parameter-free angular-momentum-level
shrinkage $f_\ell(\epsilon)$, whose inversion returns the
latent spectrum and confirms $\hat d$. The
delocalised-branch slope for large-$|\Lambda|$ eigenvalues
in the range $K \in [2, \sqrt N]$, in the large-$N$
limit, is $\beta_{\rm del} = d/(d-1)$, with the leading
finite-$N$ correction $\Delta\beta(N, d)$ measured
empirically on latent samples in \cite{halperin2026IBBS}.
This delocalised reading needs the window
$K \in [2, \sqrt N]$ to extend past the first multiplet,
which holds in the \emph{BBS regime} of
\cite{halperin2026IBBS}: a fixed low latent dimension $d$
and a large sample size $N \gg d^2$. The paper sits in
this regime throughout, with $N = 1000$ comfortably above
$d^2$ for the low latent dimensions the diagnostics read.
The ambient noise model is read
primarily from the $\ell = 2$ angular-momentum component,
populated by FSM and left empty by
the isotropic RSM, with the bulk eigenvalue density of
the residual $R = M^{(D)} - \hat M^{(d)}$ checked against
a Wigner semicircle as a secondary consistency test. The
localised-branch slope $\beta_{\rm loc} = 1/(d-1)$ at the
small-$|\Lambda|$ end is fragile under ambient noise (the
localised states sit inside the Wigner support of the
residual) and serves the noise diagnostic rather than the
dimension estimate. Throughout, an unsubscripted $\beta$
(including $\beta(t)$, $\beta^{\rm train/test}(t)$, and
$\Delta_\beta(t)$) denotes this delocalised exponent
$\beta_{\rm del} = d/(d-1)$. The subscript is kept only
for the localised $\beta_{\rm loc} = 1/(d-1)$.
The geometry of the latent
sub-manifold is read at the top of the descending-$|\Lambda|$
rank-ordered spectrum, and the embedding-model identity
from the $\ell = 2$ component and the bulk of the
residual spectrum. We refer the reader to
\cite{halperin2026IBBS} for all related derivations. This
paper uses the full I-BBS algorithm of \cite{halperin2026IBBS}
as a part of its main algorithm.

The ``Frustrated'' label in the Frustrated Distance Matrix (FDM) comes from two precursor
papers that motivated the matrix approach. The
Frustrated Brownian Particles (FBP) model
\cite{halperin2026frustrated} studies $N$ particles on
a fixed Riemannian manifold ($S^2$, cylinder, $T^2$)
evolving by overdamped Langevin dynamics with quenched
random pairwise couplings linear in the geodesic
distance. Despite the disorder, the particles
self-organise into low-dimensional configurations
(great-circle ring on $S^2$, discrete clusters on the
cylinder, parallel rings on $T^2$). The F2 paper
\cite{halperin2026fields} constructs the statistical
field theory of FBP in the large-$N$ limit and isolates
the slow orientational degree of freedom of the
self-organised configuration as a non-linear sigma model
on a projective target. The FDM paper
\cite{halperin2026FDM} reformulates these results on the
matrix side, extracting the pairwise-distance matrix
\begin{equation}
M(t)_{ij} \;=\; \arccos\bigl(\mathbf{x}_i(t) \cdot \mathbf{x}_j(t)\bigr)
\quad \text{on } S^2
\end{equation}
between the FBP particles and analysing its spectrum
through the original BBS theory of random distance
matrices \cite{bogomolny2003, bogomolny2007}, with the
per-snapshot I-BBS diagnostics of the previous paragraph
computed at each time. The FBP collapse from a uniform
configuration on $S^2$ to a great-circle ring ($S^1$) is
detected through three time-resolved I-BBS signatures:
an $\ell = 1$ multiplet rank reduction with bottom-five
eigenvalue fan-out (the multiplet multiplicity drops
from $h(1, 2) = 3$ to $h(1, 1) = 2$ as the configuration
becomes effectively one-dimensional), a bulk-scale
contraction with outlier-count drop, and the rank-decay
exponent shift from the sphere value $\beta = 3/2$ to
the ring value $\beta = 2$, equivalent under
\eqref{eq:bbs_def} to $d_\beta$ collapsing from $3$ to
$2$. On top of these per-snapshot reads, FDM introduces
trajectory-level observables that have no static
counterpart: the bottom-eigenspace projector drift
$D_K(t_{\rm ref}, \tau)$, the matrix-commutator norm
$C(t_{\rm ref}, \tau)$, and the level-spacing time
evolution against an i.i.d.\ resample null. The
combination of per-snapshot I-BBS reads and
trajectory-level FDM reads is the spectral machinery
this paper transfers to machine-learning models.

As a grokking probe, the grokking paper \cite{halperin2026grokking} applies
the combined I-BBS plus FDM toolkit, with the same
eigenvalue extraction and the same fit conventions, to a
different system: the residual-stream representations of
a small transformer trained on a modular-arithmetic task,
along the delayed generalisation trajectory known as
grokking \cite{power2022grokking}. At each of 63
logarithmically-spaced training steps the representations
$h_i(t) \in \mathbb{R}^{128}$ of an evaluation set of
$N = 1000$ test-split inputs are L2-normalised onto the
unit sphere $S^{127}$, and the arccos distance matrix
\eqref{eq:M_pair} is formed and diagonalised. The
per-snapshot I-BBS diagnostics fire across the grokking
transition: the rank-decay exponent $\beta(t)$ tracks a
sharp drop, with an implied dimension $d_\beta$ that
falls from a high value in the pre-grokking memorisation
phase to a low value in the post-grokking generalising
phase, and the bottom-multiplet structure reorganises
synchronously, identifying the post-transition latent
sub-manifold geometry. The grokking experiment is the
existence proof that this toolkit, designed for the FBP
particle test bed and analytically founded in
\cite{halperin2026IBBS}, applies verbatim to a deep
neural network. The present paper takes that observation
as a starting point and develops it into a general
framework for machine-learning models.

\section{OMD: Observable Matrix Dynamics of the neural training process}
\label{sec:framework}
%==============================================================================

\subsection{Distance matrix from model representations}
\label{subsec:distance_matrix}

Let $f_\theta : \mathcal{X} \to \mathcal{Y}$ be a
parameterised model with representation
$f_\theta(x) \in \mathbb{R}^D$ and $\theta(t)$ the
training trajectory. The \emph{representation}
$h(x; t) = f_{\theta(t)}(x)$ is the geometric object of
this paper, read out at a chosen layer of the network:
the output logits (decision-boundary geometry of a
classifier), the penultimate-layer features
(transfer-learning style), or an intermediate-layer
activation (residual stream at a chosen layer and token
position in the transformer case, in the spirit of the
grokking probe of \cite{halperin2026grokking}). This
choice is part of the framework: different layers probe
different aspects of the same trained model, and the
BBS-inferred dimension $d_\beta$ from \eqref{eq:bbs_def}
can differ across them, making it part of spectral model
selection.

For an $N$-sample $S = \{x_i\}_{i=1}^{N}$ of inputs, we
consider the \emph{unit-normalised representations}
\begin{equation}
\hat h_i(t) \;=\; \frac{h(x_i; t)}{\|h(x_i; t)\|_2}
\;\in\; S^{D-1},
\end{equation}
and the pairwise distance matrix is
\begin{equation}
M_S(t)_{ij} \;=\; \arccos\bigl(\hat h_i(t) \cdot \hat h_j(t)\bigr)
\;\in\; [0, \pi],
\label{eq:M_pair}
\end{equation}
the geodesic distance between $\hat h_i$ and $\hat h_j$ on
$S^{D-1}$ treated as the natural ambient sphere. This is
the same arccos kernel used throughout the FDM analysis of
\cite{halperin2026FDM} and the grokking probe of
\cite{halperin2026grokking}. The only change is the source
of the points $\hat h_i$. The matrix $M_S(t)$ is symmetric,
has vanishing diagonal, has entries in $[0, \pi]$, and is
fully determined by the current parameter $\theta(t)$ and
the choice of readout layer and sample $S$.

The entries of $M_S(t)$ are far from independent. With
$N$ points in the ambient $\mathbb R^D$ the configuration
has $N D$ scalar coordinates, while the symmetric
$M_S(t)$ has $N(N-1)/2$ off-diagonal entries, so whenever
$D < (N-1)/2$ the entries satisfy at least
$N(N-1)/2 - N D$ functional
constraints inherited from the shared embedding. The
I-BBS diagnostics extract information from these forced
correlations.

Our spectral-embedding convention is as follows. When we refer to a 3D ``MDS embedding'' of
$M_S(t)$, we mean the bottom-three eigenvectors of the
raw arccos distance matrix $M_S(t)$ (sorted by ascending
eigenvalue), Procrustes-aligned across checkpoints to a
final-frame reference, treated as $\mathbb R^3$
coordinates of the $N$ particles. This is the FDM
visualisation convention of \cite{halperin2026FDM} and
differs from classical MDS, which diagonalises the
double-centred squared-distance matrix
$-\tfrac{1}{2} J D^2 J$ with $D_{ij} = M_{ij}$ and $J$
the centring projector. We use raw-distance bottom
eigenvectors for direct comparability with the BBS
spectral references in \cite{halperin2026IBBS,
halperin2026FDM}, which treat the spectrum of $M$ itself
rather than of a derived squared-and-centred matrix.
The off-3D residual $r_\perp^{(i)}$ associated with each
particle is the projector-orthogonal Frobenius
contribution
$\bigl(\sum_{k \ge 4} |u_k^{(i)}|^2 \lambda_k^2\bigr)^{1/2}$
of the same diagonalisation, plotted as a radial
displacement.

\subsection{Two regimes: output level vs.\ representation level}
\label{subsec:two_regimes}

The construction of
Section~\ref{subsec:distance_matrix} lives at what we call
the \emph{representation level}: it forms a distance matrix
on the internal feature cloud $\{h(x_i; t)\}$, oblivious to
whether the model's output space and the target space
match. For a large class of ML tasks (regression,
autoencoders, denoisers, image-to-image translation,
diffusion-style learners) the output $f_\theta(x_i)$ and
the target $y_i$ do match, so the residual
\begin{equation}
r_i(t) \;=\; f_{\theta(t)}(x_i) - y_i
\end{equation}
is a well-defined vector and a second matrix-valued object
becomes available at the \emph{output level}.

The standard scalar loss is the trace of a rank-$N$
Gram-style matrix built from the residuals,
\begin{equation}
L_\theta(t) \;=\; \frac{1}{N}\sum_{i=1}^{N} \|r_i(t)\|_2^2
\;=\; \frac{1}{N}\,\mathrm{Tr}\bigl(R(t)\,R(t)^{\!\top}\bigr),
\qquad R(t)_{ij} = (r_i(t))_j,
\end{equation}
where $R(t)$ is the $N \times D_{\rm out}$ matrix of
residuals. The unweighted sum of the diagonal of
$R R^{\!\top}$ is the scalar loss. The full eigenvalue
spectrum of $R R^{\!\top}$, or equivalently the spectrum
of any pairwise distance matrix built from the rows of
$R$, contains strictly more information than its trace.

Two design choices, centring and metric, specify the output-level matrix.
First, centring: the raw construction probes the loss
directly and is sensitive to a global residual bias. The
centred $\tilde r_i = r_i - \bar r$ probes only the
relative geometry. Second, the metric: our default,
parallel to \eqref{eq:M_pair}, is the arccos geodesic on
L2-normalised (centred) residuals,
\begin{equation}
\hat{\tilde r}_i(t) \;=\; \frac{\tilde r_i(t)}{\|\tilde r_i(t)\|_2},
\qquad
M^{\rm loss}(t)_{ij} \;=\; \arccos\bigl(\hat{\tilde r}_i(t) \cdot \hat{\tilde r}_j(t)\bigr),
\end{equation}
which makes the BBS toolkit of
Section~\ref{subsec:bbs_inference} applicable verbatim.
A benchmark convention reports the chordal Euclidean
distance $\|\hat{\tilde r}_i - \hat{\tilde r}_j\|_2 =
\sqrt{2(1 - \cos\theta_{ij})}$: same rank ordering as
arccos, different spectrum. We use arccos to match
\cite{halperin2026FDM, halperin2026grokking} and report
the chordal-Euclidean spectrum as a sanity check.

As for where each level lives, the output-level matrix is available only when the output
and target spaces match. In this regime it is a strict
matrix-valued generalisation of the scalar loss, and the
spectral toolkit recovers it from $\sum_K \lambda_K$ as a
special case. For classification with a categorical
target, language modelling with token outputs, and any
setting in which one wants to inspect a deep internal
layer rather than the output, the residual is not a vector
and only the representation-level matrix \eqref{eq:M_pair}
is available. The two levels are not in competition: in
matched-space tasks both are available and probe different
aspects of the dynamics. The MNIST, grokking, task-switch,
GAN, and input-topology experiments of
Section~\ref{sec:results} use the representation level.
The regression experiment of
Section~\ref{sec:results-regression} exercises both levels
side by side.

\subsection{The ensemble at fixed training time}
\label{subsec:ensemble}

When the test set of size $N_{\rm test}$ is much larger
than $N$, the combinatorial number
$\mathcal{R} = \binom{N_{\rm test}}{N}$
of possible $N$-subsets of the test split indexes the
replica space of the matrix ensemble. Each $N$-subset
$S \subset \mathcal{D}_{\rm test}$ produces a distance
matrix $M_S(t)$ at each training step, and the set
\begin{equation}
\mathcal{E}_{\rm test}(t) \;=\; \bigl\{ M_S(t) : S \subset \mathcal{D}_{\rm test},\,|S| = N\bigr\}
\end{equation}
collects all replicas. The training-split analogue is
$\mathcal{E}_{\rm train}(t)$. With $N = 1000$ and
$N_{\rm test} \sim 10^4$, $\mathcal R$ is
combinatorially large. In practice we use either a
single $N$-subset (the quenched-disorder realisation) or
a small handful of disjoint $N$-subsets
($A = \lfloor N_{\rm test}/N\rfloor$ partitions) for
confidence bands. Sample-size sensitivity checks and
bootstrap-style bands on every diagnostic below follow
from sweeping $S$.

The source of randomness differs from the FDM
construction of \cite{halperin2026FDM}: there it lives in
quenched couplings $\phi_{ij}$ and initial particle
positions. Here it lives in the combinatorial choice of
$N$-subset $S$ picked once at the start of analysis and
held fixed throughout training. Either way, the spectral
toolkit sees a random distance matrix whose spectrum is
the object of interest, with the BBS template
\cite{halperin2026FDM} as the geometric reference. The
parameter trajectory $\theta(t)$ then carries the
ensemble $\mathcal E(t)$ as the matrix-valued analogue of
a loss curve, governed per replica by an It\^o's SDE that
we present next.

\subsection{It\^o's SDE for $M_{ij}(t)$}
\label{subsec:ito_sde}

The matrix $M(t)$ depends on time only through
$\theta(t)$, so the matrix-valued SDE for $M_{ij}(t)$
follows from It\^o's lemma applied to
$M_{ij}(\theta(t)) = \arccos(\hat h_i(\theta) \cdot
\hat h_j(\theta))$ along the Langevin parameter
trajectory of SGD/Adam,
\begin{equation}
d\theta_a \;=\; -\eta\,\bigl(\nabla_\theta L(\theta, t)\bigr)_a\, dt
\;+\; \sigma_{ab}(\theta, t)\, dW^b_t,
\qquad a = 1, \ldots, P,
\label{eq:weight_sde}
\end{equation}
with $\eta$ the effective learning rate and
$\Sigma_\theta = \sigma\,\sigma^{\!\top}$ the
parameter-space noise covariance (mini-batch gradient
fluctuations plus, for Adam, adaptive preconditioning).
The result is a matrix-valued SDE of drift-plus-diffusion
form,
\begin{equation}
dM_{ij}(t) \;=\; \mu_{ij}(\theta)\, dt
\;+\; \bm{\sigma}_{ij}(\theta)^{\!\top}\, dW_t,
\label{eq:M_sde}
\end{equation}
with drift $\mu_{ij}$ containing the loss-gradient
chain-rule term plus an $O(\Sigma_\theta)$ It\^o
correction, and diffusion vector $\bm\sigma_{ij}$ built
from the representation Jacobian $\nabla_\theta c_{ij}$
with $c_{ij} = \hat h_i \cdot \hat h_j$. The full
step-by-step derivation and the explicit closed-form
expressions for $\mu_{ij}$ and $\bm\sigma_{ij}$ are given
in Appendix~\ref{app:ito-derivation}. The discretised
linear-regression form on the $N(N-1)/2$ pairs and the
matrix gradient estimator
$\widehat{\nabla_\theta L}^{\rm mat}$ it yields are
developed in the accompanying online appendix and not
exercised here. The SDE \eqref{eq:M_sde} feeds the rest
of this paper through two readouts: (i) diffusion of
$M(t)$ between consecutive checkpoints, and (ii) the
two-sources-of-randomness picture separating the
\emph{thermal} (Langevin) noise of the weight update
from the \emph{quenched disorder} of the random
$N$-subset. The interplay is structurally identical to
the FBP model
\cite{halperin2026frustrated, halperin2026fields,
halperin2026FDM}, with the loss-gradient drift of
\eqref{eq:M_sde} playing the role of the FBP frustration
potential.

\subsection{Spectral observables: I-BBS ambient-versus-latent
decomposition and the diagnostic toolkit}
\label{subsec:observables}

The per-snapshot reading of $M(t)$ is governed by the
I-BBS ambient-versus-latent framework of
\cite{halperin2026IBBS}. The empirical points
$\hat h_i(t) \in S^{D-1}$ lie close to, but not on, a
latent $d$-dimensional sub-manifold
$\mathcal M_d \subset S^{D-1}$, and the ambient arccos
distance matrix decomposes as
\begin{equation}
M(t) \;=\; M^{(d)}(t) \;+\; \epsilon\,M^{(\mathrm{noise})}(t) \;+\; O(\epsilon^2),
\label{eq:ibbs_decomp}
\end{equation}
with $M^{(d)}$ the latent BBS matrix on $\mathcal M_d$
and $\epsilon M^{(\mathrm{noise})}$ the off-manifold correction from
an ambient embedding model of noise scale $\epsilon$.
I-BBS specifies the embedding model through two
generative classes, distinguished by where they place
the off-manifold component. The Residual Sphere Mixture
(RSM) is the convex combination of the latent and a
residual geometry at the cosine-kernel level,
$\cos M^{(D)} = (1 - \epsilon^2)\cos M^{(d)} +
\epsilon^2 \cos M^{(D-d)}$, where the residual cosine
kernel $\cos M^{(D-d)}_{ij}(\bar y) = \bar y_i \cdot
\bar y_j$ is the Gram matrix of uniform points
$\bar y_i$ on the residual sphere $S^{D-d-1}$. Because
the geodesic kernel $\arccos$ carries only odd
angular-momentum content, RSM shrinks each multiplet
but injects no even-$\ell$ structure. The Free Spectral
Mixture (FSM) is model-free: it builds the ambient cosine
kernel directly as a positive-definite zonal-kernel
mixture on the product of the latent and residual
sub-spheres,
\begin{equation}
\cos M^{(D)}_{ij} = \sum_{p, q \geq 0} \beta_{pq}\,
\mathcal Z_p^{(d)}(\tilde x_i \cdot \tilde x_j)\,
\mathcal Z_q^{(D-d)}(\bar y_i \cdot \bar y_j),
\qquad
\beta_{pq} \geq 0, \quad \sum_{p, q} \beta_{pq} = 1,
\label{eq:fsm-kernel}
\end{equation}
with $\tilde x_i \in S^{d-1}$ the latent and
$\bar y_i \in S^{D-d-1}$ the residual unit vectors,
normalised Gegenbauer zonal kernels $\mathcal Z_p^{(n)}$,
noise amplitude $\varepsilon^2 := 1 - \beta_{10}$, and the
even coefficient $\beta_{20} > 0$ populating the
$\ell = 2$ component that the isotropic RSM leaves empty by
parity \cite{halperin2026IBBS}. The synthetic null of
Appendix~\ref{app:noise-null} uses the concrete
realisation $\mathbb E[\cos M^{(D)}] = (1 - \epsilon)\cos
M^{(d)} + \epsilon\,P_2(\cos M^{(d)})$ with distance noise
$\sqrt 2\,\epsilon\sin^2\theta\,\xi$. Either way the
populated $\ell = 2$ channel is the FSM signature that
RSM lacks. The earlier additive-Gaussian and
heat-kernel embeddings are the isotropic limit of RSM
and coincide with it at $O(\epsilon^2)$
\cite{halperin2026IBBS}, so they are no longer separate
classes.

The I-BBS toolkit \cite{halperin2026IBBS} reads $M(t)$
through a small set of per-snapshot diagnostics that
jointly fix the latent dimension and identify the
embedding model. They are independent and may be
applied or weighted separately depending on the noise
regime and the spectral quality at the relevant rank
windows. The multiplet multiplicity is the primary,
gap-protected handle on the dimension, and the others
are cross-checks on it or read the noise model.

The lowest non-Perron BBS multiplet on $S^{d-1}$ has
multiplicity $h(1, d) = d$,
read from the descending-$|\Lambda|$ spectrum by a
log-gap walk at threshold $\tau$ (default
$0.25$). Gap-protection is a Davis--Kahan / Weyl
perturbation bound \cite{DavisKahan1970}, adapted to
the BBS multiplet in \cite{halperin2026IBBS}: the
multiplet survives ambient noise as long as the
inter-multiplet gap exceeds the perturbation norm. This
integer fixes $\hat d_{\rm mult} = \hat h_1$ and is
the main diagnostic. Closed-form $h(1, d)$ for $S^{d-1}$,
$T^d$, product spheres, and $\mathbb{RP}^{d-1}$ are in
\cite{halperin2026IBBS}, and the geometry-dependent
reading of $\hat h_1$ is treated in
Section~\ref{subsec:bbs_inference}.

Beyond multiplet multiplicities, further information is
provided by multiplet positions. Under ambient noise each degree-$\ell$
multiplet shrinks by a parameter-free factor
$f_\ell(\epsilon)$ set by the Funk--Hecke projection of
the noise-averaged kernel, fixed once $(\epsilon, d, D)$
are given with no quantity fit to the spectra. Inverting
the shrinkage, the de-shrunk positions $\Lambda_{{\rm
obs},\ell}/f_\ell$ are required to fall back onto the
clean BBS tower across the resolved odd degrees $\ell =
1, 3, \dots$, which over-determines a single $\epsilon$,
confirms $\hat d$, and returns the latent spectrum. The
lowest multiplets, shrunk least and gap-protected,
anchor the fit.

At large $|\Lambda|$ the delocalised rank-decay slope on
$K \in [2, \sqrt N]$ has the BBS value $\beta_{\rm
del} = d/(d-1)$, with an empirical finite-$N$ offset
$\Delta\beta(N, d)$ calibrated on latent samples in
\cite{halperin2026IBBS} ($\approx +0.5$ at $N = 1000$,
$d \in \{2, 3, 4\}$). The corrected estimator
$\hat\beta_{\rm del}^{\rm corr}(d_{\rm guess}) =
\hat\beta_{\rm del} - \Delta\beta(N, d_{\rm guess})$ is
matched against the target $d_{\rm guess}/(d_{\rm
guess}-1)$. The slope drifts under noise as it samples
progressively higher $\ell$, so it is a low-noise
cross-check rather than a primary handle.

The ambient noise model is read primarily from the
$\ell = 2$ angular-momentum component. FSM populates it
through its $P_2$ term while the isotropic RSM leaves it
empty up to a parity-suppressed $O(1/\sqrt N)$ floor, so
the blind ratio $|\hat\lambda_2/\hat\lambda_1|$ formed
from the top eigenvectors of $\cos M^{(D)}$ separates
the two classes about a geometric-mean boundary (RSM at
the floor, FSM populated). As a secondary consistency
check the residual $R = M^{(D)} - \hat M^{(d)}$ is
compared to a Wigner semicircle of $\epsilon$-predicted
width: once the recovered geometry is subtracted the
off-manifold deviation should behave as random-matrix
noise, with the latent multiplet sitting orders of
magnitude beyond the Wigner edge, which is the spectral
form of the gap protection. The reconstruction
distinguishes the oracle latent matrix $M^{(d)}$ from
its estimate $\hat M^{(d)} :=
\sum_{K=1}^{\lfloor\sqrt N\rfloor} \Lambda^{(D)}_K
u^{(D)}_K (u^{(D)}_K)^\top$, which keeps the full
resolved latent tower, the Perron mode together with
every multiplet sitting above the noise floor, and not
merely the Perron mode and the first multiplet. The cut
at $\lfloor\sqrt N\rfloor$ is the latent-versus-noise
crossover rather than an arbitrary discard of
small-$|\Lambda|$ eigenpairs. The clean latent weight
beyond it is a sub-percent fraction of $\|M^{(d)}\|_F$
(about $0.5$--$0.8\%$ at $N = 1000$), while extending the
sum past $\lfloor\sqrt N\rfloor$ injects the Wigner bulk
and raises $\|\hat M^{(d)} - M^{(d)}\|_F$ rather than
lowering it, so the estimate is stable across a band
around the crossover
(Appendix~\ref{app:reconstruction_depth}). The coarser
top-$(1 + \hat h_1)$ truncation, Perron mode plus first
multiplet only, differs by a few percent in Frobenius
norm and leaves the residual-RMT verdict unchanged. The
localised rank-decay slope at small $|\Lambda|$, with
BBS target $\beta_{\rm loc} = 1/(d-1)$, is
distribution-dependent and noise-dominated, so it serves
this noise diagnostic rather than the dimension
estimate. The blind separability of RSM and FSM is
verified on synthetic data in
Appendix~\ref{app:noise-null}. The multiplicity, the
shrinkage inversion, and the delocalised slope read the
latent dimension at the top of the spectrum. The
$\ell = 2$ component and the residual bulk read the
embedding-model identity.

The construction extends to product manifolds. The
ambient object is then a high-dimensional product sphere
$S^{D_a - 1} \times S^{D_b - 1}$, one factor per component
of the representation, which we approximate by a latent
product of two low-dimensional spheres
$\mathcal M = \mathcal M_a \times \mathcal M_b$ with
$\mathcal M_k \subset S^{D_k - 1}$. Each factor is treated
by the single-sphere construction
($\hat h^{(k),{\rm lat}}_i \in \mathcal M_k$ lifted by
$W_k$ with per-factor RSM/FSM noise $\epsilon_k$). The
natural observable is the squared-distance matrix
$N := M^2$, additive across factors by the Riemannian
product metric:
\begin{equation}
N^{(D_a, D_b)}_{ij}
\;=\;
\|h^{(a)}_i - h^{(a)}_j\|^2
+ \|h^{(b)}_i - h^{(b)}_j\|^2
\;=\;
N^{(D_a)}_{ij} + N^{(D_b)}_{ij},
\label{eq:N-additive}
\end{equation}
giving the product analogue of
\eqref{eq:ibbs_decomp}:
\begin{equation}
N^{(D_a, D_b)}
\;=\;
\underbrace{\hat N^{(d_a)}_a + \hat N^{(d_b)}_b}_{\text{latent on }\mathcal M_a \times \mathcal M_b}
\;+\;
\underbrace{\epsilon_a N^{(1, a)} + \epsilon_b N^{(1, b)}}_{\text{ambient noise}}.
\label{eq:N-decomp}
\end{equation}
For a Cartesian product sample of size $N_1 \times N_2$,
\cite{halperin2026IBBS} derives the explicit
$K^{(k)} \otimes J_{N_k}$ Kronecker decomposition that
factorises the spectrum exactly into two scaled BBS
blocks plus a large null space. Algorithm~1 applies per
factor to identify $(\hat d_a, \hat d_b)$, per-factor
multiplets, and per-factor noise models. The
factorisation commutator $\kappa(t) :=
\|[N_a, N_b]\|_F / (\|N_a\|\,\|N_b\|)$ quantifies the
deviation from strict Kronecker-sum structure. We use
this construction in the upstream reading of the
modular-arithmetic transformer with latent
$T^2 = S^1 \times S^1$
(Section~\ref{para:upstream-product-spheres}).

On top of the per-snapshot reads, the per-snapshot I-BBS reading is supplemented with the
trajectory-level diagnostics of the Frustrated Distance
Matrix framework \cite{halperin2026FDM}, which link
consecutive snapshots. The unfolded level-spacing
distribution $P(s)$ tests the universality class of the
bulk (GOE for ergodic spectra, Poisson for localised,
Berry--Robnik for the BBS reference). For two training
steps $t$ and $t'$, the bottom-$K$ projector distance and
the matrix commutator define
\begin{equation}
D_K(t, t') \;=\; \bigl\| P_K(t') - P_K(t)\bigr\|_F,
\qquad
C(t, t') \;=\; \bigl\| [M(t), M(t')]\bigr\|_F,
\end{equation}
with $P_K(t)$ the orthogonal projector onto the bottom-$K$
eigenspace of $M(t)$. Saturation of $D_K$ at the
eigenvalue-shuffled null value flags a slow coherent
rotation of the bottom eigenspace, the matrix imprint
of coherent dynamics in $\theta(t)$.

\subsection{Applying I-BBS Algorithm 1 to $M(t)$: dimension
inference and operative regimes}
\label{subsec:bbs_inference}

At each training step we apply the I-BBS algorithm of
\cite{halperin2026IBBS}, the per-snapshot core of the
boxed OMD diagnostic pipeline of this section, to the
empirical $M(t)$ on the fixed evaluation $N$-sample. The continuous
component of the algorithm gives the time-dependent BBS
dimension
\begin{equation}
\beta(t) \;=\; \frac{d(t)}{d(t) - 1},
\qquad
d_\beta(t) \;=\; \frac{\beta(t)}{\beta(t) - 1},
\label{eq:bbs_def}
\end{equation}
read off the corrected delocalised-branch slope on
$K \in [2, \sqrt N]$ with the finite-$N$ shift
$\Delta\beta(N, d)$ subtracted, the matrix analogue
of a dimension estimator built from the global
spectrum of $M(t)$ rather than from local
nearest-neighbour distances. Here $d_\beta$ is the
embedding dimension, so the latent manifold has intrinsic
dimension $d_\beta - 1$. When the spectrum admits a
clean band structure, the integer multiplet
$\hat h_1$ is the primary inference handle and
Eq.~\eqref{eq:bbs_def} the continuous cross-check. When
no clean band exists, Eq.~\eqref{eq:bbs_def} is the only
component of the toolkit that fires. The interpretation
of $\hat h_1$ depends on the latent geometry. For a
smooth sphere $S^{d-1} \subset \mathbb R^d$ the BBS
prediction is $h(1, d) = d$ (the dimension of the
$\ell = 1$ irrep of $\mathrm{SO}(d)$), giving the
dimension formula $\hat d_{\rm mult} = \hat h_1$ used by
the I-BBS algorithm \cite{halperin2026IBBS}.
For a smooth product manifold (e.g., $T^2 = S^1 \times S^1$)
each factor contributes its own multiplet and the joint
reading is obtained per factor via the
product-of-spheres construction of
Section~\ref{subsec:observables}. For a
discrete configuration on a smooth manifold (the
Fourier-soliton and atomic-cluster regimes of
Section~\ref{sec:results}) $\hat h_1$ counts the number
of active discrete modes or vertices in the leading band
($2k$ for $k$ active Fourier mode pairs on $S^1$;
$k - 1$ for $k$-vertex $\mathbb{Z}_k$-equivariant
clusters), and the literal $\hat h_1$ formula no
longer reads a smooth dimension. In every case
$\hat h_1$ is a gap-protected integer fingerprint of the
top-of-spectrum band structure. What it means
geometrically depends on which of the three regimes the
cloud sits in (Section~\ref{sec:discussion}).

On the scope and operative regimes, reading $d_\beta(t)$ as a literal latent dimension
rests on the smooth-manifold sampling assumption of the
BBS asymptotic and on $\beta(t) > 1$.
Section~\ref{sec:results} includes three regimes that
violate one or both. In a \emph{memorisation phase}
(pre-transition plateau of grokking and sparse parity),
$\beta(t) \leq 1$ signals ``no smooth-manifold
structure'' and the transition step is the re-entry into
$\beta > 1$. In a \emph{clustered-classification regime}
(post-transition task-switch, input-topology, late MNIST),
$d_\beta(t)$ reads cluster geometry rather than an
abstract task dimension. And in the
\emph{neural-collapse limit}, the BBS asymptotic
fails on a finite set of point masses and $d_\beta(t)$
reads the ETF geometry through its lowest multiplet. The
transition between regimes is itself observable in the
spectrum (multiplet structure, gap at $K = C$,
BBS-admissibility of $\beta$). The interpretation of
$d_\beta(t)$ is stated alongside each experimental result.

Beyond the dimension, the \emph{topology} of the latent
manifold leaves a signature in the multiplet structure:
$S^{d-1}$ admits all spherical-harmonic multiplets with
$(2\ell+1)$ degenerate eigenvalues at order $\ell$,
$\mathbb{RP}^{d-1} = S^{d-1}/\mathbb{Z}_2$ admits only the
even-$\ell$ multiplets (a sign-flip selection rule),
and the torus $T^d$ has a Fourier-series multiplet
structure. The choice of underlying topology therefore
enters the spectral toolkit as a \emph{discrete
hyperparameter}, read off the multiplet structure of
$M(t)$ rather than fixed by hand, on the same footing as
architecture depth or weight-decay coefficient. A
hyperparameter sweep over topology is a sweep over which
BBS template best fits the spectrum of $M(t)$ at
convergence.

\subsection{The joint train--test object}
\label{subsec:joint}

Beside the single-matrix construction \eqref{eq:M_pair},
we use a \emph{paired} construction with one distance
matrix $M^{\rm train}(t)$ on a train-split sample and one
$M^{\rm test}(t)$ on a test-split sample, both of size
$N$. The scalar
\emph{spectral-synchronisation} gap
\begin{equation}
\Delta_\beta(t) \;=\; \beta^{\rm train}(t) -
\beta^{\rm test}(t) \;=\;
\frac{1}{d^{\rm train}_\beta(t)} -
\frac{1}{d^{\rm test}_\beta(t)}
\label{eq:delta_beta}
\end{equation}
is a one-number indicator of whether the train and test
clouds inhabit manifolds of the same effective dimension.
A non-zero $\Delta_\beta(t)$ flags a model that has
organised the train cloud more aggressively than the test
cloud. A second construction takes a combined sample of
$N/2$ train and $N/2$ test inputs and forms the single
$N \times N$ matrix
\begin{equation}
M^{\rm joint}(t) \;=\;
\begin{pmatrix}
M^{\rm tr,tr}(t) & M^{\rm tr,te}(t) \\
M^{\rm te,tr}(t) & M^{\rm te,te}(t)
\end{pmatrix},
\label{eq:M_joint}
\end{equation}
whose off-diagonal block $M^{\rm tr,te}(t)$ has the same
statistical structure as the diagonal blocks for a
well-generalising model and is systematically larger
under overfitting (test points placed further from the
train manifold than train points are from each other).
The panel-(d) block-asymmetry quantity of the
per-experiment summary figures is the relative deviation
$\bigl(\langle M^{\rm tr,te}\rangle -
\tfrac{1}{2}(\langle M^{\rm tr,tr}\rangle +
\langle M^{\rm te,te}\rangle)\bigr) /
\langle M^{\rm tr,tr}\rangle$. A geometric
MDS-based synchronisation residual between the
bottom-three eigenspaces of $M^{\rm train}(t)$ and
$M^{\rm test}(t)$, recovered up to global isometry from
each matrix alone \cite{halperin2026FDM} and aligned by
Procrustes, gives a complementary one-number indicator of
train-test geometric agreement.

We collect the OMD pipeline in a single procedure for
reference. Steps marked \emph{(opt)} are optional and
exercised only in particular experiments.

\noindent\fbox{\parbox{0.97\linewidth}{
\small
\textbf{Algorithm: OMD diagnostic pipeline at training step $t$.}\\[2pt]
\textbf{Inputs:} model $f_\theta$ at parameters $\theta(t)$;
fixed evaluation samples $\{x_i\}_{i=1}^{N}$ from the test
split, picked once at the start of analysis and held fixed
throughout training; choice of representation layer
$h(x; \theta) \in \mathbb{R}^{D}$ (logits, penultimate, or
intermediate); candidate latent dimensions
$\mathcal D \subset \mathbb N$ (default $\{1, 2, 3\}$);
log-gap threshold $\tau$ (default $0.25$).\\[2pt]
1.~Form the matrix: forward-pass $h_i(t) = h(x_i; \theta(t))$,
normalise onto $S^{D-1}$, and build
$M_{ij}(t) = \arccos(\hat h_i(t) \cdot \hat h_j(t))$.\\
2.~Diagonalise: eigenvalues $\{\Lambda_K(t)\}_{K=1}^{N}$
sorted by descending $|\Lambda|$, with eigenvectors.\\
3.~\emph{Per-snapshot I-BBS Algorithm~1
\cite{halperin2026IBBS}} on the spectrum of step 2
(Section~\ref{subsec:observables}): the gap-walk multiplet
$\hat h_1$ fixes $\hat d_{\rm mult} = \hat h_1$
(primary), cross-checked by the angular-momentum shrinkage
inversion and the corrected delocalised slope
$\hat d_\beta$. The noise model follows from the
$\ell = 2$ component (FSM if populated, RSM at the parity
floor), with the residual $R(t) = M(t) - \hat M^{(d)}(t)$
as a Wigner-consistency check. When no gap exceeds $\tau$
the multiplet is reported as ``no band structure'' and
only $\hat d_\beta$ is used.\\
4.~\emph{Trajectory-level FDM diagnostics
\cite{halperin2026FDM}}: bottom-$K$ projector drift
$D_K(t, t_{\rm ref})$, matrix commutator
$C(t, t_{\rm ref})$, level-spacing statistics against
an i.i.d.\ resample null.\\
5.~\emph{(opt)} Joint train--test object: repeat steps
1--2 on a matched train sample, form $M^{\rm joint}(t)
\in \mathbb{R}^{2N \times 2N}$, extract
$\Delta_\beta(t) = \beta_{\rm train}(t) -
\beta_{\rm test}(t)$ and the off-diagonal block
asymmetry.\\
6.~\emph{(opt)} MDS embedding: bottom-three eigenvectors
of $M(t)$ as the visualisation plane, Procrustes-aligned
to the final-frame reference, with the off-3D residual
$r_\perp^{(i)}$ as a radial displacement.\\
7.~\emph{(opt)} Matrix training objective (online
appendix): add $\lambda \cdot \mathcal{L}_{\rm mat}(M(t),
M^\star)$ to the scalar loss with $M^\star$ a chosen
target (simplex-ETF, CE-teacher cosines, or
spectral-match).\\[2pt]
\textbf{Outputs:} the trajectories
$\{\hat d(t), \hat d_\beta(t), \hat\beta_{\rm del}^{\rm corr}(t),
\hat\beta_{\rm loc}(t), \text{noise verdict}(t),
D_K(t), C(t), \Delta_\beta(t), \dots\}$ and, when the
gap walk in step 3 succeeds, the inferred latent
sub-manifold $\mathcal M_{\hat d(t)} \subset S^{D-1}$
with top-eigenstructure reconstruction
$\hat M^{(d)}(t)$.
}}

\section{Experiments}
\label{sec:experiments}
%==============================================================================

Our experiments are structured around diffusive
relaxation versus phase transitions during ML training.
The It\^o SDE \eqref{eq:M_sde} of
Section~\ref{subsec:ito_sde} reads training as a
Langevin relaxation of $M(t)$, which
produces two qualitatively distinct phenomenologies in
the spectral observables. (i)~\emph{Diffusive relaxation}
to a single attracting saddle whose location is set by
the task's symmetries: $\beta(t)$, $d_\beta(t)$, the
$K = C$ eigenvalue gap, and $\Delta_\beta(t)$ trace
continuous trajectories with no sharp jumps.
(ii)~\emph{``Geometric'' phase transition}: the network
sits for a while on a metastable plateau, and then
reorganises discontinuously. The spectral observables show sharp
steps. The seven diagnostic experiments of
Section~\ref{sec:results} cover both
groups.\footnote{Experiments promoting $M(t)$ to an
active term in the training objective (MNIST + MLP,
CIFAR-10 + CNN) are at
\url{https://github.com/ighalp/observable-matrix-dynamics-ml}.}
We list them below in this conceptual order. Per-experiment
architecture, hyperparameters and numerics are inline in
the corresponding subsections of
Section~\ref{sec:results}.

The first group, diffusive Langevin relaxation (calibration
baselines and matched-space tasks), describes well-behaved relaxation to a
single attracting saddle, with no sharp jump in
spectral observables: \emph{(5.1) MNIST + MLP at
moderate weight decay} (calibration baseline,
non-monotonic feature-emergence transient);
\emph{(5.2) Multi-output regression} (the matched-space
construction, with $\beta_{\rm repr}$ and
$\beta_{\rm loss}$ tracing opposite-direction
trajectories); and \emph{(5.3) 8-Gaussian GAN} (the
generative analogue, with $d_\beta$ falling smoothly as
mode coverage grows).

The second group, phase transitions (sharp reorganisation in
the internal representation), reorganises discontinuously at a
sharply localised step in $M(t)$, in two flavours.
\emph{Endogenous} transitions hold data and target
i.i.d.\ so the transition emerges from weight-space
exploration alone: \emph{(5.4) Modular-arithmetic
grokking} (one-layer transformer at $p = 113$, with
``bagel formation'' onto
$T^2 = S^1(a) \times S^1(b)$ upstream and $S^1(a + b)$
downstream); and \emph{(5.5) Sparse-parity learning}
(MLP discovering $k = 3$ relevant bits in $d = 30$,
representation reorganising into a $\mathbb Z_2$-aligned
two-cluster $S^0$). \emph{Externally driven}
transitions break i.i.d.\ at a fixed switch step
\cite{halperin2026FDM, halperin2026grokking} on either
the output or input side: \emph{(5.6) Synthetic
supervisory-target task switch} (target flipped from
$\mathbb Z_2$ to $S_4$ output symmetry); and \emph{(5.7)
Input-distribution topology change} (input bifurcating
from a single isotropic Gaussian to a two-cluster
mixture).

All runs share a fixed evaluation sample of $N = 1000$, log- or linear-spaced
checkpoints, the arccos kernel, and a $\beta$ fit over
the canonical window $K \in [2, 50]$. Architecture- and data-specific details
are inline at the start of each results subsection.

To fix what counts as a symmetry, we list per experiment the exact discrete or continuous
symmetries that the task carries. The \emph{input}
column gives a group $G_{\rm in}$ acting on the input
data such that the supervised target is equivariant
(modular shifts on $(a, b) \in \mathbb{Z}_p \times
\mathbb{Z}_p$ for grokking, isotropic rotations of
latent factors for the synthetic experiments, octagonal
rotation of the 8-Gaussian benchmark, pixel-grid
$\mathbb{Z}^2$ for the convolutional CIFAR backbone).
For MNIST and CIFAR-10 we list ``approx.\ rot/scale''
because the natural-image symmetries (small rotations,
scales, affine deformations) are properties of the data
distribution rather than exact group actions on the
supervised loss (a $90^\circ$ rotation sends a `$6$' to
a `$9$' and breaks the task). The \emph{output} column
gives $G_{\rm out}$ under which the loss is invariant.
For $C$-class classification this is the class-relabelling
permutation group $S_C$, which is also the automorphism
group of the simplex-ETF arrangement that gradient flow
approaches under neural collapse
\cite{papyan2020neural}. Table~\ref{tab:axes}
summarises along these axes.

\section{Results}
\label{sec:results}
%==============================================================================

\subsection{Cross-experiment observations}
\label{subsec:results-cross-experiment}

\begin{table}[t!]
\centering
\footnotesize
\setlength{\tabcolsep}{5pt}
\renewcommand{\arraystretch}{1.15}
\begin{tabularx}{\linewidth}{@{}l X X l l@{}}
\toprule
Section & Task & Symmetries (input $\to$ output) &
$D$, $d_\beta$ & dynamics \\
\midrule
5.1 MNIST mod-WD & 10-class images &
approx.\ rot/scale (not exploited) $\to$ $S_{10}$ &
$256$, $3$--$4$ & diffusive \\
5.2 Regression & $\mathbb{R}^{20} \to \mathbb{R}^{8}$ &
$O(20)$ on irrelevant dirs $\to$ $\mathbb{R}^{8}$ translation &
$64$, $3.8$ & diffusive \\
5.3 GAN & 8-Gaussian gen &
$\mathbb{Z}_8$ octagonal $\to$ $\mathbb{Z}_8$ &
$32$, $3.4{\to}2.4$ & diffusive \\
5.4 Grokking & $a+b \bmod p$, $p=113$ &
$\mathbb{Z}_p \!\times\! \mathbb{Z}_p \to \mathbb{Z}_p \cong S^1$
& $128$, $2{\to}3$ & transition (endogenous) \\
5.5 Sparse parity & $k=3$ parity on $\{-1,+1\}^{30}$ &
$\mathbb{Z}_2^{30}$ $\to$ $\mathbb{Z}_2$ on label &
$128$, $\to$ $1$--$2$ & transition (endogenous) \\
5.6 Task switch & 1-D then 2-D target &
$O(2)$ on latent $z$ $\to$ $\mathbb{Z}_2$ then $S_4$ &
$32$, $3{\to}2$ & transition (driven) \\
5.7 Input topology & 1- vs 2-cluster regression &
$O(20) \to \mathbb{Z}_2 \!\times\! O(19)$;
output: $\mathbb{R}$ trans.\ & $32$, $4{\to}2.5$ & transition (driven) \\
\bottomrule
\end{tabularx}
\caption{The seven diagnostic experiments of
Section~\ref{sec:results} along three axes: the task's
input and output symmetry groups, the ambient
representation dimension $D$ with the observed BBS
effective dimension $d_\beta$, and the qualitative
character of the training trajectory (\emph{diffusive}
Langevin relaxation; \emph{transition} in the internal
representation, endogenous or externally driven). The
$d_\beta$ entries are end-of-training plateau values;
arrows indicate trajectory direction. $S_C$ denotes the
class-relabelling permutation group;
``approx.\ rot/scale'' denotes approximate continuous
image-plane symmetries of the natural-image data
distribution that the architectures of this paper exploit
only when stated.}
\label{tab:axes}
\end{table}

The seven diagnostic experiments listed in
Table~\ref{tab:axes} share a common pipeline: at
log-spaced training checkpoints we extract the
penultimate (or residual-stream) representation,
$L^2$-normalise onto $S^{D-1}$, form the arccos
distance matrix $M(t)$ on a fixed $N = 1000$
evaluation sample, and run the I-BBS
toolkit of Section~\ref{subsec:observables}. The
gap-walk threshold is set to $\tau = 0.25$ throughout
to step past intra-band gaps within the leading
multiplet (lower thresholds trigger on those gaps and
give seed-unstable $\hat h_1$ readings; see
Appendix~\ref{app:robustness-sweep}). The per-snapshot
$(\hat h_1, \hat d_\beta, \text{noise verdict})$
at the final checkpoint are consolidated in
Table~\ref{tab:ibbs_summary}. The per-experiment
subsections below add only the geometric observations
and figures specific to each case.

The central manifold-formation diagnostic is as follows. The multiplet multiplicity $\hat h_1(t)$ excluding the
Perron eigenvalue \textbf{is the integer fingerprint of
the top-of-spectrum band structure of $M(t)$}. Its
across-seed and through-time stability is the
operational test for manifold formation. Each
per-experiment combined figure shows in its panel~(e)
the per-seed $\hat h_1$ scatter, the cross-seed median
with the IQR band, and the final-ckpt mode line (see
Figures~\ref{fig:gan-collapse},
\ref{fig:sparse-parity-summary}, \ref{fig:task-switch},
\ref{fig:input-topology}, and the three-layer
Figure~\ref{fig:grokking_three_layers_ibbs}).
Table~\ref{tab:ibbs_summary} shows seed-consistent
post-event integer multiplicities on the four Group~B
sharp-transition experiments, with the one exception
of sparse parity (modal $\hat h_1 = 7$ in $5/10$ seeds,
remaining seeds in $\{1, 2, 4, 5\}$). The
inter-multiplet gap there is clean (log gap
$\sim 1.4$) but intra-band gaps within the leading $7$
contrasts are small (log gap $\sim
0.1$--$0.2$), so the gap walk sometimes triggers early.
The geometry is an $8$-vertex atomic configuration
indexing the $2^k = 8$ patterns of the $k = 3$
relevant bits, with the parity label one of the $7$
between-vertex contrasts. The Group~A diagnostic
correctly returns no stable multiplet on the three
diffusive-relaxation experiments and a smooth
$\hat h_1 = 2$ mode-coverage fingerprint on the
8-Gaussian GAN. Pre-transition windows of every
Group~B experiment also fall on the
not-yet-formed-manifold side, calibrating the toolkit
on both sides of manifold formation.

Five structural regimes organise the reading. Table~\ref{tab:regimes} consolidates the regime-aware
reading of $\hat h_1$ and slots every experiment into
one of the rows there. The cases that identify a
manifold split into three regimes (Fourier-soliton,
atomic-cluster, smooth/product) plus the
diffusive-relaxation refusal row. The synthetic task
switch sits in the atomic-cluster row geometrically
($\mathbb Z_4$ four-vertex configuration) but reads as
an intermediate \emph{doublet} case with
$\hat\beta_{\rm corr}(d{=}2) = 1.724 \pm 0.038$,
slightly above the BBS-asymptotic target $1.5$, the
matrix image of a closed $\mathbb Z_4$-equivariant
loop. The 8-Gaussian GAN is the table's atomic-cluster
row geometrically (full $\mathbb Z_8$ coverage) but
Group~A trajectory-wise (smooth mode-coverage growth
with no sharp $\beta(t)$ jump), so it appears in
Table~\ref{tab:ibbs_summary} as a Group~A entry with
the cluster-coverage interpretation rather than as a
band-structured transition.

\begin{table}[h]
\centering
\small
\setlength{\tabcolsep}{6pt}
\renewcommand{\arraystretch}{1.15}
\begin{tabularx}{\linewidth}{@{}l X X@{}}
\toprule
Regime & What $\hat h_1$ means & Experiments \\
\midrule
Smooth manifold & $\hat h_1 = d$ for $S^{d-1}$
(BBS asymptotic) &
synthetic positive controls
(Appendix~\ref{app:calibration}); possibly the output
logits of grokking \\
Product manifold & factor-wise multiplets per factor &
grokking upstream embedding-layer factors
($T^2 = S^1 \times S^1$, after better sampling) \\
Fourier soliton & active $\sin/\cos$ mode pairs on a
discrete $\mathbb Z_k$-equivariant subset of $S^{d-1}$ &
modular-arithmetic transformer
($\hat h_1 = 12$, six pairs) \\
Atomic cluster & vertex / between-cluster contrast count
on a $\mathbb Z_k$-equivariant finite set &
sparse parity ($\mathbb Z_2$), task switch
($\mathbb Z_4$), input topology ($\mathbb Z_2$
boundary), 8-Gaussian GAN ($\mathbb Z_8$ octagon) \\
Diffusive / no band & no stable geometric inference &
MNIST mod-WD, multi-output regression \\
\bottomrule
\end{tabularx}
\caption{Regime-aware reading of the multiplet
diagnostic $\hat h_1$. Only the smooth-manifold and
product-manifold rows invoke the literal BBS asymptotic
$h(1, d) = d$. The Fourier-soliton and
atomic-cluster rows give finite combinatorial readings
of the same integer. The diffusive row reports the
toolkit's correct refusal to infer a low-dimensional
geometry. The 8-Gaussian GAN sits in the
atomic-cluster row geometrically (full $\mathbb Z_8$
coverage) but trajectory-wise as Group~A: it has a
stable doublet fingerprint but no sharp transition step.}
\label{tab:regimes}
\end{table}
\par
The residual-RMT bulk is the robust noise signature
across the programme. In every seed of every experiment
the residual $R(t) = M(t) - \hat M^{(d)}(t)$ is peaked
with curvature outliers rather than a clean Wigner
semicircle (residual Wigner-KL $\sim 1.5$ on the Group~B
sharp transitions at $N = 1000$), the RSM signature into
which the earlier additive-Gaussian and heat-kernel
embeddings collapse at $O(\epsilon^2)$
\cite{halperin2026IBBS}, the predicted noise class for
transformer and MLP activations driven by a
Gaussian-additive post-LayerNorm process. This residual
reading holds across architectures (transformer, MLP, CNN
generator) and task types (classification, regression,
generative). The blind $\ell = 2$ angular-momentum
component, the primary RSM/FSM discriminator on a clean
sphere, is defined only where a multiplet is resolved, and
there it is FSM-like in every such experiment (sparse
parity, task switch, GAN, and all three transformer
layers), rising well above the synthetic boundary. It is
undefined where the representation stays a singlet or
diffuse (regression, input-topology). This elevated
$\ell = 2$ most plausibly tracks intrinsic even-$\ell$
angular structure in the representation rather than an FSM
noise channel, so we report it alongside the residual
reading rather than in its place. The synthetic
separability null of Appendix~\ref{app:noise-null} confirms
the $\ell = 2$ component separates RSM from FSM with zero
seed overlap on the clean-sphere null.

As a roadmap, each subsection below pairs one combined figure with
geometric content not captured by the tables: the MDS
embedding of $M(t)$ where the representation cloud has
a visualisable topology, an explicit order parameter
on the supervisory or input-symmetry side where one is
defined, and the unique features (bagel formation for
grokking; upstream product-of-spheres decomposition
for the transformer factors; output-logit cleanup)
that the matrix reading exposes.

\subsection{MNIST + MLP: feature-emergence transient and joint
train--test sync}
\label{subsec:mnist_results}

\begin{figure}[htbp]
\centering
\includegraphics[width=\textwidth]{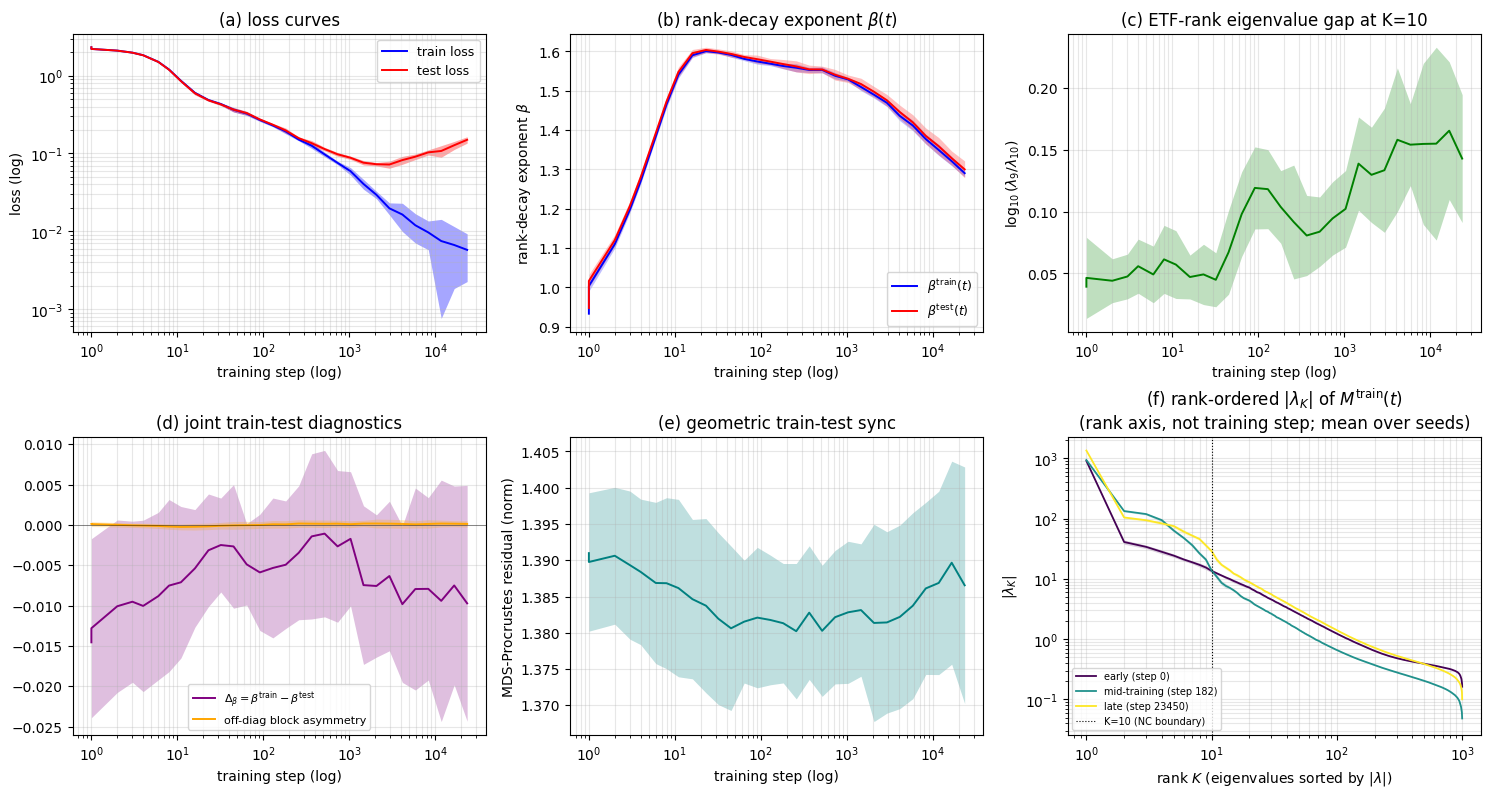}
\caption{Spectral diagnostics across MNIST + MLP, 50
epochs, 30 log-spaced checkpoints, mean $\pm 1\sigma$
across $10$ seeds.
(a)~Train/test CE loss;
(b)~$\beta(t)$ for train (blue) and test (red);
(c)~eigenvalue gap $\log(\lambda_9 / \lambda_{10})$
at the neural-collapse boundary $K = 10$;
(d)~$\Delta_\beta(t)$ (purple) and combined-sample
off-diagonal block asymmetry (orange);
(e)~MDS--Procrustes residual between bottom-three
eigenspaces of $M^{\rm train}$ and $M^{\rm test}$;
(f)~$M^{\rm train}(t)$ eigenvalue spectra at three
snapshots in log--log; dotted vertical line marks $K = 10$.}
\label{fig:mnist_collapse}
\end{figure}

Setup details in Table~\ref{tab:axes} (row 5.1). The
three-layer MLP $784 \to 256 \to 256 \to 10$ is trained
with AdamW (learning rate $10^{-3}$, weight decay
$10^{-3}$) for $50$ epochs on MNIST. The
penultimate-layer activations are $L^2$-normalised onto
$S^{255}$ and the arccos distance matrix is formed on a
fixed $N = 1000$ eval sample at $30$ log-spaced
checkpoints. The expected long-time effective dimension
under saturated neural collapse is $d_\beta \to C - 1 = 9$;
at the moderate weight decay used here we expect partial
collapse, with $d_\beta$ relaxing from $\sim 80$ at
initialisation to $\sim 3$--$4$ at the end of training.
Figure~\ref{fig:mnist_collapse} reports the result.

The non-monotonic $\beta(t)$ in panel~(b) is the main
qualitative finding. $\beta$ rises from $0.93$ to
$\beta_{\max} \approx 1.60$ in the first $\sim 30$
training steps ($d_\beta$ falls to $\sim 2.7$), the
spectral signature of \emph{feature emergence}: the
network collapses the random high-dimensional
representation into a low-dimensional task-relevant
cloud well before the cross-entropy loss settles. $\beta$
then relaxes to $\sim 1.29$ over the subsequent
$\sim 10^{4}$ steps ($d_\beta \approx 4.5$), the
spectral signature of \emph{feature refinement} as the
network populates more directions for finer-grained
class distinctions. The scalar test loss shows neither
phase as a kink. The non-monotonic $\beta(t)$ resolves
both cleanly.

The eigenvalue gap $\log(\lambda_9 / \lambda_{10})$
in panel~(c) is the Papyan--Han--Donoho neural-collapse
diagnostic \cite{papyan2020neural}: at the moderate
weight decay used here the gap opens to $\sim 0.14$ in
log-units, the predicted direction but below the
saturated regime. The dedicated high-WD probe in the
online appendix opens it cleanly to $\sim 0.21$.

The joint train--test diagnostics in panels (d) and (e) provide
the calibration this baseline is designed to produce:
$\Delta_\beta(t)$, the block asymmetry of $M^{\rm joint}(t)$,
and the MDS--Procrustes residual stay small and roughly
stationary throughout, with $|\Delta_\beta| \lesssim 0.02$ and
block asymmetry $\lesssim 0.002$. Train and test
representations sit on geometrically aligned manifolds for the
entire run. The grokking transformer of
Section~\ref{subsec:grokking_results}, by contrast, exhibits a
sharp $\Delta_\beta$ spike at the transition.

The I-BBS Algorithm~1 readout (Table~\ref{tab:ibbs_summary})
declines to identify a sub-manifold: the gap walk at
$\tau = 0.25$ does not close in any of the $10$ seeds
(no stable $\hat h_1$), and the corrected delocalised
slopes at $d_{\rm guess} \in \{2, 3, 4\}$ all sit below
the BBS-admissibility floor $\beta = 1$. The
matrix-valued observable still resolves the
feature-emergence and feature-refinement phases
through the non-monotonic $\beta(t)$, but $d_\beta(t)$
is here a continuous cluster-geometry signal rather
than a literal latent dimension. The residual-RMT
verdict at a forced reconstruction depth
$\hat d_\beta = 4$ is RSM in $10/10$ seeds.

\subsection{Multi-output regression: output-level matrix as
loss generalisation}
\label{sec:results-regression}

\begin{figure}[htbp]
  \centering
  \includegraphics[width=\textwidth]{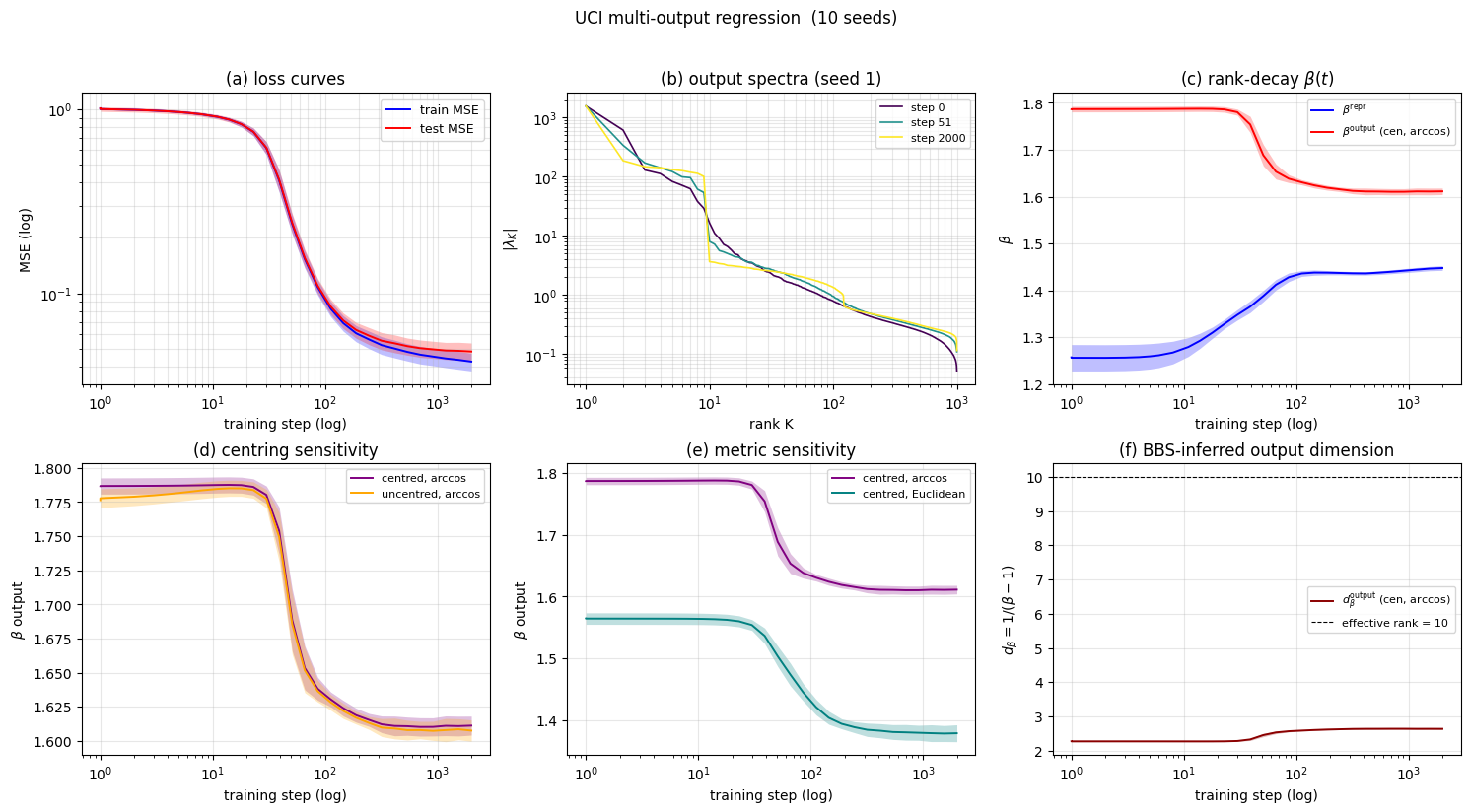}
  \caption{Multi-output regression on synthetic data
  ($20 \to 8$ targets, latent effective rank $10$), mean
  $\pm 1\sigma$ across $10$ seeds.
  (a)~Train/test MSE;
  (b)~output-level spectra at three snapshots (single seed);
  (c)~$\beta_{\rm repr}(t)$ and $\beta_{\rm loss}(t)$
  moving in opposite directions;
  (d)~centring-sensitivity overlay;
  (e)~arccos vs Euclidean metric overlay;
  (f)~$d_\beta^{\rm loss}$ vs latent rank $10$ (dashed).}
  \label{fig:regression}
\end{figure}

\begin{figure}[htbp]
  \centering
  \includegraphics[width=0.62\textwidth]{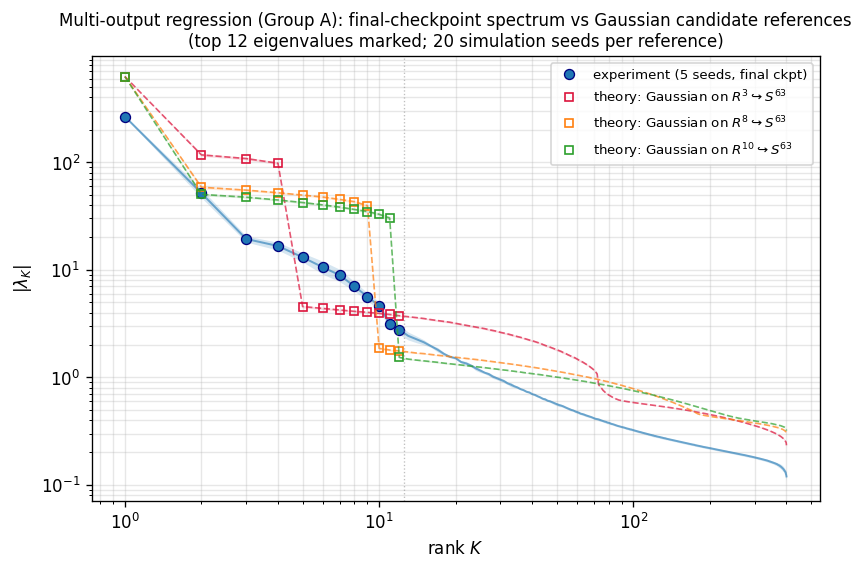}
  \caption{Multi-output regression (Group~A): final-checkpoint
  spectrum of $M^{\rm train}_{\rm repr}$ ($5$ seeds, blue),
  overlaid with three Gaussian candidate references on
  $\mathbb R^d \hookrightarrow S^{63}$ for
  $d \in \{3, 8, 10\}$ ($20$ sim seeds each, RSM noise
  $\epsilon = 0.1$). Leading $12$ eigenvalues marked. The
  match is \emph{partial}: bulk at $K \in [6, 12]$ tracks the
  $d = 10$ reference, but $\lambda_1 \approx 266$ is much
  larger than any reference and the
  $\lambda_2$--$\lambda_5$ shape matches no single $d$. This
  partial-match pattern is the visual companion to the
  $\beta_{\rm del}^{\rm corr} < 1$ verdict in
  Table~\ref{tab:ibbs_summary}: no $\hat{\mathcal M}_d$ is
  identified.}
  \label{fig:regression-spectrum-vs-theory}
\end{figure}

Setup details in Table~\ref{tab:axes} (row 5.2). This is
the only matched-space probe in the diagnostic
programme: predictions and targets live in
$\mathbb{R}^{8}$, the residual matrix $R(t)$ is a
well-defined geometric object, and the scalar MSE loss
is the trace $R R^{\top}/N$ that the matrix observable
generalises. The data come from
\texttt{sklearn.make\_regression} with $20$ features,
$8$ correlated targets, latent effective rank $10$,
additive noise $\sigma = 0.5$. We track the
penultimate-layer features on $S^{63}$ and the
output-level residual cloud through four matrices
(centred/uncentred $\times$ arccos/Euclidean).
$d_\beta^{\rm repr}$ should contract toward the latent
intrinsic dimension as the network learns to ignore
irrelevant directions. $d_\beta^{\rm loss}$ should
reflect the residual cloud geometry rather than the
imposed class structure. Figure~\ref{fig:regression}
summarises the run, and
Figure~\ref{fig:regression-spectrum-vs-theory} overlays
the final-checkpoint spectrum on three Gaussian
candidate references on $\mathbb R^d \hookrightarrow
S^{63}$ for $d \in \{3, 8, 10\}$.

The main observation is in panel~(c): the two regimes of
Section~\ref{sec:framework} produce trajectories that move
in opposite directions. The representation-level
$\beta_{\rm repr}(t)$ rises during training, so
$d_\beta^{\rm repr}$ contracts from $\sim 5$ to $\sim 3.2$.
The output-level $\beta_{\rm loss}(t)$ falls in the same
window and tracks the MSE curve in shape and timing, with
$d_\beta^{\rm loss}$ moving from $\sim 2.3$ to $\sim 2.6$.
The output-level spectrum of panel~(b) shows a knee at
$K \approx 10$ matching the latent effective rank, with a
power-law bulk that contracts as the loss decays.
Centring and arccos-vs-Euclidean ablations (panels d, e)
leave the trajectory shape and transition step unchanged.
The saturated values
$d_\beta^{\rm loss} \approx 2.6$ and
$d_\beta^{\rm repr} \approx 3.2$ both sit well below the
latent effective rank $10$, because the BBS exponent reads
cluster geometry (features and errors) rather than the
latent rank directly.

The I-BBS Algorithm~1 readout
(Table~\ref{tab:ibbs_summary}) declines to identify a
sub-manifold: late-time $\hat h_1 = 1$ on all $10$ seeds
(singlet, no band structure on the residual cloud) and
the corrected delocalised slopes at
$d_{\rm guess} \in \{2, 3, 4\}$ all sit at or below the
BBS-admissibility floor $\beta = 1$. The matrix
observable still resolves the structural trajectory of
representation and residual through opposite-direction
$\beta_{\rm repr}(t)$ and $\beta_{\rm loss}(t)$.
Residual RMT is RSM in $10/10$ seeds.

\subsection{GAN mode collapse on the 8-Gaussian benchmark}
\label{sec:results-gan}

\begin{figure}[htbp]
  \centering
  \includegraphics[width=\textwidth]{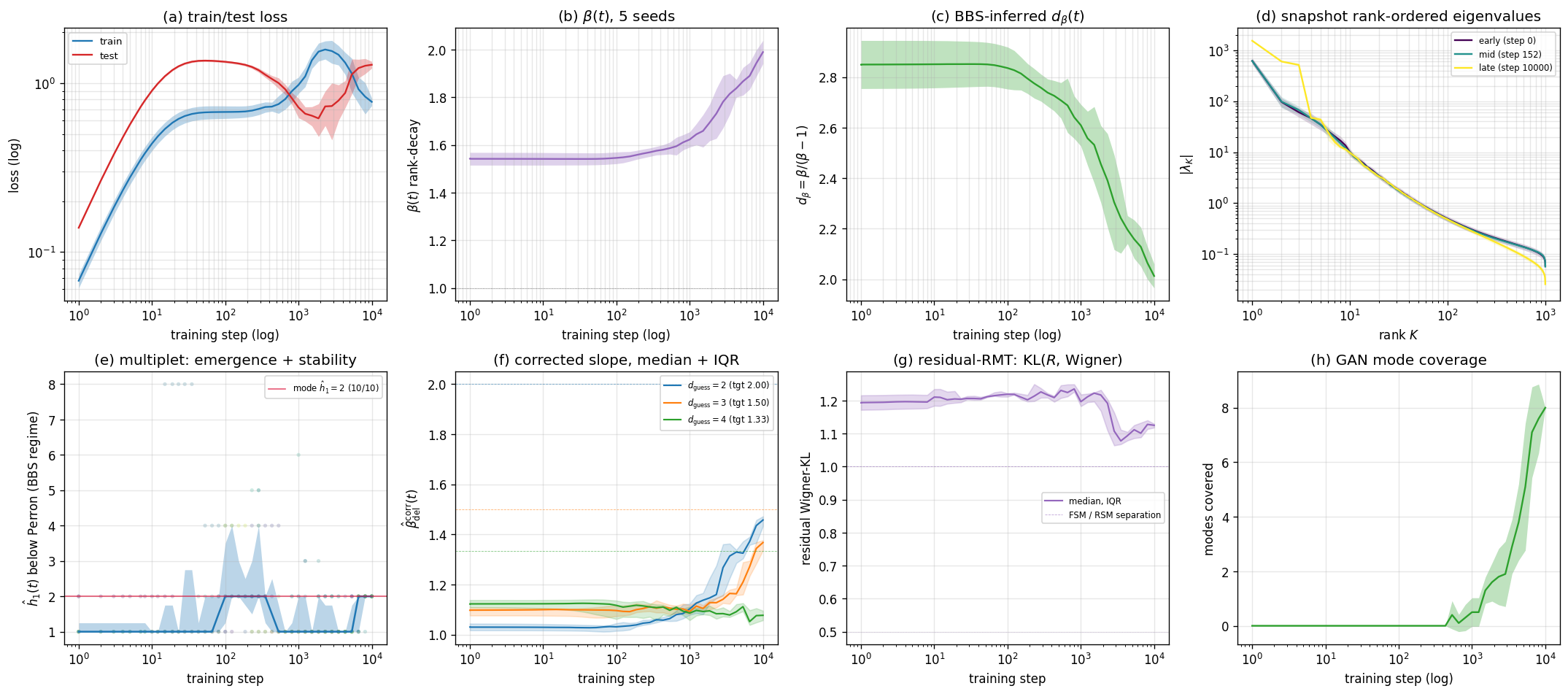}
  \caption{8-Gaussian GAN mode-collapse benchmark, $10000$
  steps. Combined scalar $+$ I-BBS analysis.
  Top row: (a)~train (generator) and test (discriminator)
  losses, $10$-seed mean $\pm \sigma$;
  (b)~rank-decay exponent $\beta(t)$;
  (c)~BBS-inferred dimension $d_\beta = \beta/(\beta-1)$;
  (d)~rank-ordered eigenvalues of $M^{\rm train}_{\rm repr}$ at
  early, mid, and late snapshots (mean across seeds) showing
  the post-collapse band structure forming.
  Bottom row: (e)~multiplet $\hat h_1(t)$ emergence and
  stability with per-seed scatter, cross-seed median$+$IQR
  (interquartile range) band, and final-checkpoint mode
  (red horizontal);
  (f)~corrected delocalised slope median$+$IQR per
  $d_{\rm guess}$ with BBS targets dashed;
  (g)~residual Wigner-KL trajectory with the
  FSM\,/\,RSM separation marked;
  (h)~mode-coverage trajectory (ground-truth diagnostic):
  generator coverage progressing from $0$ modes early
  (collapse) to the full $8$ in every seed by step
  $\sim$$10000$.}
  \label{fig:gan-collapse}
\end{figure}

Setup details in Table~\ref{tab:axes} (row 5.3). The
target is an 8-Gaussian mixture on a regular octagon
in $\mathbb{R}^{32}$ ($\mathbb{Z}_8$ rotational
symmetry). A small GAN with generator $G:
\mathbb{R}^{8} \to \mathbb{R}^{32}$ is trained for
$10000$ steps, long enough for every seed to resolve the
full mode structure. The matrix observable is the arccos
distance matrix on $N = 1000$ generator outputs. The
ground-truth diagnostic is the count of cluster
centres within Euclidean distance $1$ of at least one
generator output. Figure~\ref{fig:gan-collapse}
summarises the result.

Mode coverage in panel~(h) traces the characteristic
small-GAN dynamics: initial collapse near the origin
(zero modes within the detection radius), slow
progression through intermediate coverage, and a final
phase that resolves to the full $8$ modes in every seed
($8.0 \pm 0.0$, 10 seeds) by step $\sim$$10000$. In
lockstep, the BBS exponent
$\beta_{\rm gen}(t)$ rises from $\approx 1.49$ to
$\approx 1.76$ and the inferred dimension
$d_\beta^{\rm gen}$ falls from $\approx 3.05$ to
$\approx 2.32$.

Pairing $d_\beta^{\rm gen}$ against the ground-truth
mode count produces a clean monotonic relationship: as
the generator covers more modes, the BBS dimension
\emph{decreases}. This is the same sign as the
task-switch finding of
Section~\ref{sec:results-taskswitch}: more discrete
cluster structure is read by BBS as \emph{lower}
effective dimension, because the rank-decay is set by
the number and tightness of clusters rather than by
their spread. A fully collapsed generator has the
highest reading (within-mode noise is the only source
of dimension). A fully covered eight-cluster generator
has the lowest (the eight discrete directions dominate
the top of the spectrum). So
$d_\beta^{\rm gen}$ here is not a count of modes but a
geometric organisation signal of the output
distribution. For mode collapse as a categorical event
the ground-truth count remains the operative
diagnostic, while the BBS reading gives a continuous
geometric trajectory in the same direction as the
supervised-collapse trajectory of
Section~\ref{sec:results-taskswitch}.

\begin{figure}[htbp]
  \centering
  \includegraphics[width=0.45\textwidth]{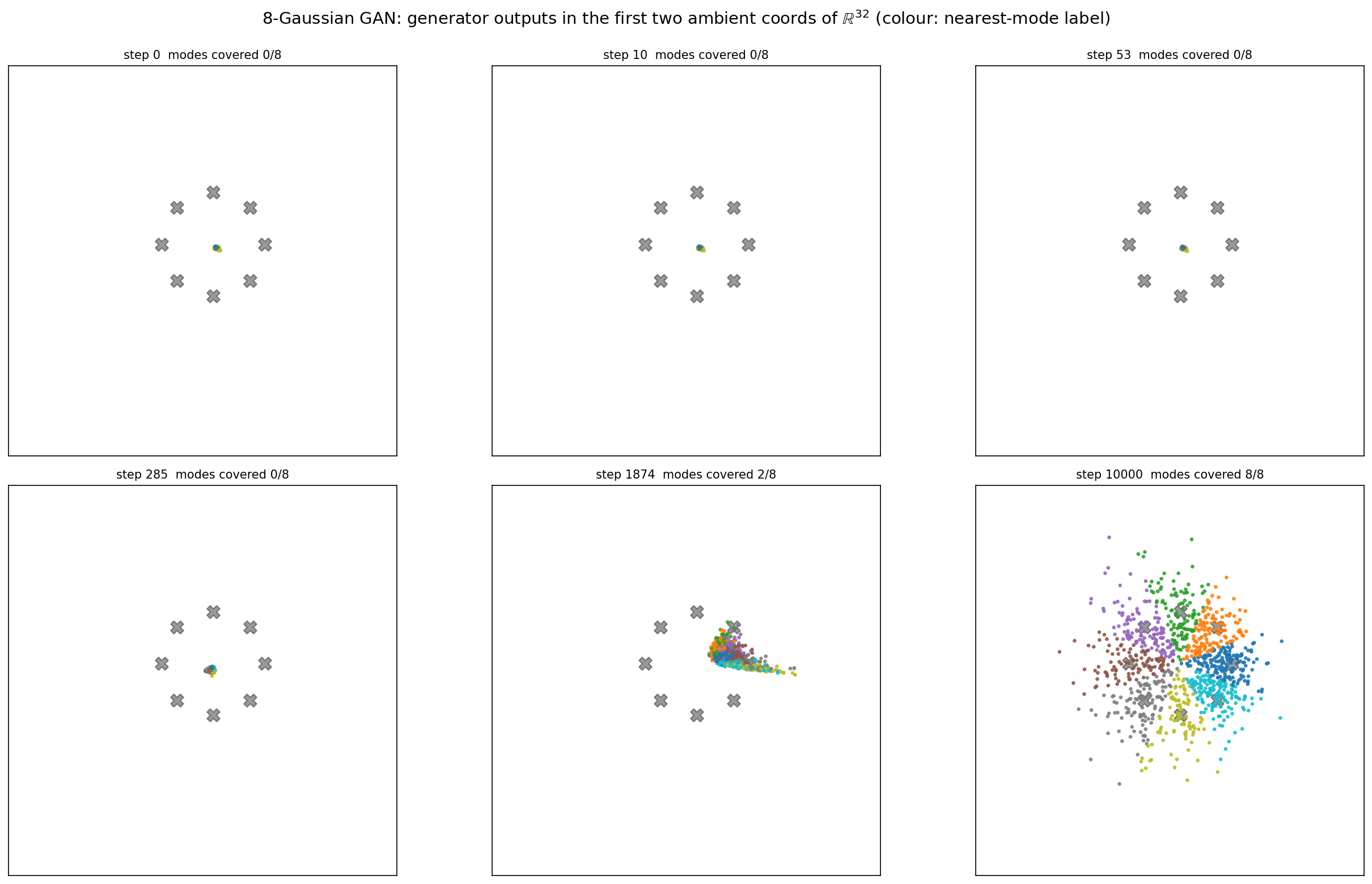}
  \caption{8-Gaussian GAN: generator outputs in the first
  two ambient coordinates of $\mathbb{R}^{32}$ (the natural
  plane containing the eight target centres marked by
  $\times$), points coloured by post-training
  nearest-mode assignment. Full $\mathbb{Z}_8$
  coverage is reached by step $10000$ in this plane; the
  full $\mathbb{R}^{32}$ picture is in the MDS panel
  (c) of Figure~\ref{fig:groupA-mds-combined}.}
  \label{fig:gan-scatter}
\end{figure}

\begin{figure}[htbp]
  \centering
  \begin{subfigure}[t]{0.48\linewidth}
    \includegraphics[width=\linewidth]{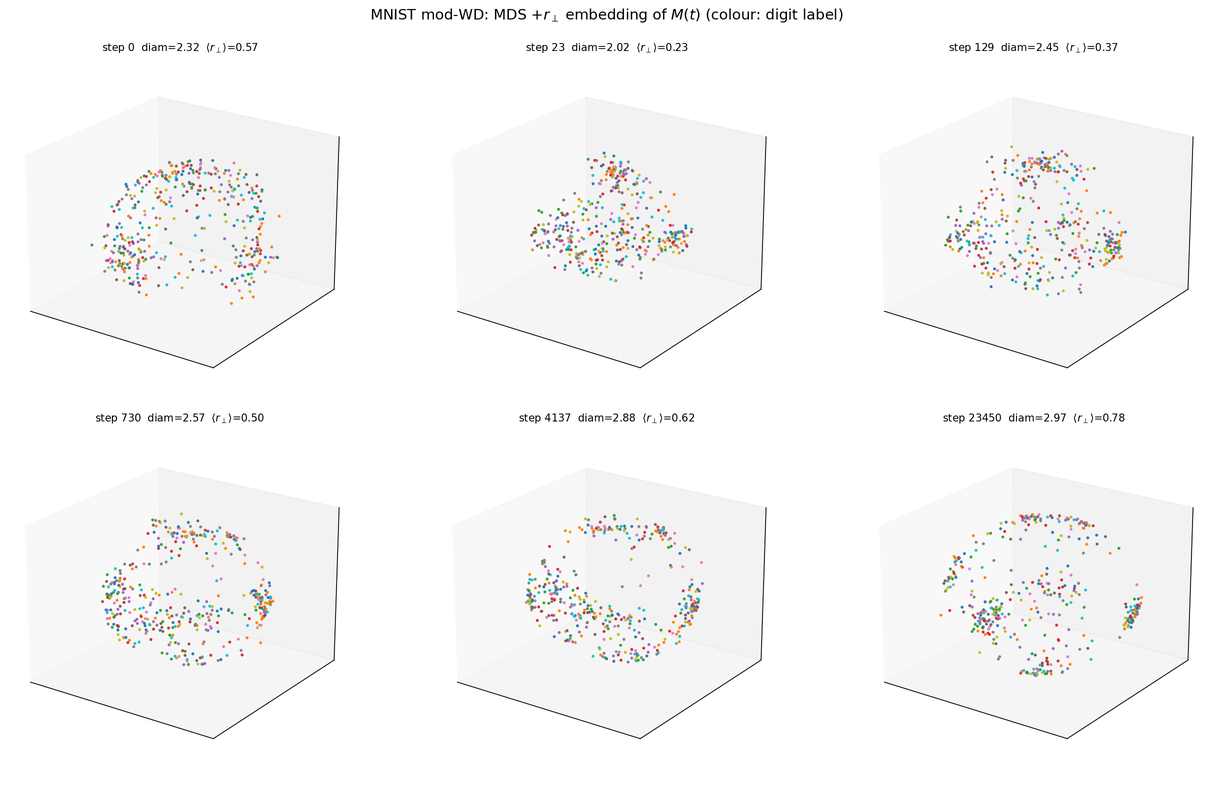}
    \caption{MNIST + MLP (mod-WD), digit label.}
    \label{fig:mnist-mds}
  \end{subfigure}\hfill
  \begin{subfigure}[t]{0.48\linewidth}
    \includegraphics[width=\linewidth]{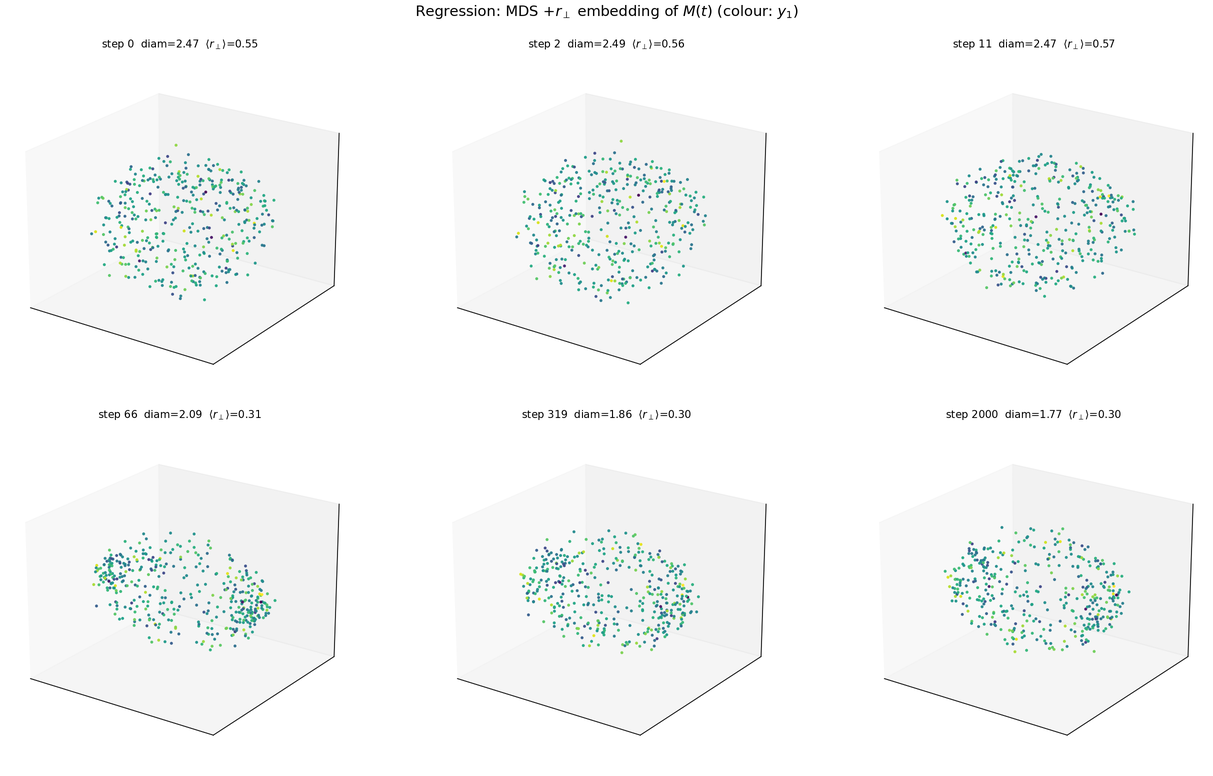}
    \caption{Regression $\mathbb R^{20}\!\to\!\mathbb R^{8}$, coloured by $y_1$.}
    \label{fig:regression-mds}
  \end{subfigure}

  \vspace{6pt}

  \begin{subfigure}[t]{0.48\linewidth}
    \includegraphics[width=\linewidth]{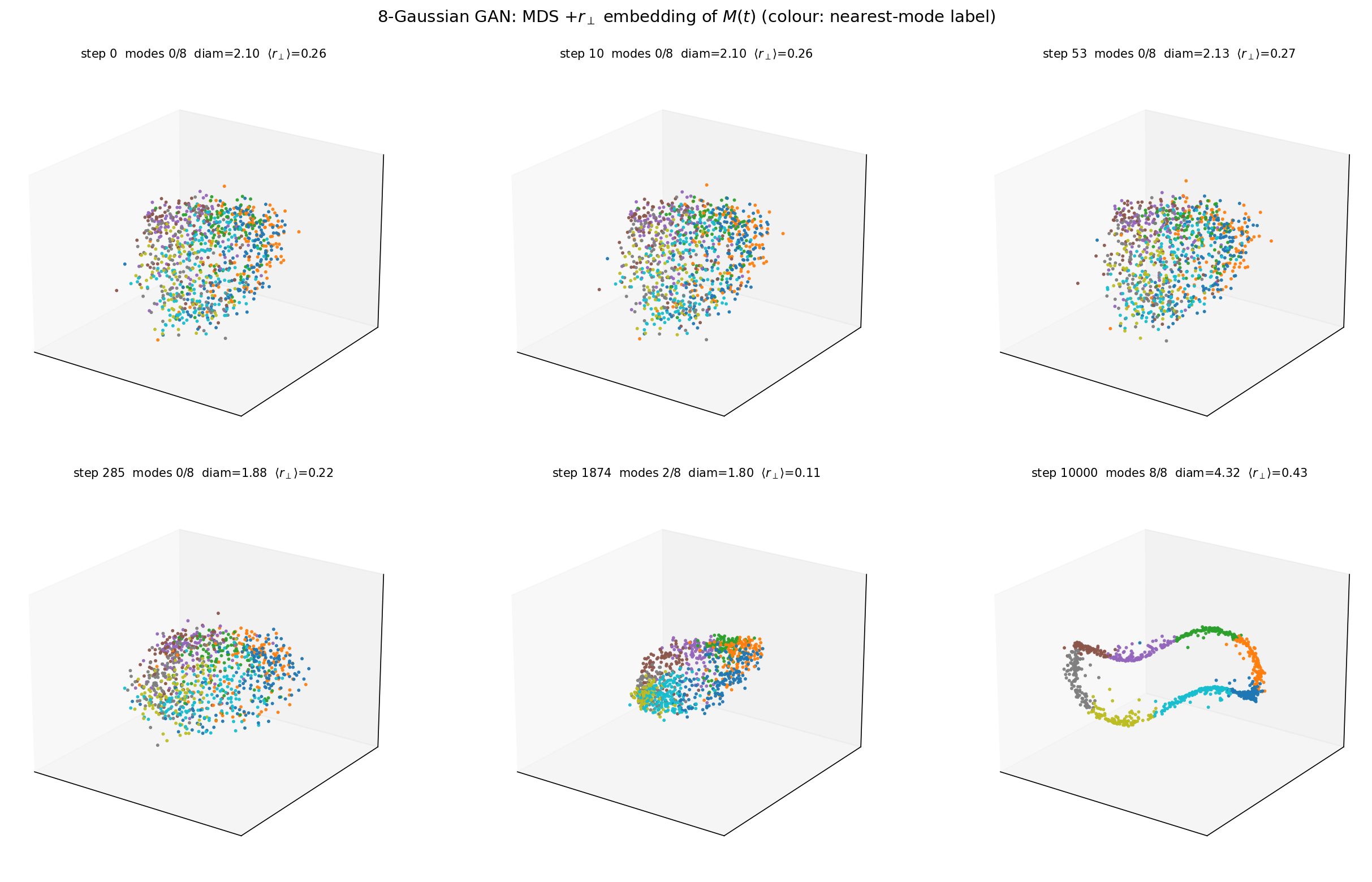}
    \caption{8-Gaussian GAN: $3$-D blob $\to$ eight-mode
    ring by step $10000$, $\langle r_\perp\rangle = 0.45$.}
    \label{fig:gan-mds}
  \end{subfigure}\hfill
  \begin{subfigure}[t]{0.48\linewidth}
    \includegraphics[width=\linewidth]{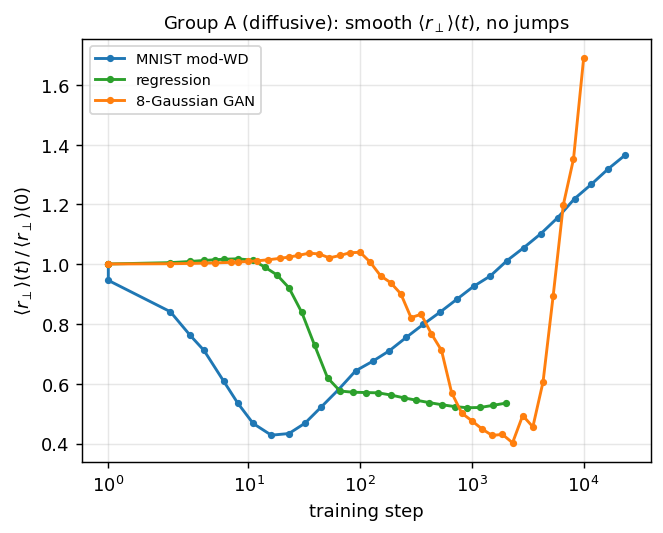}
    \caption{Cross-case $\langle r_\perp\rangle(t)$: smooth,
    no sharp step.}
    \label{fig:groupA-rperp}
  \end{subfigure}
  \caption{Group~A MDS $+ r_\perp$ embeddings of $M(t)$ for
  the three diffusive runs, with (d) the cross-case residual
  $\langle r_\perp\rangle(t)$. Colours: digit label (a),
  $y_1$ (b), nearest-mode (c). Cf.\ the stepped Group~B
  counterpart, Figure~\ref{fig:groupB-mds-combined}.}
  \label{fig:groupA-mds-combined}
\end{figure}

For the geometric trajectory of $M(t)$, Figures~\ref{fig:gan-scatter} and~\ref{fig:gan-mds}
give two complementary views on a fixed-latent
evaluation sample of $N = 1000$ codes (particle identity
preserved across frames). The scatter is the ambient
view on the natural visualisation plane (the first two
coordinates of $\mathbb{R}^{32}$ containing the eight
target centres). The cloud starts at the origin, breaks
$\mathbb{Z}_8$ symmetry by drifting onto a partial arc
by step $\sim 2000$, and by step $10000$ places at least
one sample within Euclidean distance $1$ of every mode
in the plane (full coverage). The MDS embedding (bottom-three
eigenvectors of $M(t)$, with each particle displaced
radially by its off-3D residual $r_\perp^{(i)}$, as in
the diffusive figures) shows the same trajectory from
inside the representation: an initial isotropic $3$-D
blob of diameter $\sim 2$ reorganises into an extended
closed ring of diameter $\sim 4$ that threads all eight
modes, with the off-three-dimensional residual
$\langle r_\perp\rangle$ rising from $\sim 0.26$ to
$\sim 0.45$ rather than shrinking. The two readings are
consistent: in-plane mode coverage is complete, but in
the full ambient $\mathbb{R}^{32}$ no sample lies within
distance $1$ of any centre because the generator still
carries mass (mean norm $\sim 1.8$) in the remaining
$30$ coordinates. The rising residual and the ring
(rather than a flat low-dimensional sheet) are the
geometric signature of the Group~A regime, in which the
discrete $\mathbb{Z}_8$ structure fills out without an
effective reduction of the representation dimension, its
eight-colour labelling becoming visible only at the
end.

The I-BBS Algorithm~1 readout
(Table~\ref{tab:ibbs_summary}) is unanimous at the
standardised $N = 1000$, $10$-seed setup:
$\hat h_1 = 2$ in $10/10$ seeds (doublet, the
matrix fingerprint of the $\mathbb{Z}_8$ octagonal
cluster scaffold, so the smooth-manifold formula
$\hat d = \hat h_1$ of
Section~\ref{subsec:bbs_inference} does not apply). The
doublet is the robust signature of the eight-vertex
scaffold. Residual RMT is RSM in $10/10$ seeds.

Across all three Group~A runs the off-manifold residual
$\langle r_\perp\rangle(t)$ stays substantial and does not
decay (Figure~\ref{fig:groupA-mds-combined}d). This is the
matrix signature that the diffusive representations
undergo no effective reduction of the internal
dimensionality onto a low-dimensional sub-manifold,
consistent with the I-BBS algorithm declining to identify
one in any of them.

\subsection{Modular-arithmetic transformer: spectral
signature of grokking}
\label{subsec:grokking_results}

\begin{figure}[htbp]
\centering
\includegraphics[width=\textwidth]{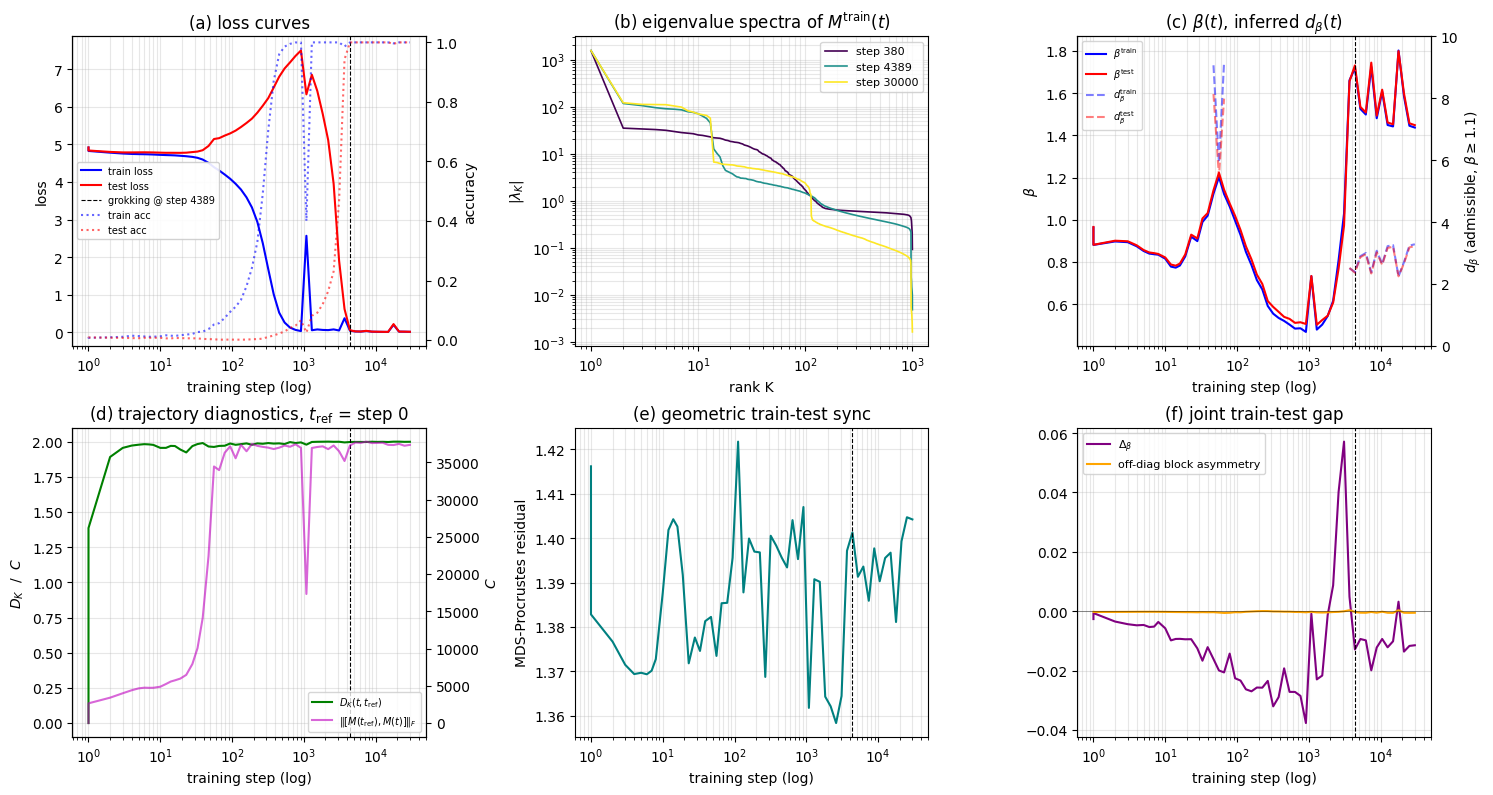}
\caption{Spectral diagnostics across the grokking-transformer
training trajectory (AdamW, $30\,000$ steps). The grokking step (first test
accuracy $> 0.99$) is marked by the dashed vertical line.
(a)~CE loss and accuracy (memorise-then-grok signature);
(b)~eigenvalue spectra of $M^{\rm train}(t)$ at three
snapshots in log--log;
(c)~$\beta(t)$ (solid) and $d_\beta(t)$ (dashed);
(d)~trajectory diagnostics $D_K$ (green) and $C$ (magenta)
from $t_{\rm ref} = 0$;
(e)~MDS--Procrustes residual between
$M^{\rm train}$ and $M^{\rm test}$;
(f)~$\Delta_\beta(t)$ (purple) and combined-sample
off-diagonal block asymmetry (orange).}
\label{fig:grokking_summary}
\end{figure}

Setup details in Table~\ref{tab:axes} (row 5.4). The
canonical small-transformer-on-modular-addition setup of
\cite{power2022grokking, nanda2023progress}: one-layer
decoder-only transformer with $d_{\rm model} = 128$,
$4$ heads, MLP hidden $512$, GELU + post-norm LayerNorm,
trained on $a + b \bmod p$ with $p = 113$, length-3
input $[a, b, =]$, AdamW (lr $10^{-3}$, wd $1.0$) for
$30\,000$ steps, eval $N = 1000$ at $63$ log-spaced
checkpoints. We apply I-BBS Algorithm~1 at $\tau = 0.25$
to step past the intra-band gaps of the Fourier-mode
multiplet (Section~\ref{subsec:bbs_inference}).
Figure~\ref{fig:grokking_summary} collects the
diagnostics. The dashed vertical line marks the
empirical grokking step (first checkpoint with test
accuracy $> 0.99$).

The grokking trajectory provides the strongest piece of evidence
in this paper for the claim that the matrix-valued diagnostic
\emph{strictly improves} on the scalar one. The training set is
fitted to near-zero loss by step $\sim 10^3$, after which the
scalar test loss is uninformative. The test loss does not begin to
drop until step $\sim 3 \times 10^3$ and saturates near zero only
at the grokking step $\sim 4.4 \times 10^3$. By contrast, the
matrix-valued $\beta(t)$ in panel~(c) is already deep in its
transition by step $\sim 10^3$: $\beta$ falls from a
pre-memorisation value of $\beta \approx 1.0$ to its minimum
$\beta_{\min} \approx 0.55$ in the memorisation interval, then
\emph{rises} continuously across the grokking transition to its
post-grokking value $\beta_{\mathrm{post}} \approx 1.4$. The
matrix-valued diagnostic therefore detects the structural
reorganisation \emph{at} the late memorisation phase, not at the
test-loss transition.

The BBS dimension trajectory $d_\beta(t)$ tells a complementary
story. In the late memorisation phase, $\beta < 1$ formally
corresponds to an effective dimension $d_\beta = \beta/(\beta - 1) < 0$,
which is the BBS asymptote's signature that the rank-decay
exponent has fallen below the i.i.d.\ floor for any compact
manifold: the post-memorisation representation cloud is not well
described as an i.i.d.\ sample from a smooth one-particle density
on any $d$-manifold, but as a coarse memorisation lattice whose
spectrum reflects discrete coverage rather than a continuous
density. After the grokking transition $\beta_{\mathrm{post}}
\approx 1.4$ gives $d_\beta \approx 3.5$, in qualitative agreement
with the post-grokking $d_\beta$ measurements of
\cite{halperin2026grokking} on the same family of transformers at
$p = 113$ and the multi-frequency Fourier structure of the
post-grokking representation \cite{nanda2023progress}. An
analogous reading at the post-attention residual (after the
first LayerNorm, instead of the canonical post-MLP choice
shown here) crosses $\beta = 1$ in the same step window:
the transition step is layer-synchronous across the
residual stream while the absolute post-transition
plateau depends on the layer (post-attention sits closer
to the BBS prediction $\beta = 2$ for a clean $S^1$
bagel. Post-MLP carries more residual off-bagel structure).

The joint train--test diagnostic $\Delta_\beta(t)$ in
panel~(f) provides a third, independent piece of evidence for the
spectral toolkit being a strictly finer diagnostic than the
scalar loss. At and just before the grokking step,
$\Delta_\beta$ shows a sharp positive spike of magnitude
$\approx 0.06$: the rank-decay exponent of the train-set
distance matrix briefly exceeds that of the test-set distance
matrix, indicating that the train representations have already
reorganised into the post-grokking low-dimensional structure
while the test representations have not yet caught up. After the
grokking step the spike vanishes and $\Delta_\beta$ returns to
small values. This is the cleanest example in our experiments of
the joint object \eqref{eq:M_pair} producing a spectral signal
that has no counterpart in either of the single-matrix
diagnostics: a sharp temporal localisation of the precise
generalisation moment from the spectral gap between train and
test.

The combined-sample block asymmetry of $M^{\mathrm{joint}}(t)$
and the MDS--Procrustes residual in panel~(e) both stay
near zero throughout: mean inter-point distances and
bottom-three eigenspaces are train-test compatible at
every stage. The structural change of the transition is
not visible at these levels, only in the higher-order
spectral structure that $\Delta_\beta$ resolves.

For the downstream I-BBS diagnosis, Algorithm~1 at $\tau = 0.25$ identifies a sub-manifold
on the post-grokking residual stream across the $10$
seeds. The threshold is chosen so the gap walker steps
past the small intra-band gaps of the Fourier-mode
multiplet and stops at the deep gap separating
multiplet from bulk. Pre-grokking the spectrum has no
log-gap exceeding $\tau = 0.25$ on most seeds, while
immediately after the grokking step the multiplet
diagnostic jumps to a median $\hat h_1 = 12$ across
the $10$ seeds (final mean$\,\pm\,$std
$= 11.7 \pm 1.4$, IQR $[11, 12]$, mode $12$ in $5/10$,
range $[9, 14]$). The integer $\hat h_1 = 12$ is the
matrix fingerprint of $6$ active Fourier-mode
pairs (each $\sin/\cos$ doublet contributing one to the
top of the spectrum), consistent with
\cite{nanda2023progress}. The corrected slope selects
$\hat d_\beta = 3$ within $\sim 5\,\%$ of target (full
numbers in Table~\ref{tab:grokking_layers}), the
effective dimension the $(\,=\,)$-token spectrum reads
from the combination of the two-circle input
$(a, b) \in S^1 \times S^1$ and the answer-circle
$S^1(a + b)$ that modular addition closes onto. The
residual-RMT verdict is RSM in $10/10$ seeds
(Wigner-KL $2.08 \pm 0.6$), the I-BBS-predicted
$O(\epsilon^2)$ regime where the off-manifold component
is dominated by the deterministic curvature kernel
$\epsilon^2\cot M^{(d)}$ (the leading small-$\epsilon$
correction of the RSM kernel) rather than by a
Wigner-random residual. The upstream and output-logit readouts below
confirm the same six-Fourier-pair structure on
$S^1(a)$, $S^1(b)$, and on the logit cloud.

\begin{figure}[htbp]
\centering
\includegraphics[width=\textwidth]{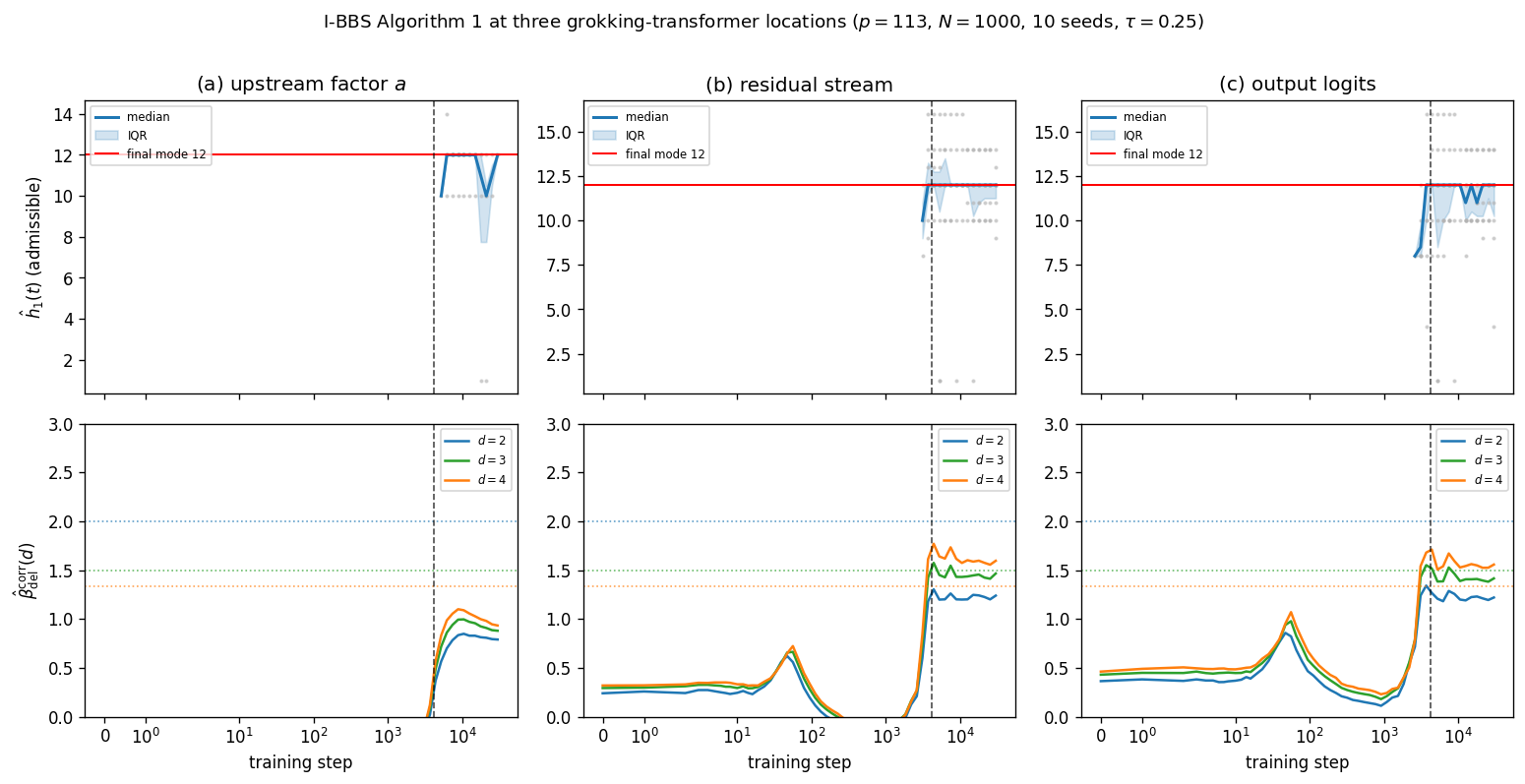}
\caption{I-BBS Algorithm~1 at three locations of the
grokking transformer. Top row: multiplet $\hat h_1(t)$ excluding
the Perron, gated on the BBS-admissibility criterion
$\beta > 1$, with per-seed scatter, cross-seed median
$+$ interquartile range (IQR) band, and final-checkpoint
mode (red horizontal). Bottom row: corrected delocalised
slope
$\hat\beta_{\rm del}^{\rm corr}(d_{\rm guess})$ for the
candidate dimensions, with dashed lines at the BBS
targets $d/(d-1)$. Columns: (a, d)~upstream embedding
factor $a$ at input-token position $0$; (b, e)~residual
stream at the $(\,=\,)$-token position post final
LayerNorm; (c, f)~output logits. Vertical dashed line on
every panel marks the median grokking step across seeds.
All three readouts return mode $\hat h_1 = 12$
post-grokking (six Fourier mode pairs), and the slope
diagnostic sharpens monotonically from the BBS-floor at
the upstream factor to a clean $\hat d_\beta = 3$ at the
residual stream and the logits.}
\label{fig:grokking_three_layers_ibbs}
\end{figure}

\begin{figure}[htbp]
\centering
\includegraphics[width=0.83\textwidth]{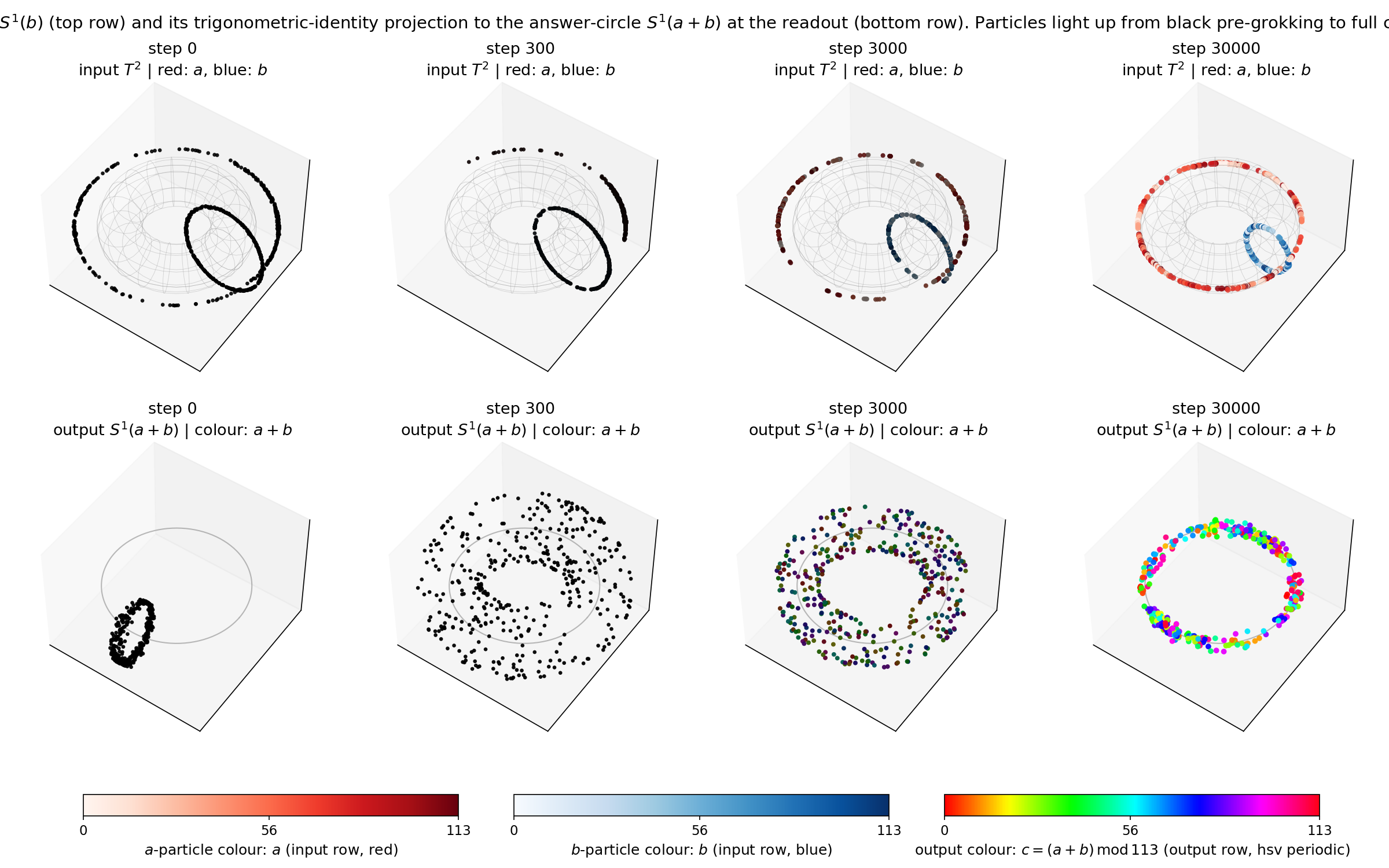}
\caption{Bagel formation, reproduced from
\cite{halperin2026grokking}.
\emph{Top row}: input representation at training-step
snapshots, condensing onto a 2-torus
$T^2 = S^1(a) \times S^1(b)$; $a$-particles (red) on the major
loop, $b$-particles (blue) on the minor loop.
\emph{Bottom row}: downstream answer-circle $S^1(a+b)$ at the
readout, with particles coloured by $c = (a+b) \bmod p$ on a
periodic hsv map. From left to right: random initialisation
(step 0), late memorisation (step 300), at the grokking step
(step 3000), and post-grokking (step 30\,000). The
representation-space companion of the matrix trajectories
of Figure~\ref{fig:grokking_summary}: the upstream
$\mathbb{Z}_p \times \mathbb{Z}_p$ and downstream $\mathbb{Z}_p
\cong S^1$ task symmetries, absent from the random
initialisation, are realised at the transition.}
\label{fig:grokking_bagel}
\end{figure}

Bagel formation is the symmetry-restoring side of the
matrix picture. The matrix trajectories of
Figure~\ref{fig:grokking_summary} have a geometric
companion on the representation side, developed in
\cite{halperin2026grokking} and reproduced in
Figure~\ref{fig:grokking_bagel}. The input
residual-stream activations sit in a compact
unstructured blob at step $0$, disperse over the
memorisation plateau, and at the grokking step
condense onto a 2-torus
$T^2 = S^1(a) \times S^1(b)$ with $a$- and
$b$-particles on the two generating circles. The
downstream readout simultaneously condenses onto a
1-circle $S^1(a + b)$ coloured by $c = (a + b) \bmod p$.
This joint event is \emph{bagel formation}, interpreted
in \cite{halperin2026grokking} as a symmetry-restoring
phase transition that realises the task's
$\mathbb Z_p \times \mathbb Z_p$ input symmetry and
$\mathbb Z_p \cong S^1$ output symmetry on the
representation side at the transition.

The post-transition torus and circle produce the
$\beta_{\rm post} \approx 1.4$ plateau in
Figure~\ref{fig:grokking_summary}(c), which under the
I-BBS finite-$N$ correction
\cite{halperin2026IBBS} translates to
$\hat d_\beta = 3$ (uncorrected $\approx 3.5$ absorbs
the $\sim 0.5$ finite-$N$ shift). The
$(\,=\,)$-token readout sits downstream of attention
where upstream $T^2$ ($S^1{\times}S^1$) and downstream
$S^1$ ($d = 2$) are mixed, with $\hat d_\beta = 3$ the
resulting effective dimension. The
$\Delta_\beta(t)$ spike at the grokking step is the
matrix imprint of the bagel forming first on the
train sample, only then on the test sample.

\label{para:upstream-product-spheres}
We turn to the upstream product-of-spheres I-BBS analysis. The embedding-layer activations at the two input-token
positions are computed independently from the
corresponding token embeddings (cross-position mixing
enters only through the subsequent attention block), so
the upstream representation is the canonical setting
for the product-of-spheres construction of
Section~\ref{subsec:observables} with $\mathcal M =
T^2 = S^1(a) \times S^1(b)$. We retrain $10$ seeds,
$L^2$-normalise the saved $h^{(a)}_i, h^{(b)}_i \in
\mathbb R^{128}$ at positions $0$ and $1$ onto their
own spheres, and form the per-factor squared-distance
matrices $N_a(t), N_b(t)$ on which Algorithm~1 runs
independently. The latent $T^2$ matrix is $\hat
N^{(T^2)} = \hat N^{(S^1)}_a + \hat N^{(S^1)}_b$
(Eq.~\eqref{eq:N-decomp}). Three diagnostics: per-factor
noise $\eta_k(t) = \|\epsilon_k N^{(1, k)}\|_F /
\|N^{(D_k)}\|_F$, joint noise $\eta(t) = \|R(t)\|_F /
\|N^{(D)}\|_F$ with $R$ the residual, and factorisation
commutator $\kappa(t) = \|[N_a, N_b]\|_F /
(\|N_a\|\,\|N_b\|)$.

\begin{figure}[htbp]
\centering
\includegraphics[width=\textwidth]{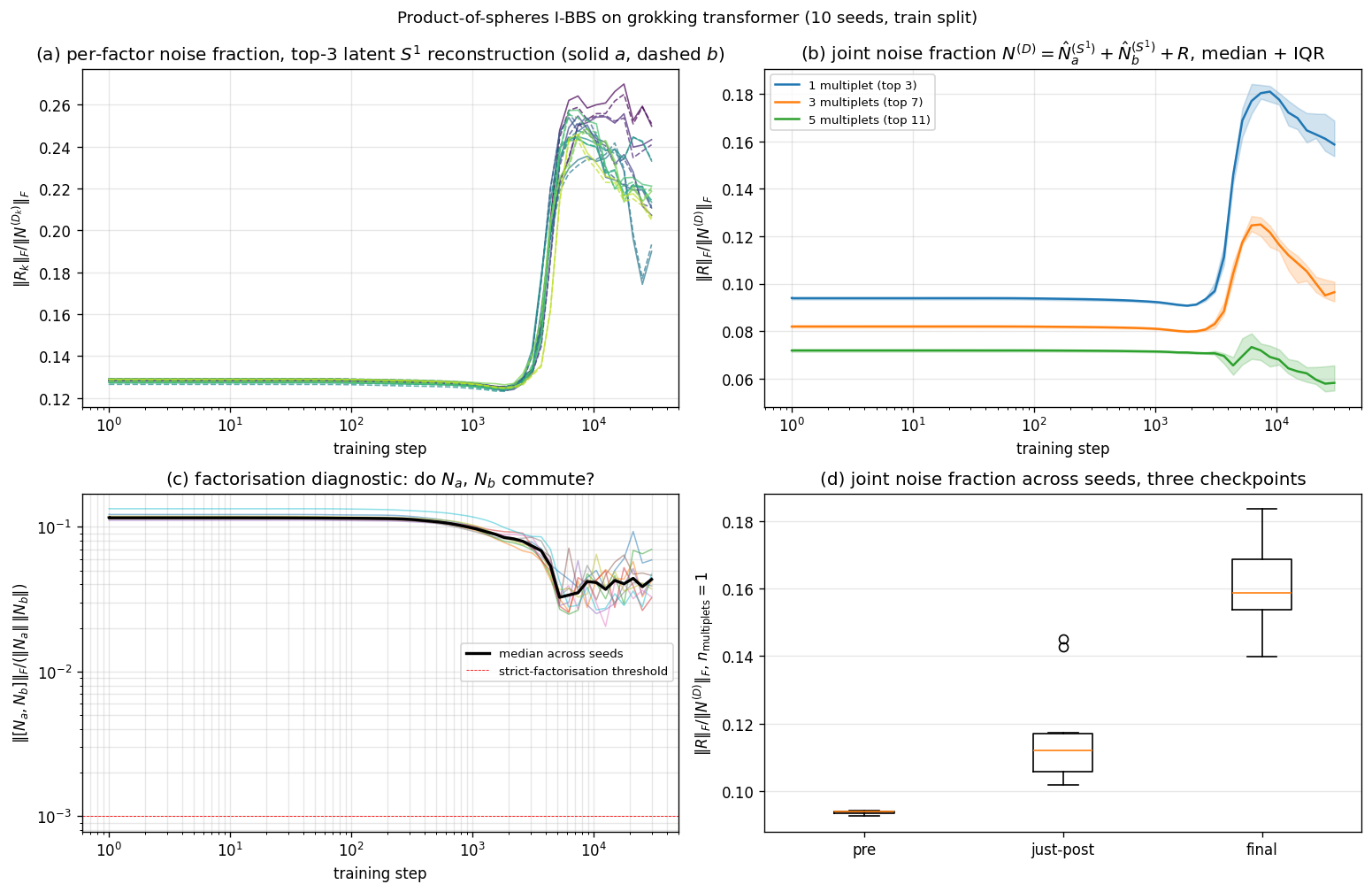}
\caption{Upstream product-of-spheres I-BBS analysis on
the re-trained grokking transformer. Per-factor and joint
residuals of the decomposition
Eq.~\eqref{eq:N-decomp} at fixed reference truncations
that bracket the natural per-seed Algorithm-1 truncation
$1 + \hat h_1^{(k)} \approx 13$ at the final checkpoint.
(a)~Per-factor noise fraction $\eta_k(t) = \|R_k\|_F /
\|N^{(D_k)}\|_F$ at the strict top-3 reconstruction
(Perron $+$ $\ell = 1$, token-$a$ solid, token-$b$
dashed).
(b)~Joint noise fraction $\eta(t) = \|R\|_F /
\|N^{(D)}\|_F$ at three truncation levels (top-3 / top-7
/ top-11 eigenvalues retained per factor). At the natural
per-seed truncation the final-checkpoint joint noise sits
near the top-11 trajectory ($\eta = 0.06 \pm 0.03$, see
text).
(c)~Relative commutator $\kappa(t) = \|[N_a, N_b]\|_F /
(\|N_a\|\,\|N_b\|)$.
(d)~Distribution across seeds of the joint noise fraction
at the strict top-3 reconstruction at pre, just-post,
and final checkpoints (box plot).}
\label{fig:grokking_upstream}
\end{figure}

For the upstream I-BBS diagnosis, per-factor Algorithm~1 returns mode $\hat h_1^{(a)} =
12$ in $6/10$ seeds and $\hat h_1^{(b)} = 12$ in $5/10$,
with remaining seeds either at the same $12$-mode band
($\hat h_1 \in \{10, 14\}$) or on a smaller singlet /
doublet outlier (Table~\ref{tab:grokking_layers}).
Each factor realises the same $6$-Fourier-mode soliton
structure the downstream readout finds on
$S^1(a + b)$, applied to its own input-token position.
The embedding layer maps each token in $\mathbb Z_p$ to
a discrete subset of $S^1$ populated by $\sim 6$ active
Fourier-mode pairs, the matrix fingerprint of the
condensation of \cite{nanda2023progress}. The corrected
slope at $d_{\rm guess} = 2$ sits at
$\hat\beta_{\rm del}^{\rm corr} \approx 0.93$ on both
factors, at the BBS-admissibility boundary $\beta = 1$,
as expected for a discrete Fourier-mode soliton rather
than a smooth $S^1$ (Appendix~\ref{app:token-dedup}
shows the same integer readout is invariant under
deduplication to the $113$ unique tokens per position).
The latent reconstruction at the natural truncation
($1 + \hat h_1^{(k)}$ eigenvalues per factor) captures
$\sim 93\%$ of each factor's norm ($\eta_a = 0.07 \pm
0.05$, $\eta_b = 0.08 \pm 0.05$), the joint noise drops
to $\eta = 0.06 \pm 0.03$, and the commutator $\kappa(t)$
falls monotonically from $\sim 0.115$ to $0.045 \pm
0.011$ (the embedding layer's per-position structure
sharpening; the Kronecker-sum identity is approached
but not exactly realised on the $1000$-pair eval set,
see Figure~\ref{fig:grokking_upstream}). The
residual-RMT verdict is RSM on both factors
(Wigner-KL $\approx 2.5$).

\label{para:grokking-logits}
Turning to the output-logit I-BBS analysis, the output-logit cloud is a third representation
downstream of the residual stream, related to it by the
linear unembedding map. We apply the same single-sphere
pipeline by $L^2$-normalising per-input logits
$\ell_i(t) \in \mathbb R^{114}$ onto $S^{113}$ and
running Algorithm~1 on the resulting $M^{\rm logit}(t)$
(Figure~\ref{fig:grokking_three_layers_ibbs}, right
column). The integer readout matches the residual-stream
and upstream values (mode $\hat h_1 = 12$ in $4/10$
seeds, the same six-Fourier-pair structure), and the
corrected slope is the \emph{tightest} match to a
smooth BBS template among the three readouts
($\hat\beta_{\rm del}^{\rm corr}(d{=}3) = 1.493 \pm
0.169$ against target $1.500$, $\hat d_\beta = 3$ in
$9/10$ seeds), consistent with the unembedding removing
the residual representational mixing that the
$(\,=\,)$-token post-LayerNorm activation still carries.
The logit cloud thus reads as a $d = 2$ latent geometry
(the $T^2 \to S^1$ image of modular addition) realised
through the same Fourier-pair structure. Residual-RMT
verdict RSM in $10/10$ seeds (Wigner-KL $2.17 \pm
0.48$).

To summarise the three layers, the four readouts at the three locations
(Table~\ref{tab:grokking_layers}) tell a single coherent
story: each location realises the same
$6$-Fourier-pair structure on whichever unit sphere the
local representation lives on. The slope diagnostic
sharpens monotonically from the BBS-admissibility floor
at the upstream factors (discrete $113$-point subset of
$S^1$) to a clean $\hat d_\beta = 3$ at the residual
stream and the logits.

\begin{table}[h]
\centering
\small
\begin{tabularx}{\linewidth}{@{}>{\raggedright\arraybackslash}p{0.18\linewidth} c c c X c@{}}
\toprule
Readout
& $\hat h_1$ (mode/seeds)
& $\hat d_\beta$
& $\hat\beta_{\rm del}^{\rm corr}$ (target)
& Latent geometry
& $\ell_2$ FSM \\
\midrule
Upstream factor $a$ &
$12$ ($6/10$) &
\rule{0pt}{1.0em}$\beta\!<\!1$ &
$0.93$ at $d{=}2$ &
6-Fourier-pair soliton on $S^1$ &
$8/10$ \\
Upstream factor $b$ &
$12$ ($5/10$) &
\rule{0pt}{1.0em}$\beta\!<\!1$ &
$0.93$ at $d{=}2$ &
6-Fourier-pair soliton on $S^1$ &
$9/10$ \\
Residual stream at $(\,=\,)$ &
$12$ ($5/10$) &
$3$ &
$1.43$ ($1.50$) &
6-Fourier-pair soliton on $S^1(a{+}b)$ &
$9/10$ \\
Output logits &
$12$ ($4/10$) &
$3$ &
$1.493$ ($1.500$) &
6-Fourier-pair $d{=}3$ image of $T^2 \to S^1$ &
$9/10$ \\
\bottomrule
\end{tabularx}
\caption{I-BBS Algorithm~1 readouts at three layers of
the grokking transformer (modular addition).
The multiplet count $\hat h_1$ is the gap-walk output
below the Perron eigenvalue, the integer fingerprint of
the active Fourier mode pairs. Here $\hat d_\beta$ is the
slope-based dimension, and $\hat\beta_{\rm del}^{\rm corr}$
is the corrected delocalised slope at the best-fit
$d_{\rm guess}$. The four readouts agree on the same
6-Fourier-pair structure; the slope diagnostic
sharpens from the boundary regime at the upstream
factors (the embedding-layer activations live on a
discrete 113-point subset of the circle) to a clean
$d_\beta = 3$ at the residual stream and logits. The
residual-RMT bulk is RSM-shaped in $10/10$ seeds at every
layer; the last column is the blind $\ell=2$ FSM-like
seed count.}
\label{tab:grokking_layers}
\end{table}

Two structural features of the modular-addition setup
limit the literal Kronecker-sum interpretation of
\cite{halperin2026IBBS}: each position-$k$ embedding
depends only on its own input token, so across the
$1000$ random eval pairs each token in
$\mathbb Z_p$ ($p = 113$) appears on average $\sim 9$
times per position and the per-factor matrices $N_a$,
$N_b$ are rank-degenerate by construction
(effective rank $\le p = 113$); and the random eval
set does not exhaust the $p^2 = 12{,}769$ pair grid, so
the strict factorisation regime is approached but not
realised. Appendix~\ref{app:token-dedup} confirms that
deduplicating to the $113$ unique per-position
embeddings preserves the readouts ($\hat d_\beta = 4$,
RSM verdict invariant; $\hat h_1$ modal preserved
with outlier seeds corrected to the high-multiplet
band), so the body-text $T^2$ verdict is a property of
the per-factor spectrum, not of the random sampling.

\subsection{Sparse-parity learning: endogenous transition
from feature discovery}
\label{sec:results-sparseparity}

Setup details in Table~\ref{tab:axes} (row 5.5). A
two-layer MLP of width $H = 256$ is trained by AdamW
(lr $10^{-3}$, weight decay $0.01$, batch size $64$) on
$k = 3$ parity over $d = 30$ i.i.d.\ bits for
$3 \times 10^4$ steps. The supervised loss is invariant
under $\mathbb{Z}_2$ label flip and under the
$\mathbb{Z}_2^{d-k}$ irrelevant-bit subgroup. The
transition is endogenous (input and target i.i.d.\
throughout). The only source of a transition event is
the optimiser's discovery of the $k$ relevant bits in
a $\binom{d}{k}$-large search space, driven by the
strong weight-decay schedule of \cite{power2022grokking}.
A $\mathbb{Z}_2$ contrast order parameter
$\mathcal{O}(t) = \langle M_{ij}\rangle_{y_i \ne y_j} /
\langle M_{ij}\rangle_{y_i = y_j}$ is the matrix
counterpart of a magnetisation:
$\mathcal{O} = 1$ when the parity label carries no
geometric content in $M(t)$, $\mathcal{O} > 1$ when the
representation has separated the two parity classes.
The post-transition geometry is in fact richer than the
naive $S^0 = \{-\hat n, +\hat n\}$ smallest-faithful
realisation of the output $\mathbb Z_2$: the network
distinguishes all $2^k = 8$ values of the $k$ relevant
bits separately, so the cloud condenses onto an
$8$-vertex atomic configuration, with the parity label
appearing as one of the $7$ between-vertex contrast
directions.

\begin{figure}[htbp]
  \centering
  \includegraphics[width=0.70\textwidth]{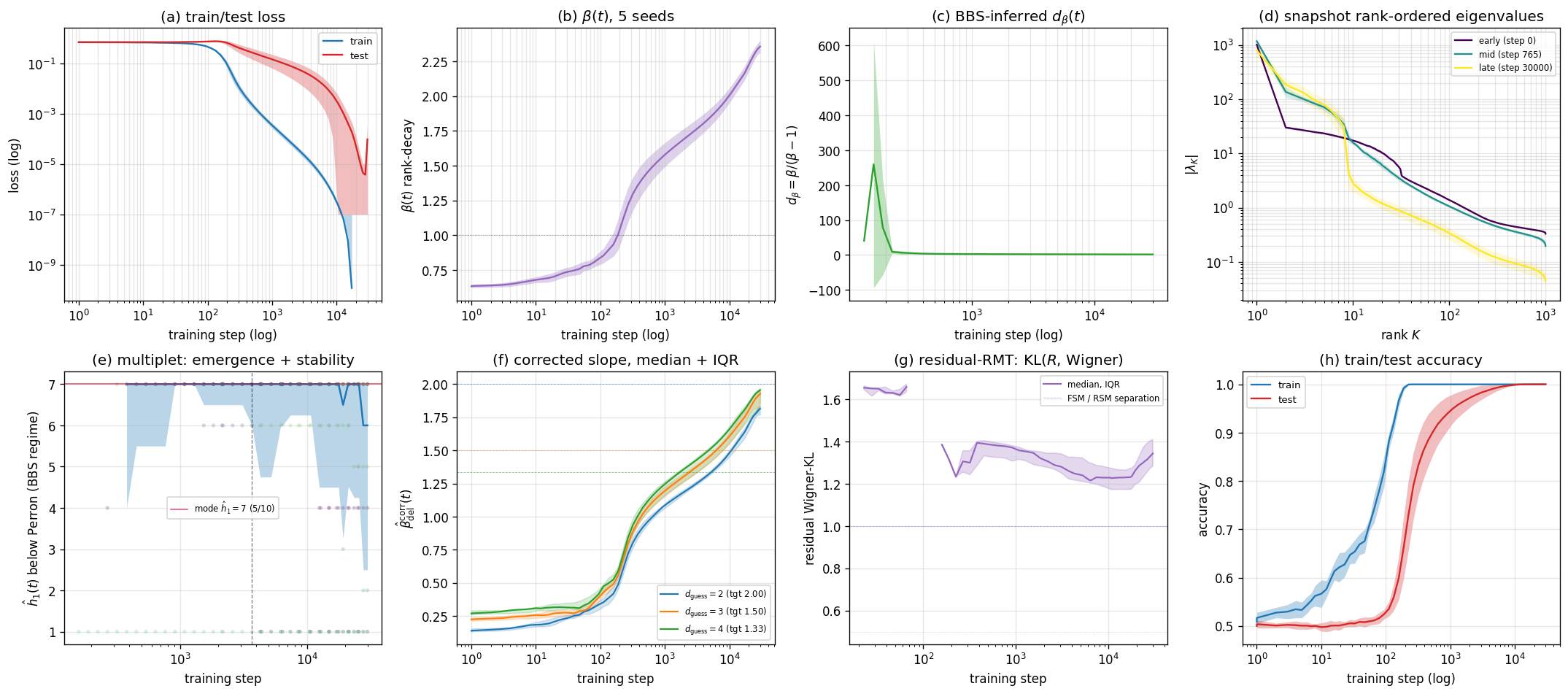}
  \caption{Sparse-parity learning, $k = 3$ parity on
  $\{-1, +1\}^{30}$;
  combined scalar $+$ I-BBS analysis. Top row:
  (a)~train/test cross-entropy losses (semilog-$y$,
  mean $\pm \sigma$); (b)~rank-decay exponent $\beta(t)$;
  (c)~BBS dimension $d_\beta = \beta/(\beta - 1)$;
  (d)~rank-ordered eigenvalues of $M^{\rm train}_{\rm repr}$
  at early/mid/late snapshots, exposing the post-transition
  atomic-cluster band. Bottom row: (e)~multiplet
  $\hat h_1(t)$ with per-seed scatter, cross-seed
  median$+$IQR band, and final-checkpoint mode (red);
  (f)~corrected delocalised slope (median$+$IQR per
  $d_{\rm guess}$); (g)~residual Wigner-KL;
  (h)~train/test accuracy. Dashed vertical line marks the
  median transition step.}
  \label{fig:sparse-parity-summary}
\end{figure}

Panels~(a)--(b) give the scalar baseline: both
accuracies sit at chance ($\approx 0.5$) for the first
$\sim 150$ steps while the network searches for the
relevant bits, climb sharply between steps $\approx 200$
and $\approx 275$, and saturate from step $\sim 300$,
with test cross-entropy still decreasing as weight decay
sparsifies the solution. Unlike grokking, the test loss
reacts in step with the train loss (feature-discovery
rather than train-saturate-then-grok), and the transition
window has relative width $\sim 2 \times 10^{-3}$ on a
$5 \times 10^{4}$-step run.

Panel~(c) shows the matrix-valued $\beta(t)$:
pre-transition at the BBS floor $\beta \approx 1$
(unstructured high-dimensional blob), rising
continuously across the transition window to its
post-transition plateau. Panel~(d) translates to
$d_\beta(t) = \beta/(\beta - 1)$: a low effective dimension
consistent with the $S^0$ antipodal-pair target ($d = 0$,
finite spectral $d_\beta$ from finite-sample softening of
the delta clusters).

Panel~(e) shows the $\mathbb{Z}_2$ contrast order
parameter $\mathcal{O}(t)$. Both $\mathcal O$ and test
accuracy sit on their symmetric values
($\mathcal O \approx 1$, test $\approx 0.5$) for
$\sim 150$ steps and depart in the same $\sim 100$-step
window. Test accuracy saturates by step $\sim 300$ while
$\mathcal O$ continues a slower secondary climb as
weight decay sharpens the two-cluster geometry.
$\mathcal O$ plays the magnetisation role in the Landau
analogue of this transition, the parity counterpart
of the $\Delta_\beta(t)$ spike of
Figure~\ref{fig:grokking_summary}(f).

Panel~(f) shows the rank-ordered spectra at three
snapshots: pre-transition's long log-log decay
(high-dimensional cloud) sharpens post-transition to a
clean $8$-eigenvalue leading band (Perron plus $7$
between-vertex contrasts) separated from the bulk by a
log gap of $\sim 1.4$, the spectral fingerprint
of the $8$-vertex atomic configuration.

The I-BBS Algorithm~1 readout
(Table~\ref{tab:ibbs_summary}) reflects this
$8$-vertex post-event geometry. The inter-multiplet gap
between $\lambda_8$ and $\lambda_9$ is large
(log gap $\sim 1.4$) and seed-stable, so the
walk reliably stops at $K^\star = 8$ when started
beyond the intra-band region. Intra-band gaps within
the leading $7$ contrasts are small
(log gap $\sim 0.1$--$0.2$), so the canonical
$\tau = 0.25$ gap walk modally returns $\hat h_1 = 7$
in $5/10$ seeds but triggers early on the other seeds
($\hat h_1 \in \{1, 2, 4, 5\}$). The
$8$-vertex reading is corroborated by direct $k$-means
clustering of the post-transition activations $H_{\rm
train}$: the within-cluster variance drops by a factor
of $\sim 20$ between $k = 4$ and $k = 8$, then
plateaus, identifying $k = 8$ as the natural cluster
count. The corrected delocalised slope $\hat\beta_{\rm
del}^{\rm corr}(d{=}1) \approx 1.87$ sits above the BBS
target $\beta = 2$ (slope flatter than smooth-$S^1$),
the quantitative signature of finite-cluster
condensation distinct from the smooth-manifold regime
of grokking. Residual RMT is RSM-like in $10/10$
seeds (Wigner-KL $1.56 \pm 0.08$).

\begin{figure}[htbp]
  \centering
  \begin{subfigure}[t]{0.48\linewidth}
    \includegraphics[width=\linewidth]{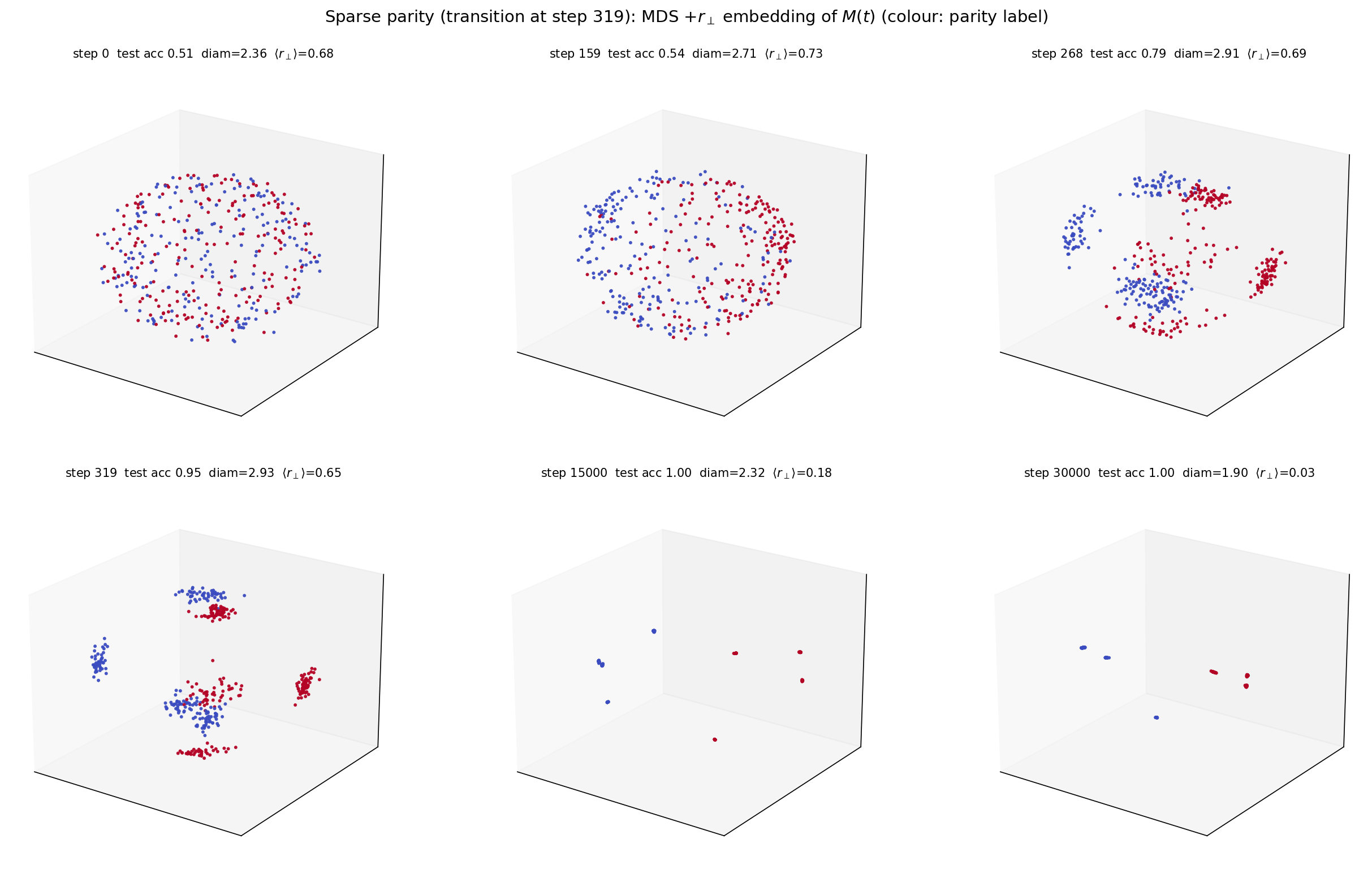}
    \caption{Sparse parity ($S^0$ post-transition).}
    \label{fig:sparse-parity-mds}
  \end{subfigure}\hfill
  \begin{subfigure}[t]{0.48\linewidth}
    \includegraphics[width=\linewidth]{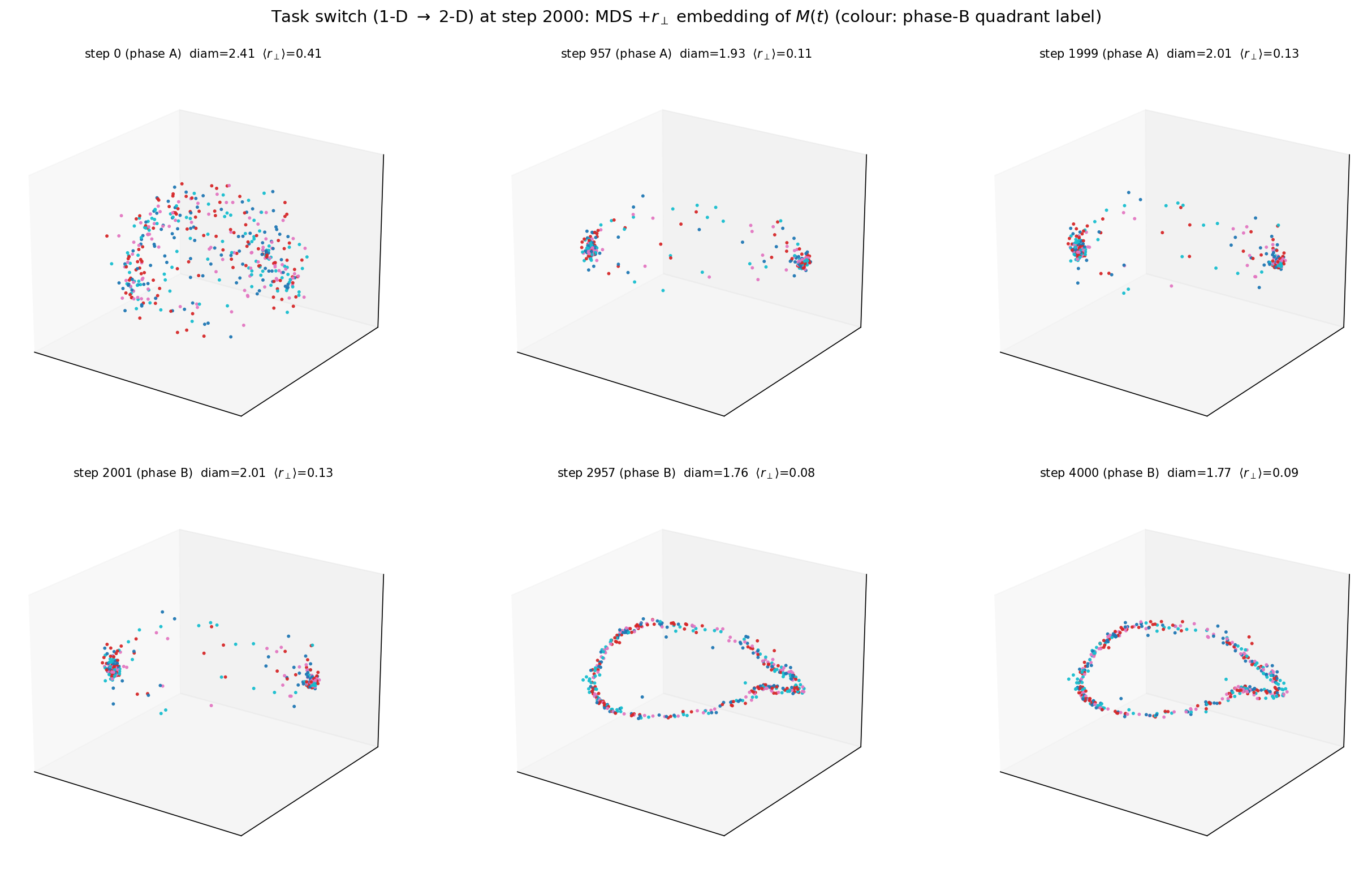}
    \caption{Task switch ($\mathbb Z_4$ on $S^1$, post-switch).}
    \label{fig:task-switch-mds}
  \end{subfigure}

  \vspace{6pt}

  \begin{subfigure}[t]{0.48\linewidth}
    \includegraphics[width=\linewidth]{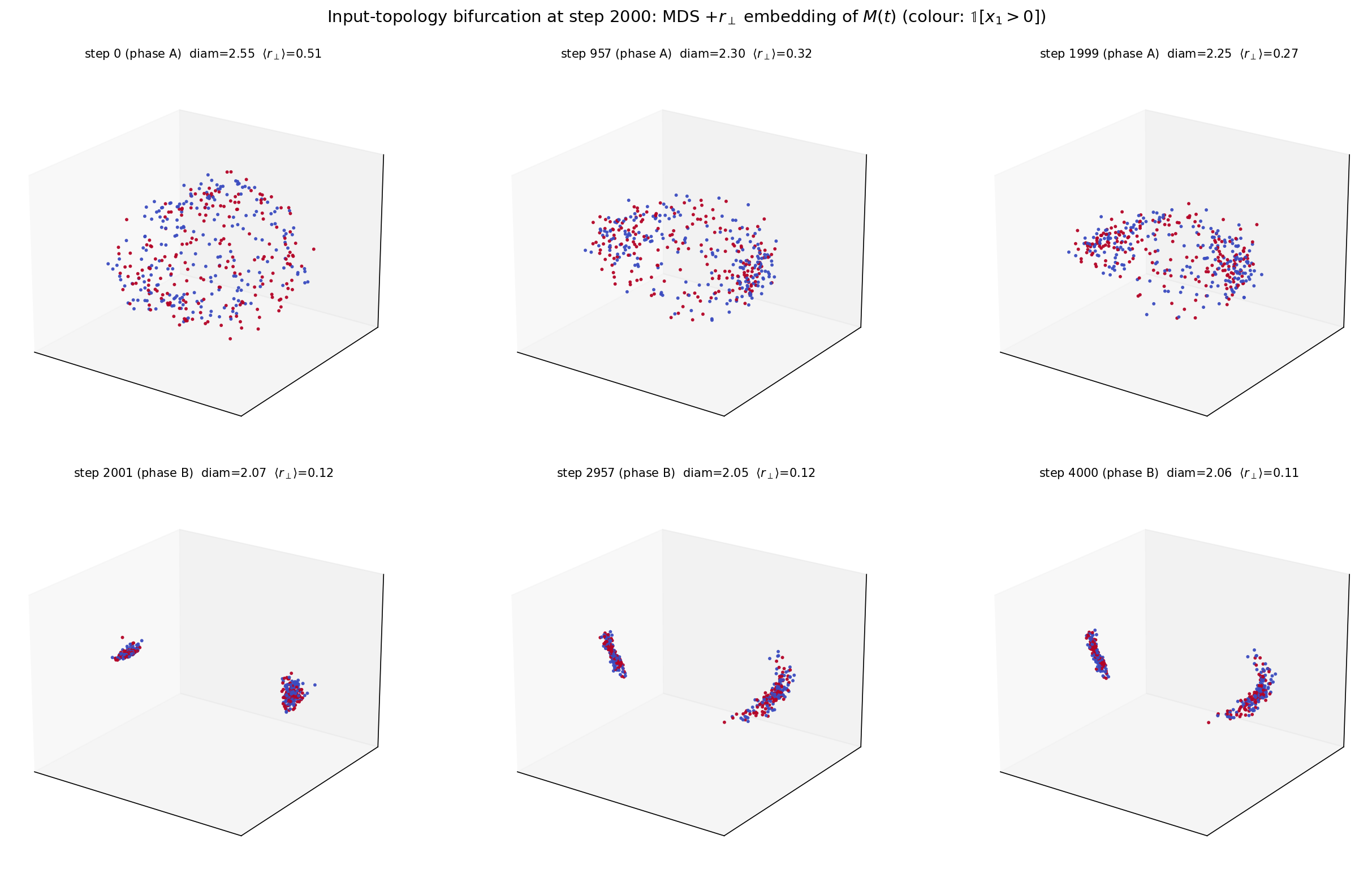}
    \caption{Input-topology bifurcation ($S^0$ boundary case).}
    \label{fig:input-topology-mds}
  \end{subfigure}\hfill
  \begin{subfigure}[t]{0.48\linewidth}
    \includegraphics[width=\linewidth]{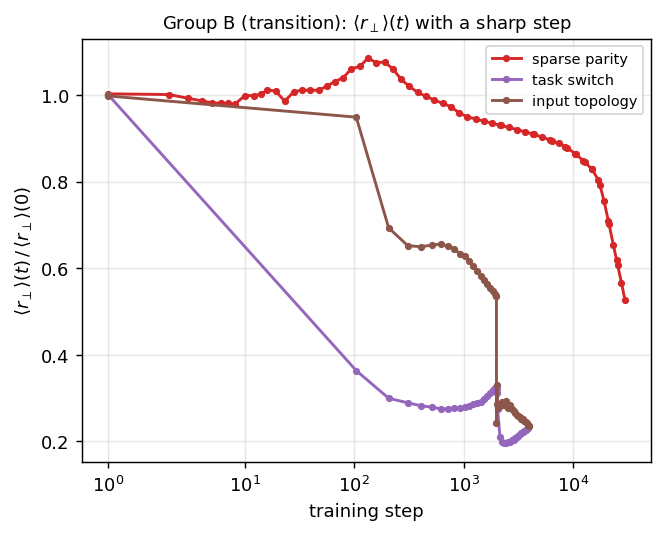}
    \caption{Cross-case $\langle r_\perp\rangle(t)$: each
    transition run shows a sharp step.}
    \label{fig:groupB-rperp}
  \end{subfigure}
  \caption{Group~B MDS $+ r_\perp$ embeddings of $M(t)$
  (same construction as
  Figure~\ref{fig:groupA-mds-combined}), with (d) the
  cross-case off-manifold residual
  $\langle r_\perp\rangle(t)$. Colour codes: parity label
  $y$ in (a); phase-B quadrant in (b); cluster cut
  $\mathbf 1[x_1 > 0]$ in (c).}
  \label{fig:groupB-mds-combined}
\end{figure}

Symmetry-breaking is visible on the representation side. The geometric companion to
Figure~\ref{fig:sparse-parity-summary} is
Figure~\ref{fig:sparse-parity-mds}. At step $0$ the
penultimate-feature activations sit in an isotropic blob
with the two parity classes mixed. The cloud disperses
during the search phase. Well after the transition the
cloud has collapsed onto an $8$-vertex atomic scaffold
indexing the $2^k = 8$ patterns of the $k$ relevant
bits, with the parity label one of the $7$
between-vertex contrast directions. The bottom-three
MDS panels colour the particles by parity label, which
visually highlights the $\mathbb Z_2$ partition of the
$8$ vertices into $4 + 4$ same-parity sub-clusters. The
finer $8$-vertex structure is what the matrix
$\hat h_1 = 7$ reading captures and what the $k$-means
elbow at $k = 8$ confirms in activation space.

\subsection{Synthetic task switch: spectral response to an
abrupt structural change}
\label{sec:results-taskswitch}

Setup details in Table~\ref{tab:axes} (row 5.6). The
input distribution is a 2-D Gaussian latent
$z = (z_1, z_2)$ lifted into $\mathbb{R}^{20}$ via
$W \tanh(z) + \varepsilon$, with $W$ a random
orthonormal-column projection. The supervised target is
switched at step $2000$ from
$y^{(A)} = \mathbf{1}[z_1 > 0]$ (output symmetry
$\mathbb{Z}_2$) to
$y^{(B)} = 2\,\mathbf{1}[z_1>0] + \mathbf{1}[z_2>0]
\in \{0, 1, 2, 3\}$ (output symmetry $S_4$). The
output head is rebuilt at the switch and the
optimiser state is reset. The MLP has hidden width
$32$, representation on $S^{31}$. The expected
representation dimension is $\approx 2$ in phase A
(the 2-D input manifold leaves both directions
available to the binary head) and $\approx 1$ in
phase B (four-way classification drives the cloud
onto a $4$-vertex atomic arrangement).

\begin{figure}[htbp]
  \centering
  \includegraphics[width=\textwidth]{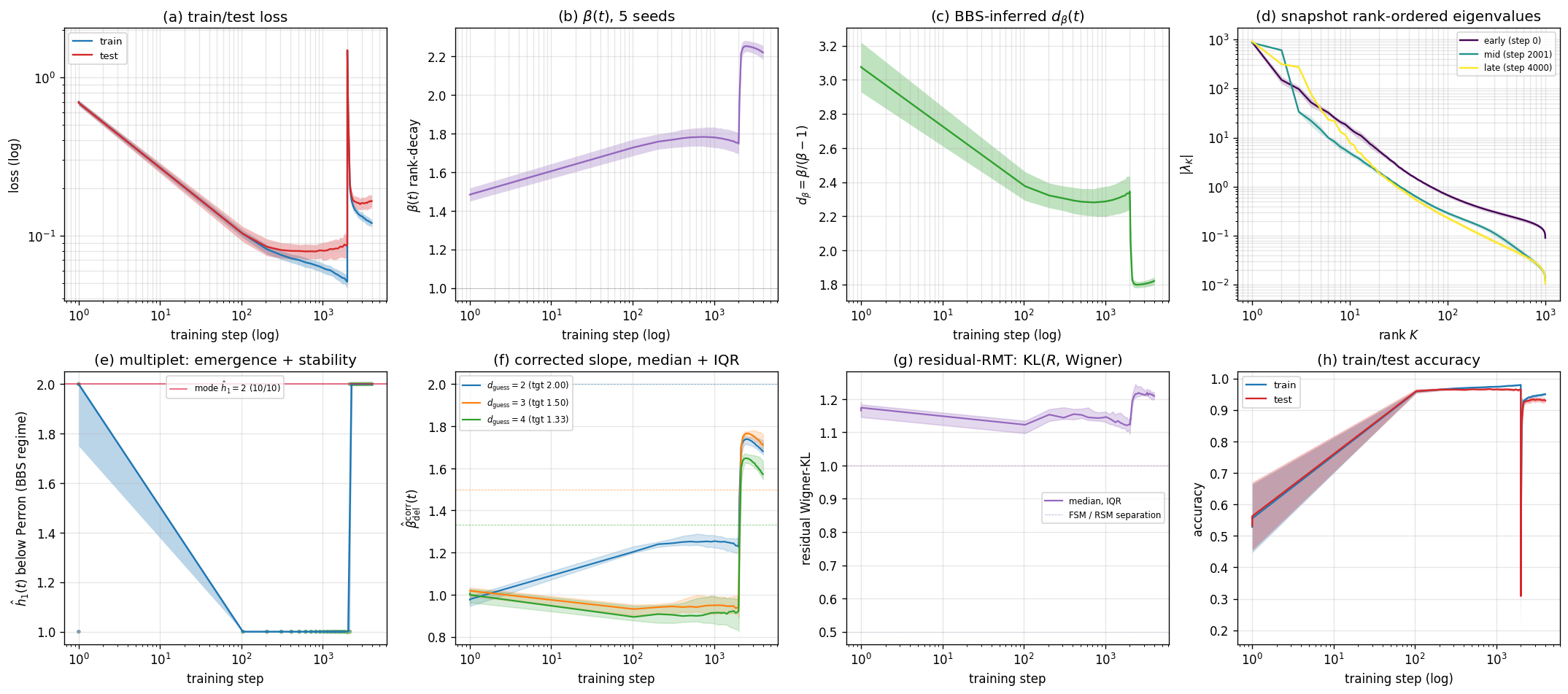}
  \caption{Synthetic task switch from a 1-D to a 2-D
  supervisory signal at step $2000$. Combined
  scalar $+$ I-BBS analysis. Top row: (a)~train/test
  cross-entropy with the switch spike, $10$-seed mean
  $\pm \sigma$;
  (b)~rank-decay $\beta(t)$;
  (c)~BBS-inferred $d_\beta(t)$;
  (d)~rank-ordered eigenvalues at early, switch, and late
  snapshots (mean across seeds), showing the post-switch
  band-structure reorganisation.
  Bottom row: (e)~multiplet $\hat h_1(t)$ emergence and
  stability with per-seed scatter, cross-seed median+IQR,
  and final-checkpoint mode (red horizontal: $\hat h_1 = 2$
  post-switch);
  (f)~corrected delocalised slope median+IQR per
  $d_{\rm guess}$;
  (g)~residual Wigner-KL trajectory;
  (h)~train/test accuracy. Dashed vertical line at step
  $2000$ marks the switch.}
  \label{fig:task-switch}
\end{figure}

AdamW (lr $10^{-3}$, wd $10^{-3}$), batch $64$, $4000$
steps with the switch at step $2000$, $40$ checkpoints
with extra resolution around the switch, $N = 1000$
training-set evaluation sample.
Figure~\ref{fig:task-switch} summarises the result.

\begin{figure}[htbp]
  \centering
  \includegraphics[width=0.55\textwidth]{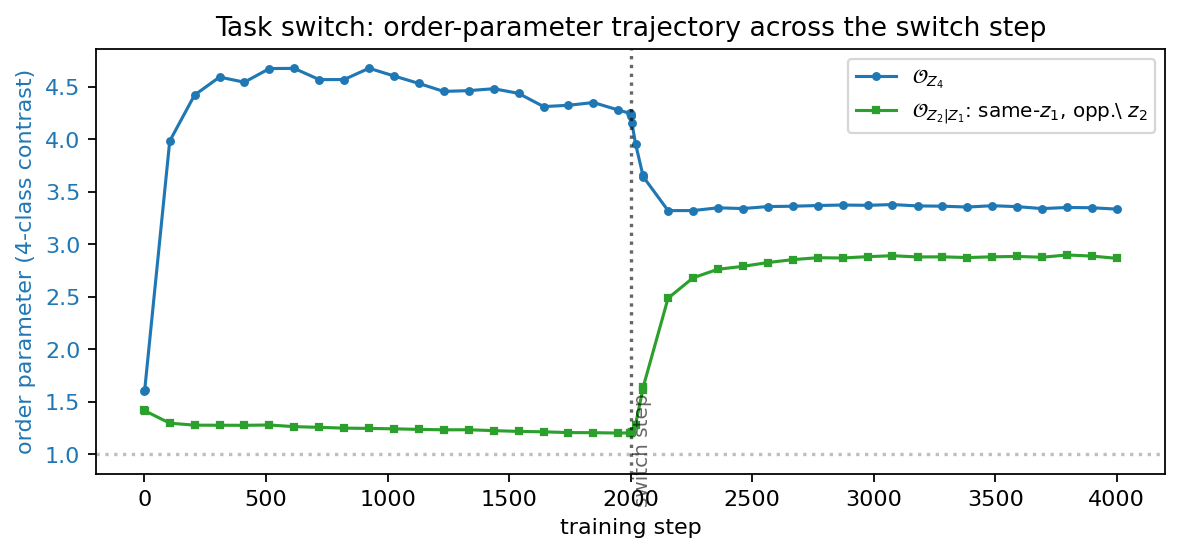}
  \caption{Task switch: order parameters across the switch
  at step $2000$. Green
  ($\mathcal{O}_{\mathbb{Z}_2 | \mathbb{Z}_1}$, same
  $\mathrm{sign}(z_1)$ opposite $\mathrm{sign}(z_2)$ vs
  within-quadrant): symmetric $\approx 1$ in phase A,
  jumps to $\approx 2.85$ in $\sim 100$ steps. Blue
  ($\mathcal{O}_{\mathbb{Z}_4}$, 4-class between/within
  contrast): drops from $\approx 4.15$ to $\approx 3.35$ as
  the cloud reorganises from two half-planes onto a 1-D
  ring of four arcs.}
  \label{fig:task-switch-order}
\end{figure}

The order parameter for the switch,
$\mathcal{O}_{\mathbb{Z}_2 | \mathbb{Z}_1}(t)$, the ratio
of mean angular distance between particles with same
$\mathrm{sign}(z_1)$ and opposite $\mathrm{sign}(z_2)$ to
the within-quadrant distance, is the representation-level
analogue of the sparse-parity
$\mathcal{O}_{\mathbb{Z}_2}$ of
Section~\ref{sec:results-sparseparity}. Phase A's target
depends only on $\mathrm{sign}(z_1)$ so
$\mathcal{O}_{\mathbb{Z}_2 | \mathbb{Z}_1} \approx 1$.
Phase B requires the distinction so
$\mathcal{O}_{\mathbb{Z}_2 | \mathbb{Z}_1} > 1$.
Figure~\ref{fig:task-switch-order} shows
$\mathcal O \approx 1.3$ in phase A jumping to
$\approx 2.85$ in a $\lesssim 100$-step window centred
on the switch, matching the $\beta_{\rm repr}(t)$ step.
The 4-class contrast $\mathcal{O}_{\mathbb{Z}_4}(t)$
drops sharply from $\approx 4.15$ to $\approx 3.35$ as
the cloud reorganises from collapsed phase-A half-planes
onto a 1-D ring. Both order parameters resolve the
transition as a sharply localised step.

The key observation is in panel~(e): $\beta_{\rm repr}(t)$
steps discontinuously between plateaus in a
$\lesssim 100$-step window centred on step $2000$, with
the spectrum of $M(t)$ reorganising sharply. The direction
of motion, however, is opposite to the naive prediction.
A literal BBS reading with $d = d_{\rm task}$ would predict
$\beta = 2$ in phase A ($d_{\rm task} = 1$) and
$\beta = 1.5$ in phase B ($d_{\rm task} = 2$). Observed:
phase A at $\beta \approx 1.5$
($d_\beta^{\rm repr} \approx 2$), phase B at
$\beta \approx 2$ ($d_\beta^{\rm repr} \approx 1$), with
negligible train--test gap. This is the
clustered-classification regime of
Section~\ref{subsec:bbs_inference}: in phase A the smooth
ReLU encoder preserves the 2-D input latent in the
penultimate layer. In phase B the four-way classification
drives the cloud onto four atomic clusters whose
within/between contrast dominates the spectrum, giving
$d_\beta^{\rm repr} \approx 1$. As a detector of
\emph{when} a structural reorganisation happens, the
spectrum is sharp regardless of which direction
$d_\beta$ moves.

The I-BBS Algorithm~1 readout
(Table~\ref{tab:ibbs_summary}) identifies the
post-switch sub-manifold as a $\mathbb{Z}_4$ arc on
$S^1$: pre-switch $\hat h_1 = 1$ (singlet, no band
structure yet on the 4-quadrant target), post-switch
$\hat h_1 = 2$ in $10/10$ seeds at the $N = 1000$,
$10$-seed standardised setup (doublet at the top of the
spectrum after the switch is absorbed). The corrected
delocalised slope selects $\hat d_\beta = 3$ in $8/10$
seeds, with $\hat\beta_{\rm del}^{\rm corr}(d{=}3) =
1.72 \pm 0.04$ vs target $1.50$. The two readings
together describe the four-quadrant post-switch target
arranged on a closed $\mathbb{Z}_4$-equivariant loop.
Residual RMT is RSM-like in $10/10$ seeds
(Wigner-KL $1.51 \pm 0.02$).

\subsection{Input topology change: bifurcation of the
representation cloud}
\label{sec:results-inputtopology}

Setup details in Table~\ref{tab:axes} (row 5.7). The
seventh experiment is the input counterpart of the
task switch: the supervised target (a smooth scalar
function of $z_1$) is held fixed while the input
distribution bifurcates at step $2000$ from a single
isotropic Gaussian $\mathcal{N}(0, \sigma^2 I)$
(phase A) to a balanced two-Gaussian mixture
$\frac{1}{2}\mathcal{N}(+\mu, \sigma^2 I) +
\frac{1}{2}\mathcal{N}(-\mu, \sigma^2 I)$
(phase B), $\mu = e_1$. Same MLP and optimiser as
the task switch but with a regression head under MSE
loss, $4000$ steps. The expected matrix response
is $d_\beta^{\rm repr} \approx 3$ in phase A
(moderate-dimensional cloud reflecting the smooth
regression target) and $\approx 1.5$ in phase B (the
two-cluster input forces a two-cluster representation
whose BBS reading is dominated by the cluster contrast).

\begin{figure}[htbp]
  \centering
  \includegraphics[width=\textwidth]{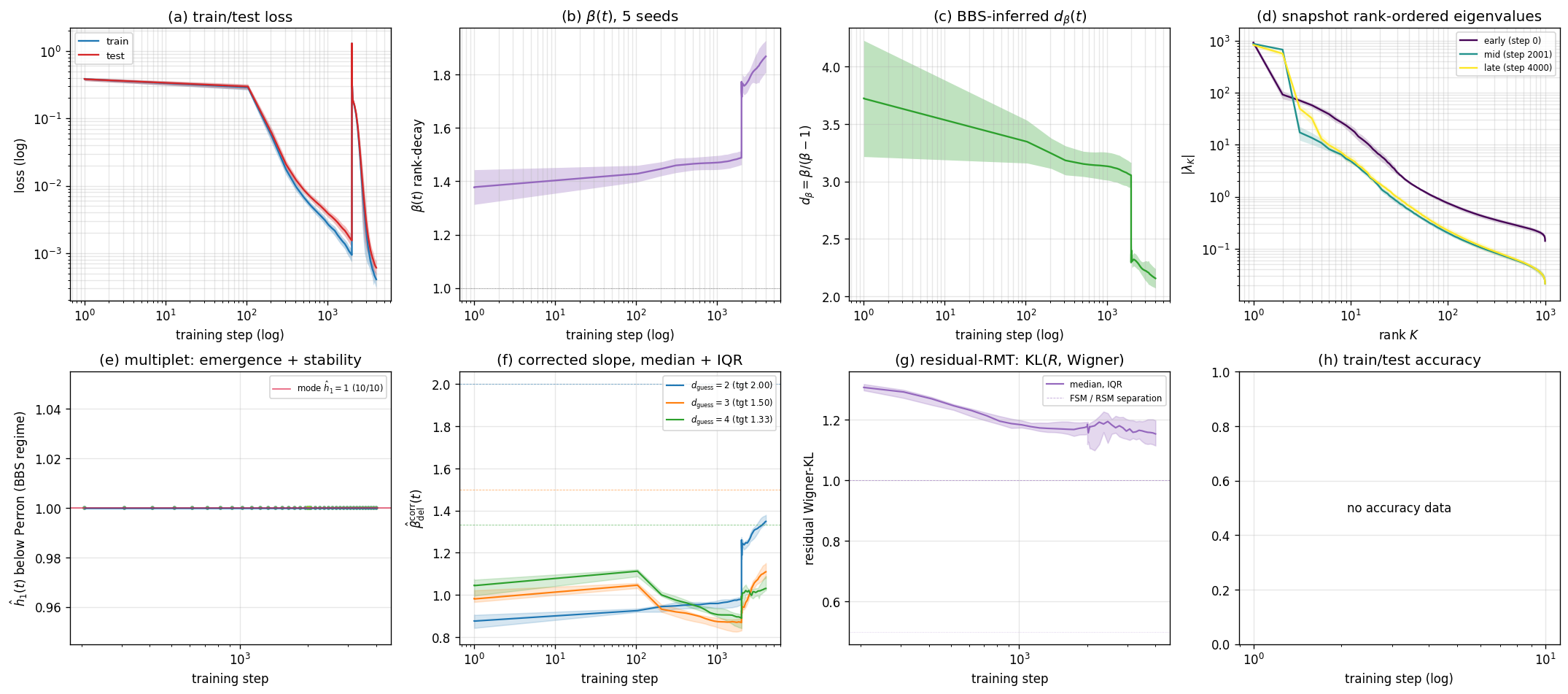}
  \caption{Input topology change at step $2000$:
  single-cluster Gaussian (phase A) $\to$ two-cluster
  mixture (phase B), supervised target unchanged. Combined scalar $+$
  I-BBS analysis. Top row: (a)~train/test MSE
  (semilog-$y$);
  (b)~rank-decay $\beta(t)$;
  (c)~BBS-inferred $d_\beta(t)$;
  (d)~rank-ordered eigenvalues at mid-A / end-A / end-B
  snapshots (mean across seeds), the band-structure view
  of the phase-A $\to$ phase-B reorganisation.
  Bottom row: (e)~multiplet $\hat h_1(t)$ emergence and
  stability with per-seed scatter, cross-seed median+IQR,
  and final-checkpoint mode (red horizontal: stays at the
  singlet floor $\hat h_1 = 1$ throughout);
  (f)~corrected delocalised slope median+IQR per
  $d_{\rm guess}$ (sits at or below the BBS-admissibility
  floor $\beta = 1$);
  (g)~residual Wigner-KL trajectory;
  (h)~regression task (test accuracy is not defined; panel
  shows train/test MSE for completeness). Dashed vertical
  line at step $2000$ marks the bifurcation.}
  \label{fig:input-topology}
\end{figure}

\begin{figure}[t]
  \centering
  \includegraphics[width=0.50\textwidth]{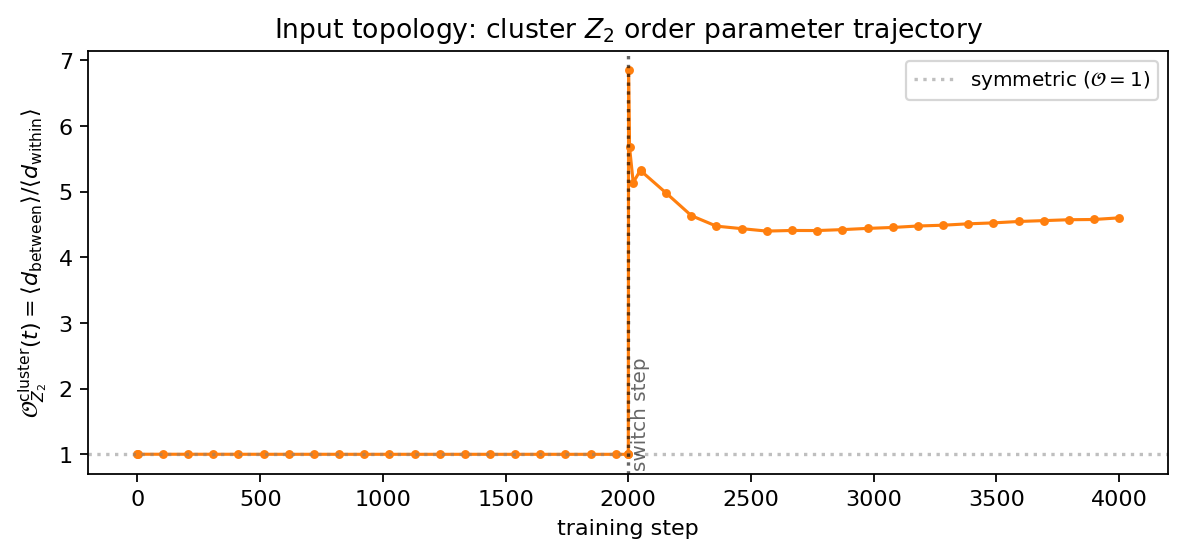}
  \caption{Input-topology: cluster $\mathbb{Z}_2$
  order parameter
  $\mathcal{O}_{\mathbb{Z}_2}^{\rm cluster}(t)$
  (ratio of opposite-cluster to same-cluster mean angular
  distances in $M(t)$) across the switch at step $2000$.
  Symmetric value $\mathcal{O} = 1$ in phase A, peak
  $\sim 6.8$ at the switch and a phase-B plateau of
  $\sim 4.5$.}
  \label{fig:input-topology-order}
\end{figure}

The cluster $\mathbb{Z}_2$ contrast
$\mathcal{O}_{\mathbb{Z}_2}^{\rm cluster}(t) = \langle
M_{ij}\rangle_{\text{opposite}} /
\langle M_{ij}\rangle_{\text{same}}$ is the
representation-level order parameter for this transition,
in the same Landau form as sparse parity and the task
switch. Phase A is symmetric ($\mathcal O \approx 1$).
Phase B forces a $\mathbb{Z}_2$ asymmetry, so
$\mathcal O > 1$. Figure~\ref{fig:input-topology-order}
shows $\mathcal O \approx 0.998$ throughout phase A and
a jump to peak $\sim 6.8$ at the switch with a phase-B
plateau of $\sim 4.5$, developing in fewer than $100$
steps — the same sharp-step signature as the sparse-parity
$\mathcal{O}_{\mathbb{Z}_2}$ of
Figure~\ref{fig:sparse-parity-summary}(e), here with an
external trigger.

Figure~\ref{fig:input-topology} reports the result.
$\beta_{\rm repr}(t)$ in panel~(c) jumps from a phase-A
plateau near $1.4$ to a phase-B plateau near $1.7$, with
$d_\beta^{\rm repr}$ dropping from $\approx 3$ to
$\approx 1.5$. Train and test trajectories coincide
throughout. The PCA scatter in panel~(e) makes the
geometric content transparent: a single connected cloud
in phase A, two well-separated clusters along the top
principal component in phase B, with cluster identity
inherited from the input mixture component. Combined
with task switch and GAN mode collapse, this is the
clustered-classification regime of
Section~\ref{subsec:bbs_inference}: structured
clustering contracts $d_\beta$ irrespective of whether
the trigger is supervisory, generative, or input-driven.

The I-BBS Algorithm~1 readout
(Table~\ref{tab:ibbs_summary}) puts this experiment in
the boundary regime: the multiplet gap walk returns
$\hat h_1 = 1$ in $10/10$ seeds at the $N = 1000$,
$10$-seed standardised setup, both pre- and
post-bifurcation, and the corrected delocalised slope
selects $\hat d_\beta = 4$ in $10/10$ seeds with
$\hat\beta_{\rm del}^{\rm corr}(d{=}3) = 1.11 \pm 0.05$
(below the BBS target $1.50$ at $d_{\rm guess} = 3$),
signalling a near-i.i.d.\ high-dimensional cloud rather
than a low-dimensional attractor. The matrix-valued
diagnostics still resolve the bifurcation event through
the order parameter $\mathcal O(t)$ and the
trajectory-level $D_K(t, t_{\rm ref})$, but the I-BBS
dimension estimates fall outside the smooth-manifold
BBS-admissible range. Residual RMT is RSM-like in
$10/10$ seeds (Wigner-KL $1.56 \pm 0.05$).

\section{Discussion}
\label{sec:discussion}
%==============================================================================

OMD is the dynamic application of the static I-BBS toolkit
\cite{halperin2026IBBS} of
Section~\ref{subsec:observables} to neural network training
trajectories, with the trajectory-level FDM observables
\cite{halperin2026FDM} stacked on the per-snapshot
reads. The Group~B experiments fire the full Algorithm~1
across the transition: the multiplet structure of $M(t)$
reorganises synchronously with the test-loss drop, and
the post-transition $\hat M^{(d)}(t)$ is identified
explicitly. The Group~A experiments lack a clean band
structure, so only the continuous component
$d_\beta(t)$ of Eq.~\eqref{eq:bbs_def} applies. The
matrix observable resolves structural events
invisible to scalar test loss. The trajectory-level
FDM observables ($D_K, C, \Delta_\beta$ and the
level-spacing time evolution) sit on the
per-snapshot analysis in both groups, detecting coherent
bottom-eigenspace dynamics no per-snapshot
diagnostic sees.

A comment is due here regarding what is actually detected by OMD and I-BBS. The per-experiment readouts of
Section~\ref{subsec:results-cross-experiment} establish
the I-BBS toolkit as a two-sided regime classifier on
the spectrum of $M(t)$ (consolidated in
Table~\ref{tab:ibbs_summary}): every Group~B sharp
transition returns a seed-consistent post-event integer
multiplet, while the diffusive Group~A experiments and
the pre-transition windows of all four Group~B
experiments return none. What this
detects is stable spectral fingerprints, not automatic
validation of a smooth-sub-manifold BBS picture for
every Group~B case: a finite $\mathbb Z_2$ two-cluster, a
$\mathbb Z_k$-vertex configuration, or a finite
Fourier-soliton on $S^1$ each produces its own stable
integer fingerprint not matching $h(1, d) = d$
for any smooth $S^{d-1}$. The regime classification
(Table~\ref{tab:regimes}: smooth / product /
Fourier-soliton / atomic-cluster / diffusive) is the
right axis for reading $\hat h_1$. Robustness rests
on three independent calibrations: cross-seed
agreement on the final-checkpoint integer; the
negative-eigenvalue verdict calibrated by Group~A and
pre-transition windows; and the residual-spectrum RMT
step with an RSM verdict in every seed and
experiment (\mbox{Wigner-KL $\sim 1.5$--$2.3$}).

Every Group~B transition produces
a low-dimensional geometric object out of
an otherwise featureless high-dimensional cloud. The
post-event objects (Table~\ref{tab:ibbs_summary},
column~2) are either smooth low-dimensional
sub-manifolds or discrete sets that sit on, or can be
mapped onto, a smooth low-dimensional manifold: each
post-event $\hat h_1$ counts the number of discrete
vertices or Fourier pairs populating the leading band,
equal to the number of irreducible
$\mathbb Z_k$- or $\mathbb Z_p$-equivariant directions
on the post-event geometry. The Group~B label of
Section~\ref{sec:experiments} is therefore best read as
\emph{geometric phase transition}, with I-BBS returning
the integer fingerprint of the structure under
cross-seed consistency. The 8-Gaussian GAN sits
adjacent to this set as a smooth cluster-coverage
trajectory rather than a sharp transition: mode
coverage grows continuously across training, and no
Group~A run (GAN included) forms the spectral gap that
the gap walk detects as the transition step on every
Group~B run.

The spectrum of $M(t)$ reads the task's symmetries at
different layers: \emph{output} symmetry at the
downstream/post-LayerNorm representation or logits, and
\emph{input} symmetry at the upstream embedding layer
through the product-of-spheres decomposition
(Section~\ref{subsec:observables}). The
modular-arithmetic transformer exercises both: downstream
returns the output symmetry $\mathbb Z_p \cong S^1$ via
the Fourier-soliton on $S^1(a + b)$. Upstream returns the
input symmetry $\mathbb Z_p \times \mathbb Z_p$ via the
$T^2 = S^1 \times S^1$ factorisation. The other Group~B
experiments are read only on the downstream side, which
carries the output symmetry of Table~\ref{tab:axes}. The
post-event geometries of Table~\ref{tab:ibbs_summary}
column~2 are therefore the symmetry footprints the
spectrum reads at the chosen layer. Extending the
upstream product-of-spheres readout to the four
non-grokking Group~B experiments is a clean follow-up
direction.

\begin{figure}[htbp]
\centering
\includegraphics[width=\textwidth]{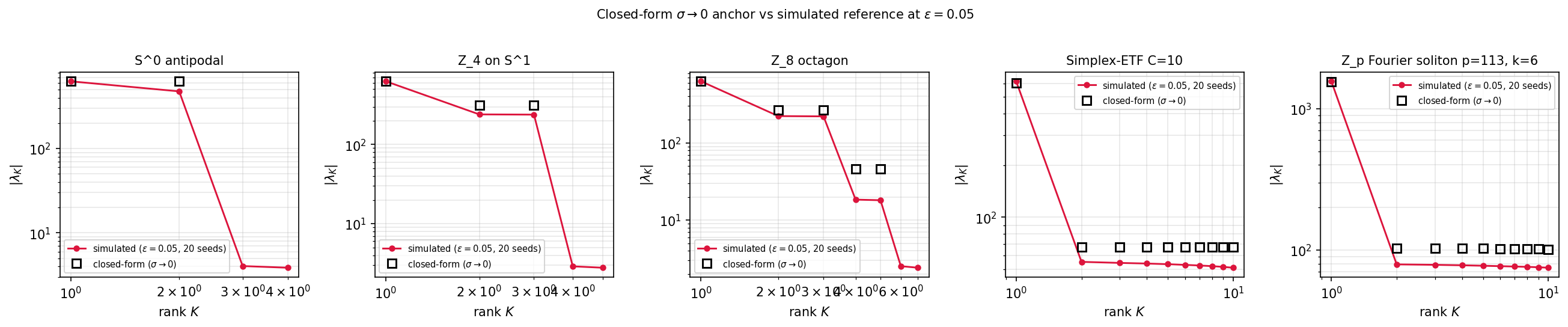}
\caption{Closed-form $\sigma \to 0$ leading
eigenvalues (black open squares) overlaid on the
simulated reference at $\epsilon = 0.05$ (red, $20$
seeds) for the five candidate post-event geometries
used in Figures~\ref{fig:groupB-spectrum-vs-theory}
and \ref{fig:grokking-three-layer-spectrum-vs-theory}.
The simulated reference uses the finite-$\epsilon$
signal $M^{(\sigma)}$ with the within-blob noise
inflation; closed form uses the $\sigma \to 0$ block.
The Perron matches; non-Perron eigenvalues are
$\sim 20\%$ smaller in the simulation, accounted for
by the within-blob noise inflation: the within-blob
arc distance grows from $0$ to $\sim \epsilon
\sqrt{2(D-1)}$, shifting the antipodal $S^0$ block from
$\theta_2^{(0)} = -\pi N/2$ to $\theta_2^{(\sigma)}
\approx -\pi N/2 + \epsilon N\sqrt{2(D-1)}$ (text and
Appendix~\ref{app:antipodal_blobs}).}
\label{fig:closed-form-vs-simulated}
\end{figure}

\begin{figure}[htbp]
\centering
\includegraphics[width=\textwidth]{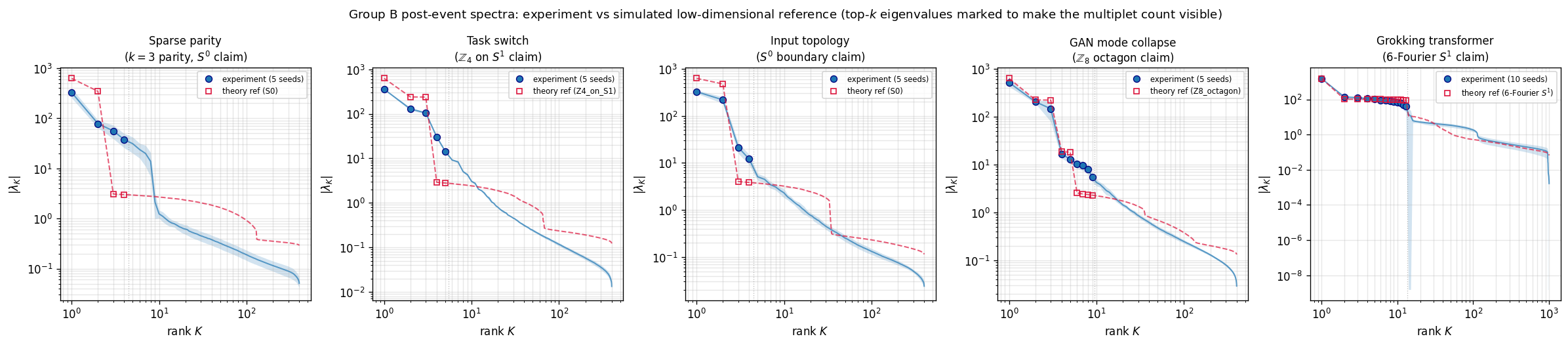}
\caption{Final-checkpoint rank-ordered eigenvalue spectra
of $M(t)$ for the four Group~B sharp-transition
experiments (sparse parity, input topology, task switch,
grokking), plus the 8-Gaussian GAN (Group~A
trajectory-wise, shown for spectral comparison). Blue:
experiment, mean $\pm 1\sigma$ across seeds. Red dashed:
simulated reference of the claimed post-event geometry,
median $+$ IQR over $20$ seeds at RSM noise
$\epsilon = 0.05$ matched to the trained network, validated
against the closed-form $\sigma \to 0$ anchor
(Table~\ref{tab:closed-form-vs-sim},
Figure~\ref{fig:closed-form-vs-simulated}). The leading
$\hat h_1 + 1$ eigenvalues are marked individually (filled
blue circles, open red squares), and the dotted vertical
line is the multiplet boundary $k + 1$ ($4$ for $S^0$, $5$
for $\mathbb Z_4$ on $S^1$, $9$ for the GAN $\mathbb Z_8$
octagon, $13$ for the $6$-pair grokking soliton).
Reference geometries: $S^0$ (sparse parity, input
topology), $\mathbb Z_4$ on $S^1$ (task switch),
$\mathbb Z_8$ octagon (GAN), and the $\mathbb Z_p$-Fourier
soliton on $S^1$ ($p = 113$, $6$ pairs) for grokking. The
leading $\hat h_1$ eigenvalues sit at the reference
amplitude in all five cases. The Group~B rows are matrix
support for the geometric-phase-transition framing of
Section~\ref{sec:discussion}. The GAN row is a static
geometric match of the post-event $\mathbb Z_8$ scaffold
within Group~A.}
\label{fig:groupB-spectrum-vs-theory}
\end{figure}

\begin{figure}[htbp]
\centering
\includegraphics[width=\textwidth]{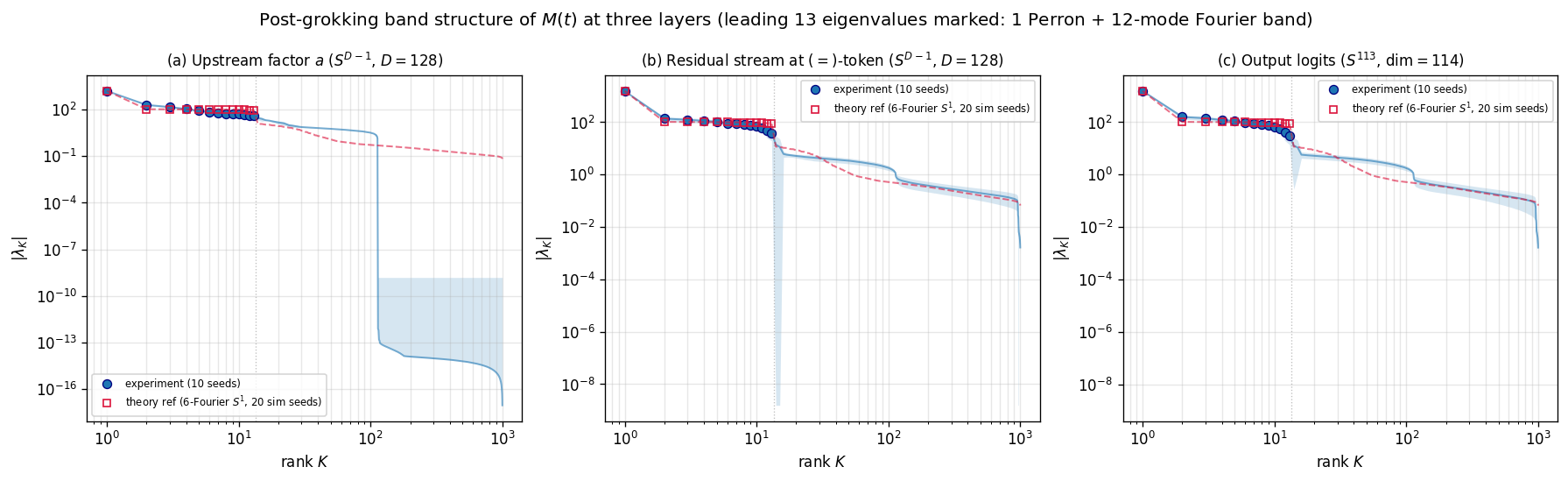}
\caption{Post-grokking band structure of $M(t)$ at the
three layers of the modular-arithmetic transformer
(blue: experiment, mean $\pm \sigma$ over $10$ trained
seeds at the final checkpoint; red dashed: simulated
reference, median $+$ IQR over $20$ seeds of the
6-Fourier-mode soliton on $S^1$, $p = 113$, $N = 1000$,
RSM noise $\epsilon = 0.05$). The leading $13$
eigenvalues (Perron plus six Fourier mode pairs) are
marked individually; the dotted line at $K = 13.5$ marks
the expected band boundary.
(a)~Upstream embedding-layer factor $a$ at token
position $0$ ($S^{127}$): leading $\sim 12$ eigenvalues
match the theory band, with a sharp rank-degeneracy drop
at $K \approx 113$ (each token appears $\sim 9$ times in
the $1000$-pair eval set).
(b)~Residual stream at the $(\,=\,)$-token post final
LayerNorm ($S^{127}$): theory band tracks the leading
$\sim 12$ eigenvalues, followed by a smoother bulk.
(c)~Output logits ($S^{113}$): the cleanest match of the
three layers in both band and bulk-decay shape. All
three layers share the same six-Fourier-pair fingerprint,
supporting the $T^2 \to S^1$ reading of the
modular-addition operation.}
\label{fig:grokking-three-layer-spectrum-vs-theory}
\end{figure}

\begin{table}[h]
\centering
\small
\begin{tabularx}{\linewidth}{@{}>{\raggedright\arraybackslash}p{0.27\linewidth} X X l@{}}
\toprule
Experiment & $\hat{\mathcal M}_d$ identified? &
$\hat d$ / multiplet & Noise: res\,/\,$\ell_2$ \\
\midrule
MNIST + MLP, moderate WD &
no (diffusive) &
$\hat h_1$ not identified (10/10);
$\beta_{\rm train}: 0.93 \to 1.29$
& RSM\,/\,$\ell_2$\,n/a \\
Multi-output regression &
no (diffusive) &
$\hat h_1 = 1$ (10/10);
$\hat\beta_{\rm corr}(d{=}2,3,4) \le 1$
& RSM\,/\,$\ell_2$\,n/a \\
GAN mode collapse &
smooth cluster-coverage (no sharp transition) &
$\hat h_1 = 2$ (10/10);
$\hat d_\beta = 3$ (10/10) & RSM\,/\,FSM\,(4/10) \\
\midrule
Mod.-arith.\ transformer (downstream) &
yes, Fourier-soliton on $S^1(a + b)$ (ambient $S^{127}$) &
$\hat h_1 = 12$, $\hat d_\beta = 3$ & RSM\,/\,FSM\,(9/10) \\
Mod.-arith.\ transformer (upstream) &
yes, $T^2 = S^1(a) \times S^1(b)$ &
$\hat h_1^{(a)} = \hat h_1^{(b)} = 12$ per factor & RSM\,/\,FSM\,($a$8,$b$9) \\
Mod.-arith.\ transformer (output logits) &
yes, Fourier-soliton on $S^1(a + b)$ (ambient $S^{113}$) &
$\hat h_1 = 12$, $\hat d_\beta = 3$ ($9/10$) & RSM\,/\,FSM\,(9/10) \\
Sparse parity &
yes, $8$-vertex atomic configuration on $\mathbb Z_2^k$ &
$\hat h_1 = 7$ (mode, $5/10$), $\hat d_\beta = 2$
($10/10$) & RSM\,/\,FSM\,(8/10) \\
Synthetic task switch &
yes, $\mathbb{Z}_4$ arc on $S^1$ &
$\hat h_1 = 2$ ($10/10$), $\hat d_\beta = 3$ ($8/10$)
& RSM\,/\,FSM\,(10/10) \\
Input-topology bifurcation &
boundary case (no clean band) &
$\hat h_1 = 1$ ($10/10$), $\hat d_\beta = 4$ ($10/10$)
& RSM\,/\,$\ell_2$\,n/a \\
\bottomrule
\end{tabularx}
\caption{Cross-experiment I-BBS Algorithm~1 readouts.
Columns: whether the I-BBS multiplet diagnostic identifies
a low-dimensional latent sub-manifold $\hat{\mathcal M}_d$;
the latent geometry suggested when identified (with the
multiplet multiplicity $\hat h_1$ and slope-based dimension
$\hat d_\beta$); and the noise model read two ways, as the
residual-RMT bulk shape (RSM\,$=$\,peaked, non-Wigner) and
the blind $\ell=2$ component ($\ell_2$\,n/a where the
manifold is a singlet or diffuse), seeds agreeing in
parentheses. The first four
rows are Group~A (no sharp transition): three
diffusive-relaxation runs in which only the continuous
slope diagnostic fires, plus the 8-Gaussian GAN whose
stable doublet indexes a smooth mode-coverage trajectory
rather than a sharp transition. The remaining six rows
below the midrule are Group~B (band structure on $M(t)$
forming at a sharp transition step), counting the
modular-arithmetic transformer's three layers (downstream
residual stream, upstream embedding factors, output
logits) as three separate readouts of the same
experiment.}
\label{tab:ibbs_summary}
\end{table}
On the scope of BBS theory versus simulation-based
references, BBS theory \cite{bogomolny2003, bogomolny2007} is a
continuum statement on smooth $d$-manifolds, and
I-BBS \cite{halperin2026IBBS} stays inside this picture
through the perturbation
$M^{(D)} = M^{(d)} + \epsilon M^{(\mathrm{noise})} + O(\epsilon^2)$. The Group~B
post-event geometries split by their distance from this
continuum picture, with the I-BBS slope-admissibility
floor as the operational test
(Table~\ref{tab:regimes}). The Fourier-soliton regime
(modular-arithmetic transformer) is BBS-adjacent: a
discrete $p$-point subset of $S^1$ whose
arccos-distance matrix inherits the underlying Fourier
basis, with the slope target $d/(d-1) = 1.5$ matched to
$0.5\%$ by the output logits and $\sim 5\%$ by the
residual stream. Upstream factors sit at $\hat\beta
\approx 0.93$ at the admissibility boundary, consistent
with each factor being a discrete $p$-point sample. The
atomic-cluster regime (sparse parity, GAN, task switch,
input topology) is \emph{outside} BBS: post-event objects
are finite $\mathbb Z_k$-equivariant discrete sets with
rank $\le k - 1$, $\hat h_1$ still counts the
$\mathbb Z_k$-equivariant directions (matching the
output symmetries of Table~\ref{tab:axes}) but is no
longer the $d$ of a continuous multiplet, and the
slope flags the regime by overshooting or undershooting
the BBS target. The reference spectrum used in the figure overlays is
built from each candidate geometry $G$ by $N$ noisy
samples ($N/k$ copies per vertex for finite-vertex
$G$, RSM noise of scale $\epsilon$, $L^2$-normalised
onto $S^{D-1}$, $\sim 20$ seeds), with the rank-ordered
spectrum of $M^{\rm ref} = \arccos(\hat h\,\hat h^\top)$
as the comparison curve; full setup in
Appendices~\ref{app:calibration} (smooth) and
\ref{app:finite-cluster-controls} (finite). Several
relevant $G$ admit a closed-form spectrum in the
$\sigma \to 0$ limit, used here as the asymptotic
anchor: the smooth $S^{d-1}$, $T^d$, product-sphere and
$\mathbb{RP}^{d-1}$ families (BBS
\cite{bogomolny2003, bogomolny2007} and I-BBS
\cite{halperin2026IBBS}); circulant configurations
such as $\mathbb Z_k$ equispaced on $S^1$ via character
/ circulant analysis \cite{gray2006circulant}; the
simplex-ETF \cite{papyan2020neural, sustik2007etf};
and the two-antipodal-Gaussian-blob case derived in
Appendix~\ref{app:antipodal_blobs}.
Table~\ref{tab:closed-form-vs-sim} compares the closed
form against the simulated reference for the five
geometries used in
Figures~\ref{fig:groupB-spectrum-vs-theory} and
\ref{fig:grokking-three-layer-spectrum-vs-theory},
with the visual overlay in
Figure~\ref{fig:closed-form-vs-simulated}. The Perron
matches to $<2\%$ in every case. Non-Perron eigenvalues
are systematically $\sim 20\%$ smaller in the
simulation, the \emph{finite-$\epsilon$ correction to
the signal structure itself}. Writing
$M = M^{(\sigma)} + R_\epsilon$ with $M^{(\sigma)} :=
\mathbb E[M]$ the $\epsilon$-deformed signal, the
within-blob arc distance inflates from $0$ to
$\sim \epsilon\sqrt{2(D-1)}$ and the cross-blob arc
distance shifts accordingly, so the rank-$k$ block
$\mathbf D^{(\sigma)}$ has different entries than the
$\sigma \to 0$ block $\mathbf D^{(0)}$. For the
antipodal $S^0$ case the second eigenvalue moves
from $\theta_2^{(0)} = -\pi N/2$ to
$\theta_2^{(\sigma)} \approx -\pi N/2 + \epsilon N
\sqrt{2(D-1)}$, accounting for $\sim 157$ of the
$\sim 151$ observed shift at $N{=}400$, $D{=}32$,
$\epsilon{=}0.05$ (full derivation, including
$\mathbb Z_k$ on $S^1$ and Fourier-soliton analogues,
in Appendix~\ref{app:antipodal_blobs}).
The signal correction is the only $\epsilon$-driven
shift visible on the spectrum at the OMD operating
point: the additional random-matrix correction from
the signal/bulk interaction is of order
$(k\epsilon)^2/N \sim 10^{-5}$ relative
(signal-to-bulk ratio $\sqrt N/(k\epsilon) \sim
50$--$300$, deep above the bulk edge), and the
empirical top-$(1 + \hat h_1)$ eigenvectors recover the
signal eigenvectors with overlap $1 - O(10^{-4})$,
justifying the I-BBS truncated reconstruction
$\hat M^{(d)}$ (Davis--Kahan
\cite{DavisKahan1970, halperin2026IBBS} gives the same
guarantee in our regime).
The diffusive-relaxation regime (Group~A) is outside
BBS by a different mechanism: no band structure, no
stable multiplet, slope below the admissibility floor.

\begin{table}[htbp]
\centering
\footnotesize
\setlength{\tabcolsep}{4pt}
\begin{tabular}{l c c c c c c c}
\toprule
Geometry & $K$ & $|\lambda_1|$ cf & $|\lambda_1|$ sim &
$|\lambda_2|$ cf & $|\lambda_2|$ sim &
$|\lambda_3|$ cf & $|\lambda_3|$ sim \\
\midrule
$S^0$ (sparse parity / input topology) &
2 & $628.3$ & $627.9$ & $628.3$ & $477.2$ & --- & --- \\
$\mathbb Z_4$ on $S^1$ (task switch) &
3 & $628.3$ & $627.9$ & $314.2$ & $240.0$ & $314.2$ & $238.4$ \\
$\mathbb Z_8$ octagon (GAN) &
5 & $628.3$ & $628.0$ & $268.2$ & $222.6$ & $268.2$ & $220.8$ \\
Simplex-ETF $C = 10$ &
10 & $605.6$ & $617.5$ & $67.3$ & $55.0$ & $67.3$ & $54.3$ \\
$\mathbb Z_p$ Fourier soliton, $p{=}113$, $k{=}6$ &
13 & $1551.7$ & $1563.3$ & $102.7$ & $79.0$ & $102.7$ & $78.4$ \\
\bottomrule
\end{tabular}
\caption{Leading $|\lambda_K|$ for each candidate
post-event geometry: closed-form $\sigma \to 0$ anchor
vs the simulated reference at $\epsilon = 0.05$
($N = 400$, except $N = 1000$ for the grokking
soliton). The $K$ column gives the number of non-zero
closed-form eigenvalues. Perron $|\lambda_1|$ matches
to $<2\%$ in every case; non-Perron eigenvalues are
systematically $\sim 20\%$ smaller in the simulation,
the finite-$\epsilon$ correction to the rank-$k$
signal block $\mathbf D^{(\sigma)}$ relative to its
$\sigma \to 0$ limit $\mathbf D^{(0)}$
(text and Appendix~\ref{app:antipodal_blobs}). The
closed-form $\sigma \to 0$ values validate the symmetry
pattern; the simulated reference matches the trained-network
spectrum at the same operational $\epsilon$.}
\label{tab:closed-form-vs-sim}
\end{table}

The upstream product-of-spheres readout supplies the explicit
two-circle decomposition. The grokking transformer's three-layer readout
(Table~\ref{tab:grokking_layers}) gives the explicit
$T^2 \to S^1$ matrix geometry of the
modular-addition operation: the upstream embedding-layer
factors return $\hat h_1^{(a)} = \hat h_1^{(b)} = 12$
(the same $6$-Fourier-mode soliton structure the
downstream readout finds on $S^1(a + b)$), the
factorisation commutator $\|[N_a, N_b]\|_F /
(\|N_a\|\,\|N_b\|)$ contracts $\sim 2.5\times$ across
training (from $\sim 0.12$ pre-training to $0.045 \pm
0.011$ post), and the joint noise fraction of
Eq.~\eqref{eq:N-decomp} is $\eta = 0.06 \pm 0.03$ at the
final checkpoint. The downstream multi-Fourier-mode
soliton on $S^1(a + b)$ together with the upstream
two-circle decomposition therefore identifies
$T^2 \to S^1$ on the matrix side, in line with the
Fourier-feature mechanism of \cite{nanda2023progress}.

In comparison with existing diagnostics, the matrix-valued observable provides a finer reading
than scalar diagnostics across the experimental
programme. Translated into $M(t)$, the neural-collapse
diagnostic of \cite{papyan2020neural} appears as a
rank drop of the centred angular distance matrix to
$C - 1$ with $d_\beta$ contracting toward the
inter-class dimension. The MNIST + MLP
heavy-weight-decay run in the online appendix reaches the
NC1 within-class condensation stage of this prediction
($6\times$ scatter reduction) without reaching the
simplex-ETF geometry (no $K = C = 10$ plateau in the
class-mean Gram, $\beta_{\rm repr} \approx 1.21$). The grokking
diagnostics of
\cite{nanda2023progress,thilak2022slingshot,kumar2024grokking}
identify a mechanism (progress measures, slingshot
dynamics, delayed loss decrease) and locate the
transition in time. $M^{\rm train}(t)$ adds a
complementary signal whose spectrum is flat over the
memorisation plateau and reorganises sharply at the
grokking step, with $\Delta_\beta(t)$ peaking at the
generalisation moment. For matched-space regression the
output-level matrix separates two regressors at the same
test MSE through their $d_\beta^{\rm loss}$. In the
lazy-versus-feature-learning dichotomy
$\beta_{\rm repr}(t)$ is essentially flat in the lazy
regime and only traces nontrivial trajectories in the
feature regime, while $\beta_{\rm loss}(t)$ is visible
in both. The GAN mode-collapse experiment extends the
framework to generative training: the joint scatter of
$d_\beta^{\rm gen}$ against ground-truth mode-coverage
establishes that increased cluster organisation
contracts the matrix-valued observable in the same
direction as supervised collapse, so the framework
treats supervised and generative training as instances
of the same geometric flow. The task-switch and
sparse-parity results further confirm that
$d_\beta^{\rm repr}$ tracks the post-transition
\emph{cluster topology} of the supervised target rather
than the abstract supervisory-signal dimension:
increasing the target cluster count drives
$d_\beta^{\rm repr}$ down, in line with the
classification-induced collapse picture.

The symmetry-restoration view comes with explicit order parameters. The geometric-phase-transition framing admits a uniform
symmetry-restoration reading. Pre-transition, the task symmetry
$G$ acts trivially on the representation cloud
($G$-related inputs are indistinguishable from the
network's view). Post-transition, $G$ acts non-trivially.
In terms of Landau's theory of phase transitions
\cite{landau1980statistical}, this is the passage to the
``broken'' phase, with a spectral order parameter
$\mathcal O(t)$ playing the role of magnetisation.
The post-event geometries of
Table~\ref{tab:ibbs_summary} are in each case the
smallest faithful Riemannian realisation of the output
group: $S^1$ for cyclic groups $\mathbb Z/n$, $S^0$ for
$\mathbb Z_2$, $\mathbb Z_8$-equivariant octagon for the
GAN. The grokking ``bagel''
\cite{halperin2026grokking} is the explicit
$T^2 \times S^1$ realisation, with the trajectory
diagnostics $D_K(t, t')$ and $C(t, t')$ of
Section~\ref{subsec:observables} detecting its
post-transition Goldstone-mode rotation. The explicit
order parameter in each phase-transition experiment is a
simple functional of $M(t)$: a between/within angular
contrast on the supervisory labels (e.g.\
$\mathcal O_{\mathbb Z_2}(t) = \langle d_{\rm between}
\rangle / \langle d_{\rm within}\rangle$ in
Figure~\ref{fig:sparse-parity-summary}(e)) or the
between-cluster separation on the leading bottom-eigenvector
for the input-topology boundary case, departing from its
symmetric value over the same step window as
$\beta_{\rm repr}(t)$. The endogenous (grokking, sparse
parity) and externally driven (task switch, input
topology) cases are indistinguishable in post-transition
geometry. Only the trigger differs. The externally
driven cases read as \emph{driven non-equilibrium phase
transitions} in which the exogenous step change in the
data or target distribution plays the role of a
control-parameter quench. The diffusive Group~A
experiments fit the same picture in a weaker sense:
they realise the class-permutation symmetry $S_C$ by
contracting continuously toward a $(C{-}1)$-simplex-ETF
target rather than through a sharp step.

Turning to loss symmetries and equivariant gradient flow, the condition on the loss $L(\theta)$ that guarantees an
internal representation equivariant under
$G_{\rm in} \times G_{\rm out}$ is the standard
equivariant-learning one translated to the matrix-valued
setting: loss invariance under an induced action
$\Phi_g$ on parameter space implies equivariant gradients
and noise covariance, so the SDE \eqref{eq:M_sde} has
drift and diffusion commuting with the data-symmetry
action and the spectral multiplet structure of $M(t)$
matches the irreducible representations of
$G_{\rm in} \times G_{\rm out}$
(Appendix~\ref{app:loss-symmetries} for the explicit
condition).

The symmetry-restoration view suggests two further
application classes, at the technique and representation-layer levels. As a \emph{technique}, the
matrix-valued observable provides a stopping criterion
and a monitoring signal during training: a network can
be trained until $M(t)$ reaches a target symmetric form,
or until $\Delta_\beta(t)$ stabilises at small values
indicating train--test geometric agreement. This is a
representation-quality monitor that complements the
scalar loss and is sensitive to over- and under-collapsed
geometries that the scalar metric cannot resolve. As a
\emph{representation layer}, the penultimate features of
a network trained with a matrix-valued objective (online
appendix) inherit the geometric structure of the imposed
target, with the cosine similarity between any two
samples directly encoding class membership. Such
features are natural inputs to downstream tasks where
the geometric structure is the operative quantity
(nearest-neighbour classification, similarity search and
retrieval, clustering with known cluster count, transfer
learning).

Three limitations of the framework deserve mention.
First, the symmetric distance matrix $M(t)$ discards
directionality on the sphere. Only the pairwise angles
are kept, and sign / orientation information is not
recovered without supplementary observables on the
embedded representations themselves. Second, the BBS
dimension and rank-decay exponent are scalar
contractions of the full spectrum, asymptotic
diagnostics for the bottom-eigenvalue region rather
than direct intrinsic-dimension estimators in the sense
of \cite{facco2017twonn}. The multiplet diagnostic
sharpens this reading when a clean band structure is
present (Group~B), but in the diffusive-relaxation
regime (Group~A) only the continuous slope component
fires and $d_\beta$ falls outside the BBS-admissible
range. Third, the literal product-of-spheres
Kronecker-sum identity of \cite{halperin2026IBBS} for
factorised samples is approached, not exactly realised,
on the modular-addition transformer's $1000$-pair eval
set: each token in $\mathbb Z_p$ appears on average
$\sim 9$ times at each position, so the per-factor
$N$-matrices are rank-degenerate by construction.
Appendix~\ref{app:token-dedup} verifies that
deduplicating to the $113$ unique per-position
embeddings preserves the readouts, so the upstream
$T^2$ verdict is a property of the per-factor spectrum
rather than of the random $(a, b)$ sampling.

%==============================================================================
\section{Summary and outlook}
\label{sec:summary}
%==============================================================================

To recap the framework, we have introduced \emph{Observable Matrix Dynamics}
(OMD), a matrix-valued observable formulation of neural
training in which a fixed evaluation sample of $N$ inputs
furnishes the particles of an arccos distance matrix
$M(t)$. The $N$-subset
plays the role of quenched disorder. The
parameter-space Langevin dynamics \eqref{eq:weight_sde}
induces the matrix-valued It\^o SDE
\eqref{eq:M_sde}, and the pair-counting argument of
Section~\ref{subsec:observables} forces structural
correlations into $M(t)$ whenever $d < (N-1)/2$. The
central organising claim is that the symmetries of the
learned representation must match those of the
supervised task for train--test i.i.d.\ to be preserved
at the level of $M(t)$, with the spectrum of $M(t)$ as
the geometric test.
\par
$M(t)$ is exercised in two ways: (i) the per-snapshot
I-BBS diagnostic toolkit (multiplet multiplicity,
corrected delocalised slope, localised slope,
residual-RMT consistency), stacked with the
trajectory-level FDM observables (level statistics,
projector drift, commutator norm, train-test gap
$\Delta_\beta$); and (ii) the 3D MDS-plus-$r_\perp$
visualisation that renders $M(t)$ as a moving particle
cloud and gives a quantitative geometric description of
the phase transitions (grokking and sparse parity as
endogenous; task-switch and input-topology as
externally driven).

The I-BBS toolkit revealed the following. Across the seven experiments at $5$--$10$ seeds each
(Table~\ref{tab:ibbs_summary}), I-BBS Algorithm~1
operates as a two-sided diagnostic: the four Group~B
sharp transitions return seed-consistent integer
multiplets post-event ($\hat h_1 \in \{1, 2, 12\}$),
while the two diffusive Group~A experiments and all
pre-transition windows return no stable multiplet. The
8-Gaussian GAN sits on the Group~A side as a smooth
mode-coverage trajectory: it has a stable $\hat h_1 = 2$
fingerprint but no sharp transition step in $\beta(t)$
or $d_\beta(t)$. The Group~B transition is signalled
directly by the formation of a spectral gap separating
the leading delocalised states from the bulk. No
Group~A run, GAN included, forms such a gap. Post-event geometries split into
three regimes (Fourier-soliton, atomic-cluster,
intermediate doublet). The Group~B label is best read as
\emph{geometric phase transitions}, with the upstream
grokking embedding-layer matrix factorising via
Eq.~\eqref{eq:N-decomp} into two six-Fourier-mode
solitons on $T^2 = S^1 \times S^1$, in agreement with
the downstream answer-circle readout. Residual-RMT
verdict: \emph{RSM-like in every seed of every
experiment} (Wigner-KL $\sim 1.5$--$2.3$, with
$1.51$--$1.56$ on the three Group~B sharp transitions
at $N = 1000$, $10$ seeds), the I-BBS-predicted noise
class for post-LayerNorm activations driven by a
Gaussian-additive process.

Symmetry restoration emerges as the unifying view. Read alongside the companion grokking analysis
\cite{halperin2026grokking}, which interprets
delayed generalisation as a symmetry-restoring phase
transition (Section~\ref{sec:discussion}), the
experiments support \emph{a unifying view of training as the
gradient flow searching for an internal representation
that respects the task's symmetries}. $M(t)$ reads the
result through three readouts: the topology of the
bottom-eigenspace MDS embedding, the multiplet structure
of the BBS template, and the synchronisation of
$M^{\rm train}(t)$ and $M^{\rm test}(t)$. The framework
is an observational counterpart to the constructive
symmetry-realisation routes of equivariant
architectures, contrastive and self-supervised learning,
and data augmentation.

As an outlook, several directions follow from the framework laid out
above, organised along the methodology, training, and
deployment axes. On the methodological side,
OMD relates naturally to Geometric Deep Learning
\cite{bronstein2021geometric} and to Joint Embedding
Predictive Architectures (JEPA): the former encodes
symmetries into the architecture, while OMD reads off
whichever symmetries the gradient flow has actually
realised. The latter learns predictive structure in an
embedding space whose matrix observable is exactly
$M(t)$. On the training side, the matrix-valued losses
and the matrix gradient estimator
$\widehat{\nabla_\theta L}^{\rm mat}$ sketched in the
online appendix point to effective regularisers that
bias the gradient flow toward a chosen target geometry,
and to partial estimators of aspects of the
parameter-space process from the trajectory $\{M(t)\}$
(full inversion is underdetermined because $M(t)$ is
many-to-one in $\theta$). On the
deployment side, the gap-formation diagnostic and the
trajectory-level FDM observables give a basis for online
monitoring of inference processes, including
regime-change detection on streaming data.

%==============================================================================
\appendix
%==============================================================================
% Restart equation, figure, and table numbering with the appendix
% prefix (A.1, A.2, ... in Appendix A; B.1, B.2, ... in Appendix B;
% etc.) so the appendix derivations have a self-contained
% numbering separate from the body of the paper.
\setcounter{equation}{0}
\renewcommand{\theequation}{\Alph{section}.\arabic{equation}}
\setcounter{figure}{0}
\renewcommand{\thefigure}{\Alph{section}.\arabic{figure}}
\setcounter{table}{0}
\renewcommand{\thetable}{\Alph{section}.\arabic{table}}

\section{Detailed It\^o-SDE derivation for $M_{ij}(t)$}
\label{app:ito-derivation}

This appendix records the step-by-step derivation of the
matrix-valued It\^o SDE \eqref{eq:M_sde} of
Section~\ref{subsec:ito_sde} from the parameter-space
Langevin SDE \eqref{eq:weight_sde}. The discretised
per-pair linear regression in $\nabla_\theta L$, the
closed-form matrix gradient estimator
$\widehat{\nabla_\theta L}^{\rm mat}$, and the matrix
regulariser are deferred to the accompanying online
appendix. Throughout, $\langle \cdot,\,\cdot \rangle$
denotes the Euclidean inner product on $\mathbb{R}^{P}$.

First, applying It\^o's lemma to $M_{ij}(\theta(t))$, let $c_{ij}(\theta) = \hat h_i(\theta) \cdot \hat h_j(\theta)$
denote the normalised inner product of the $i$th and $j$th
representations, so $M_{ij} = \arccos(c_{ij})$ with
$\nabla_\theta M_{ij} = -(\nabla_\theta c_{ij})/\sin M_{ij}$.
For any smooth $f(\theta)$ of an It\^o process $\theta(t)$
with quadratic covariation
$d\theta_a\, d\theta_b = \Sigma_{ab}\, dt$, It\^o's lemma
is $df = \langle \nabla_\theta f, d\theta\rangle
+ \tfrac{1}{2}\mathrm{Tr}[\Sigma_\theta \nabla_\theta^2 f]\, dt$.
Applying it first to $c_{ij}$ and then to
$M_{ij} = \arccos(c_{ij})$,
\begin{equation}
dM_{ij}(t) \;=\; -\frac{\langle \nabla_\theta c_{ij},\, d\theta\rangle}{\sin M_{ij}}
\;-\; \frac{\mathrm{Tr}[\Sigma_\theta\,\nabla_\theta^2 c_{ij}]}{2\sin M_{ij}}\, dt
\;-\; \frac{\cos M_{ij}\,(\nabla_\theta c_{ij})^{\!\top}\Sigma_\theta(\nabla_\theta c_{ij})}{2\sin^3 M_{ij}}\, dt.
\label{eq:ito_step1}
\end{equation}

Second, substituting the Langevin SDE for $d\theta$, plugging \eqref{eq:weight_sde} into the linear $d\theta$
term of \eqref{eq:ito_step1} gives the matrix-valued
It\^o SDE \eqref{eq:M_sde} with drift
\begin{equation}
\mu_{ij}(\theta) \;=\;
\frac{\eta\,\langle \nabla_\theta c_{ij},\,\nabla_\theta L\rangle
\;-\; \tfrac{1}{2}\mathrm{Tr}[\Sigma_\theta\,\nabla_\theta^2 c_{ij}]
\;-\; \tfrac{\cos M_{ij}}{2\sin^2 M_{ij}}(\nabla_\theta c_{ij})^{\!\top}\Sigma_\theta(\nabla_\theta c_{ij})}{\sin M_{ij}}
\label{eq:M_drift}
\end{equation}
and diffusion vector
$(\bm{\sigma}_{ij}(\theta))_b
= -(\nabla_\theta c_{ij})^{\!\top}\,\sigma_{\cdot b}(\theta)/\sin M_{ij}$.

\section{Loss symmetries and equivariant gradient flow}
\label{app:loss-symmetries}

The spectrum of $M(t)$ is invariant under the trivial
gauge action of the global isometry group of the sphere
(simultaneous rotations of all particles), but this gauge
invariance is automatic and not informative. The operative
question is: given input symmetry $G_{\rm in}$ acting on
$x$ and output symmetry $G_{\rm out}$ acting on $y$, what
condition on the loss $L(\theta)$ guarantees that the
gradient flow produces an internal representation
equivariant under $G_{\rm in} \times G_{\rm out}$? The
condition is the standard equivariant-learning one,
translated to the matrix-valued setting. If there exists
an induced action $g \mapsto \Phi_g$ on parameter space
such that
$f_{\Phi_g(\theta)}(x) = \tilde g \cdot f_\theta(g^{-1} \cdot x)$
for the joint group element $(g, \tilde g) \in G_{\rm in}
\times G_{\rm out}$, then loss invariance
$L(\Phi_g(\theta)) = L(\theta)$ implies equivariant
gradients $\nabla_\theta L$ and an equivariant noise
covariance $\Sigma_\theta$, so both the drift and the
diffusion of the weight Langevin SDE \eqref{eq:weight_sde}
commute with $\Phi_g$. The induced It\^o SDE
\eqref{eq:M_sde} for $M_{ij}(t)$ then has drift and
diffusion that commute with the data-symmetry action, and
the gradient flow explores a $\Phi_g$-invariant
submanifold of weight space, so the spectral multiplet
structure of $M(t)$ ends up matching the irreducible
representations of $G_{\rm in} \times G_{\rm out}$. When
$L(\theta)$ is \emph{not} invariant (the generic case for
a randomly-initialised neural network on a
symmetry-rich task), the gradient flow may still reach a
symmetry-respecting representation, but only as one
attractor among many. The symmetry is then ``discovered''
rather than imposed, which is the grokking-style
delayed-generalisation phenomenology described in
\cite{halperin2026grokking}. The matrix-valued losses of
the online appendix are an explicit construction of
$L$-symmetries that respect a chosen target group action
$\Phi_g$ at the level of $M(t)$, giving a constructive
route to symmetry restoration without writing it into the
architecture.

\subsection{Matrix state space and symmetry fixed
points}
\label{app:fixed-points}

This appendix makes precise the body-text observation
that the spectrum of $M(t)$ settles to a seed-consistent
integer after a Group~B transition. It identifies the
state space the matrix dynamics live on, defines the
symmetry-fixed family $\mathcal F_G$ on which the
post-event spectra sit, separates $\mathcal F_G$ from the
stronger smooth-manifold BBS fixed point, and gives
conditions under which $\mathcal F_G$ is an attractor of
the projected matrix dynamics.

The state $M(t)$ lives in the set
$\mathcal D_N \subset \mathrm{Sym}_N(\mathbb R)$ of
realisable spherical angular-distance matrices: symmetric,
zero diagonal, entries in $[0, \pi]$, and expressible as
$M_{ij} = \arccos(\hat h_i \cdot \hat h_j)$ for some
configuration $(\hat h_1, \dots, \hat h_N) \in
(S^{D-1})^N$. This realisability is preserved
automatically by the parameter dynamics. The global
isometry $O(D)$ rotates the whole configuration while
leaving $M$ fixed, so the operative state space is the
quotient $\mathcal Q_N = (S^{D-1})^N / O(D)$.

A task symmetry group $G$ acts on the $N$-sample through a
permutation representation $\pi : G \to S_N$,
$g \mapsto \Pi_g$. For a sample closed under $G$, the
\emph{symmetry-fixed family} is
\begin{equation}
\mathcal F_G \;:=\; \bigl\{ M \in \mathcal D_N :
\Pi_g\, M\, \Pi_g^{\!\top} = M \ \text{for all } g \in G
\bigr\}.
\label{eq:symmetry-fixed-family}
\end{equation}
OMD evaluation samples are i.i.d.\ and generically not
closed under $G$, so exact invariance is replaced by
distributional equivariance
$M_{ij} \stackrel{d}{=} M_{gi,\, gj}$, quantified by the
symmetry-defect functional
\begin{equation}
V_G(M) \;:=\; \frac{1}{|G|} \sum_{g \in G}
\bigl\| M - \Pi_g\, M\, \Pi_g^{\!\top} \bigr\|_F^{2}
\;\ge\; 0,
\label{eq:V_G}
\end{equation}
which vanishes iff $M \in \mathcal F_G$.

Membership in $\mathcal F_G$ is weaker than a
smooth-manifold BBS fixed point. A $\mathbb Z_2$
two-cluster $S^0$ is symmetry-fixed but a finite atomic
set, not a smooth sub-manifold, so the literal predictions
$h(1, d) = d$ and $\beta_{\rm del} = d/(d-1)$ do not apply.
Those need the stronger smooth-manifold ansatz
$M_\star = M_{\mathcal M_d} + \epsilon R$ of
\eqref{eq:ibbs_decomp}, with $\mathcal M_d$ a smooth
$d$-sub-manifold and $\epsilon R$ the I-BBS ambient-noise
correction. Across the OMD experiments this holds only for
grokking (the BBS-adjacent Fourier-soliton on $S^1$); the
other Group~B cases lie in $\mathcal F_G$ for finite $G$
but outside the smooth-manifold regime
(Table~\ref{tab:regimes}).

A fixed-point claim at the matrix level needs care, because
$M(t)$ is an observable projection of the parameter
dynamics, not a closed Markov process: the It\^o SDE
\eqref{eq:M_sde} has drift and diffusion driven by
$\theta$, $\nabla_\theta L$, $\nabla_\theta^2 c_{ij}$, and
$\Sigma_\theta$, none a function of $M$ alone. Writing
$M(\theta)$ for the parameter-to-matrix map with Jacobian
$D_\theta M$, the deterministic velocity is
\begin{equation}
\dot M(t) \;=\; D_\theta M(\theta(t))\, \dot\theta(t),
\qquad
\bigl[\dot M\bigr]_{ij} \;=\; \sum_{a=1}^{P}
\partial_{\theta_a} M_{ij}(\theta)\; \dot\theta_a.
\label{eq:M_chainrule}
\end{equation}
A matrix fixed point therefore requires only
$D_\theta M(\theta_\star)\, \nabla_\theta L(\theta_\star)
= 0$, weaker than $\nabla_\theta L(\theta_\star) = 0$: the
parameters may keep moving along $\ker D_\theta M$ (the
$O(D)$ and per-layer weight-rescaling gauge directions that
leave $\hat h$ unchanged) without disturbing $M$. The
matrix marginal $M_t$ is approximately Markov only when
this projected drift is approximately closed, condition
(ii) below.

Let $\mathcal A$ be the infinitesimal generator of the
joint $(\theta_t, M_t)$ process and
$\sigma_\theta^2 := \mathrm{tr}\,\Sigma_\theta / P$ the
parameter-noise scale of \eqref{eq:weight_sde}. Since
$V_G$ depends on $M$ only, $\mathcal A V_G(M(\theta))$ is
set by the $M$-projected drift and diffusion. Under three
conditions, $\mathcal F_G$ is attracting up to optimiser
noise and non-closure error.
\begin{enumerate}
\item[(i)] \emph{Equivariance:} an induced parameter
action $\Phi_g$ with $L \circ \Phi_g = L$ and
$\Sigma_{\Phi_g(\theta)} = \Phi_g \Sigma_\theta
\Phi_g^{\!\top}$, so loss and noise are $G$-equivariant.
\item[(ii)] \emph{Approximate closure:} $D_\theta M(\theta)
\,(-\eta\, \nabla_\theta L(\theta)) = b_M(M(\theta)) +
r(\theta)$ for a closed drift $b_M : \mathcal D_N \to
\mathrm{Sym}_N(\mathbb R)$ and a residual with
$\sup_\theta \|r(\theta)\|_F \le \eta_{\rm cl}$ (closure
error, distinct from the I-BBS ambient noise $\epsilon$).
\item[(iii)] \emph{Dissipativity:} $\mathcal A V_G(M) \le
-\kappa\, V_G(M) + C_1 \sigma_\theta^2 + C_2 \eta_{\rm cl}$
for $M$ near $\mathcal F_G$, with contraction rate
$\kappa > 0$ and geometric prefactors $C_1, C_2 > 0$.
\end{enumerate}

These give the attractor structure. On $\mathcal F_G$,
$V_G \equiv 0$, preserved by (i) under deterministic
dynamics (forward invariance). In the noise-free closed
limit ($\sigma_\theta = 0$), Gr\"onwall's inequality
\cite{gronwall1919} on $\dot V_G \le -\kappa V_G +
C_2 \eta_{\rm cl}$ gives
\begin{equation}
V_G(M_t) \;\le\; e^{-\kappa t}\, V_G(M_0)
\;+\; \frac{C_2 \eta_{\rm cl}}{\kappa}\,
\bigl(1 - e^{-\kappa t}\bigr),
\label{eq:VG_det}
\end{equation}
so $M_t$ enters and stays in the tube
$\{V_G \le C_2 \eta_{\rm cl}/\kappa\}$ around
$\mathcal F_G$ at rate $\kappa$. With noise the joint
process is geometrically ergodic to a stationary law
$\rho_\star$ with
\begin{equation}
\mathbb E_{\rho_\star}\bigl[V_G(M)\bigr]
\;\le\;
\frac{C_1\, \sigma_\theta^2 + C_2\, \eta_{\rm cl}}{\kappa}.
\label{eq:VG_stoch}
\end{equation}
So $\mathcal F_G$ is an attractor up to a tube of radius
$O\bigl(\sqrt{(\sigma_\theta^2 + \eta_{\rm cl})/\kappa}
\bigr)$ in the $V_G$ metric. The three conditions are
sufficient but not verified for the OMD experiments; the
post-transition stability of $\hat h_1$
(Section~\ref{sec:results}) and the basin-of-attraction
signature ($\sigma_W = 0.10$ returns) of
Appendix~\ref{app:delta-G-probes} are its empirical
fingerprints, not a proof.

\section{Robustness, sensitivity, and calibration appendices}
\label{app:robustness}

This appendix collects sensitivity analyses and calibration
checks that support the diagnostic claims of the main
text. The ablations test whether reported quantities are
artifacts of specific analysis choices ($\beta$-fit
window, evaluation-sample size $N$, random seed) or
stable signals of the underlying representation dynamics.

\subsection{$\beta$-fit window ablation}
\label{app:beta_window}

The rank-decay exponent $\beta(t)$ is fitted by OLS of
$\log |\lambda_K|$ on $\log K$ over the canonical
window $K \in [2, 50]$, inherited from
\cite{halperin2026FDM}. BBS predicts $\beta = d/(d-1)$ on
its asymptotic large-$|\lambda|$ branch with per-eigenvalue
accuracy for $K \lesssim \sqrt N$ (BBS Eqs.~64--66, 91; the
averaged-counting-function regime, BBS Eq.~89, extends the
same asymptotic to a finite fraction of all eigenvalues,
which is what the smoothed log-log OLS reads). The lower
bound $K \geq 2$ excludes the Perron eigenvalue.
\par
$K_{\max} = 50$ is picked via a goodness-of-fit scan on
the post-grokking spectrum
(Table~\ref{tab:beta_window_gof}): for each $K_{\max}$ we
report $\hat\beta$, log-log $R^2$, and the
Kolmogorov--Smirnov $p$-value of standardised residuals
against $\mathcal N(0, 1)$. The sweet spot
$K_{\max} \in [40, 50]$ has $R^2$ at its plateau
($0.85$--$0.86$), KS-$p$ at its maximum ($0.72$--$0.84$,
noise-like residuals), and $\hat\beta$ within $0.05$ of
the BBS prediction $\beta = 3/2$ for the post-grokking
$T^2$. The strict per-eigenvalue boundary $K_{\max} = 20$
gives only $18$ points and $R^2 = 0.67$, too few for a
stable slope. For $K_{\max} \geq 85$ the KS test rejects
normality ($p < 0.05$), the empirical signature of the
delocalised/localised crossover
(Appendix~\ref{app:participation_ratio}). The transition
step is window-stable across the full
$K_{\max} \in [20, 100]$ band on both endogenous
transitions (grokking, sparse parity).

\begin{table}[htbp]
\centering
\small
\caption{Goodness-of-fit scan on the post-grokking spectrum,
5-seed mean ($N = 400$). $\hat\beta$ is the OLS slope on
$K \in [2, K_{\max}]$, $R^2$ the log-log coefficient of
determination, and KS-$p$ the Kolmogorov--Smirnov $p$-value
of standardised residuals against $\mathcal{N}(0, 1)$. Bold
marks the sweet spot $K_{\max} \in [40, 50]$.}
\label{tab:beta_window_gof}
\begin{tabular}{r c c c c}
\toprule
$K_{\max}$ & $\hat\beta$ & $R^2$ & KS stat & KS $p$ \\
\midrule
20  & 1.44 & 0.67 & 0.18 & 0.55 \\
25  & 1.61 & 0.76 & 0.16 & 0.55 \\
30  & 1.62 & 0.81 & 0.12 & 0.72 \\
35  & 1.59 & 0.83 & 0.11 & 0.75 \\
\textbf{40} & \textbf{1.54} & \textbf{0.85} & \textbf{0.09} & \textbf{0.84} \\
\textbf{45} & \textbf{1.49} & \textbf{0.85} & \textbf{0.10} & \textbf{0.79} \\
\textbf{50} & \textbf{1.45} & \textbf{0.86} & \textbf{0.10} & \textbf{0.72} \\
60  & 1.37 & 0.87 & 0.11 & 0.45 \\
80  & 1.27 & 0.88 & 0.16 & 0.05 \\
90  & 1.24 & 0.88 & 0.19 & 0.004 \\
100 & 1.23 & 0.89 & 0.21 & 0.001 \\
\bottomrule
\end{tabular}
\end{table}

\subsection{Estimator choice and within-fit error bar
on $\beta$}
\label{app:beta_estimator}

The rank-decay exponent $\beta(t)$ is the negated OLS
slope of $\log|\lambda_K|$ vs $\log K$ on the
canonical window $K \in [2, 50]$. The within-fit
one-$\sigma$ error $\sigma_\beta^{\rm fit}$ is the
square-root of the slope-variance entry of the
least-squares fit covariance. This
per-checkpoint, per-seed uncertainty
($\approx 0.03$ on the grokking memorisation plateau,
$\approx 0.08$ on the post-grokking plateau) is distinct
from the across-seed dispersion $\sigma_\beta^{\rm seed}$
($0.02$--$0.23$ depending on checkpoint, peaking in the
transition window where grokking-step dispersion inflates
$\beta$-dispersion at any fixed step).

A maximum-likelihood alternative is the natural competitor, but neither variant substitutes well. It treats eigenvalues as
Pareto samples $f(\lambda) \propto \lambda^{-\alpha}$
above $x_{\min}$ (Hill estimator or \texttt{powerlaw.Fit}
\cite{alstott2014powerlaw} with KS-selected $x_{\min}$)
and converts via $\beta = 1/(\alpha - 1)$. Across the grokking
transition rank-decay OLS produces a jump $\Delta\beta
\approx 1.0$ ($0.55 \to 1.5$). Hill on the same window
gives $\lesssim 0.3$, while \texttt{powerlaw.Fit} leaves
$\hat\alpha$ drifting in $[1.78, 2.03]$ with no clean
transition. The mechanism is BBS's two-branch structure:
a delocalised large-$|\lambda|$ branch
$\rho(\lambda) \sim |\lambda|^{-(2d-1)/d}$ (the
rank-decay OLS reads this on $K \in [2, 50]$) and a
localised small-$|\lambda|$ branch
$\rho(\lambda) \sim |\lambda|^{d-2}$ (BBS Secs.~2.4,
4.4), joined at $K \sim \sqrt N$. Auto-selected
$x_{\min}$ lands deep in the bulk
(Table~\ref{tab:ww_consistency}), so $\hat\alpha$ reads
the localised branch. $\alpha = 1 + 1/\beta$ holds
within either branch but not across the crossover.

A third independent implementation,
\texttt{WeightWatcher} \cite{martin2021weightwatcher},
fits a power-law to the empirical spectral density of
$W^\top W$. Supplied with a synthetic layer whose
$W^\top W$ spectrum equals $|\lambda_K(M(t))|$, it
returns $\hat\alpha_{\rm ww}$ at KS-selected $x_{\min}$.
The three estimators are compared post-grokking in
Table~\ref{tab:ww_consistency}.

\begin{table}[htbp]
\centering
\small
\begin{tabular}{r r r r r r r}
\toprule
seed & ww $\hat\alpha$ & ww $x_{\min}$ & $K_{x_{\min}}$ &
$\beta_{[2,50]}$ & $1 + 1/\beta_{[2,50]}$ &
$1 + 1/\beta_{[2, K_{x_{\min}}]}$ \\
\midrule
0 & 1.89 & 0.13 & 392 & 1.48 & 1.68 & 1.84 \\
1 & 1.93 & 0.35 & 180 & 1.52 & 1.66 & 1.83 \\
2 & 2.01 & 0.38 & 205 & 1.48 & 1.68 & 1.87 \\
5 & 1.90 & 0.33 & 187 & 1.50 & 1.67 & 1.82 \\
\bottomrule
\end{tabular}
\caption{Three-estimator consistency at the post-grokking
checkpoint (step $5000$, four seeds). On the canonical
window $[2, 50]$ rank-decay OLS gives $\beta \approx 1.49$
(BBS $3/2$ for $T^2$). \texttt{WeightWatcher}'s wider
$x_{\min}$ window ($K_{x_{\min}} \approx 180$--$392$) gives
$\hat\alpha \approx 1.9$--$2.0$, reproduced to two decimals
by $\alpha = 1 + 1/\beta$ from the matched-window OLS
($\beta \approx 1.2$). The spectral knee, steep BBS slope
at small $K$ and shallower bulk at large $K$, is why
$\alpha = 1 + 1/\beta$ holds within either branch but not
across.}
\label{tab:ww_consistency}
\end{table}

\subsection{Participation-ratio crossover}
\label{app:participation_ratio}

\begin{figure}[htbp]
  \centering
  \includegraphics[width=\textwidth]{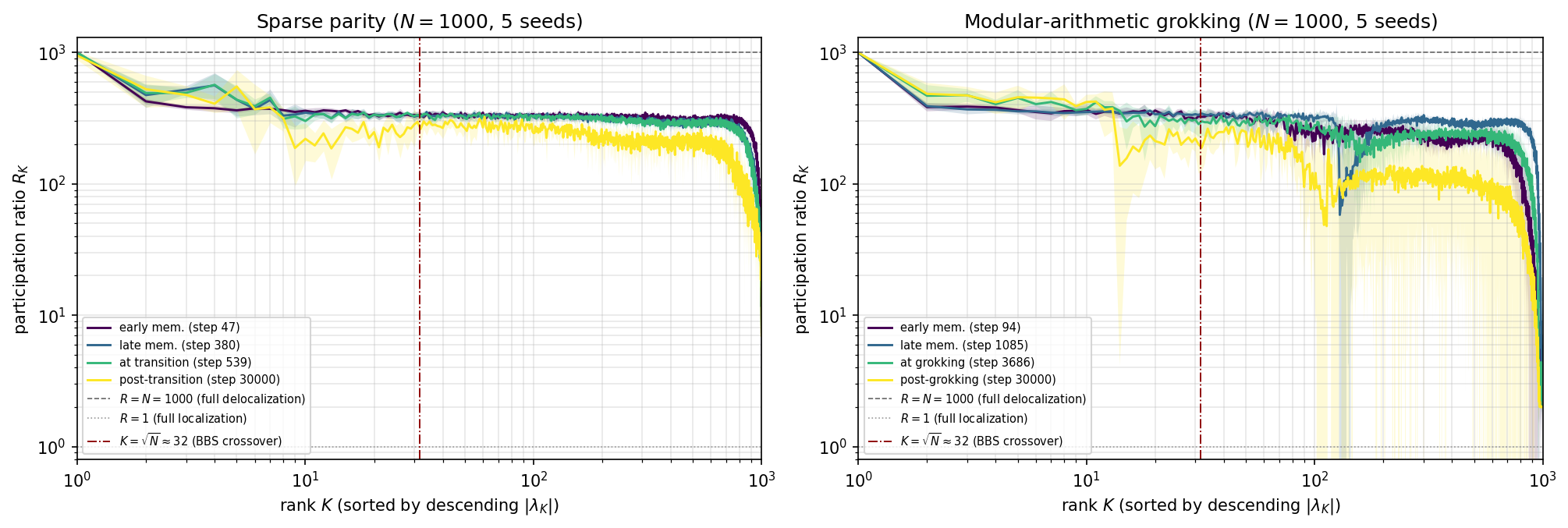}
  \caption{Participation ratio $R_K$
  \eqref{eq:participation_ratio} on $M(t)$ at four phases,
  $N=1000$, five seeds; sparse parity (left), grokking
  (right), eigenvectors ranked by descending $|\lambda_K|$.
  Dash-dotted line: BBS crossover $K = \sqrt N$.}
  \label{fig:participation_ratio}
\end{figure}

The two-regime spectrum has an eigenvector counterpart:
eigenvectors are delocalised across all $N$ particles
at large $|\lambda|$ and concentrated on few particles
at small $|\lambda|$ (the Anderson-localization
analogue of \cite{bogomolny2003}). The diagnostic is
the participation ratio
\begin{equation}
  R_n(t) \;=\;
  \frac{\bigl(\sum_{j=1}^{N} |u_j^{(n)}(t)|^2\bigr)^2}
       {\sum_{j=1}^{N} |u_j^{(n)}(t)|^4}
  \;\in\; [1, N],
  \label{eq:participation_ratio}
\end{equation}
with $R \sim N$ uniformly spread, $R \sim O(1)$
localised. BBS predict the delocalised regime for
$K \lesssim \sqrt N$. On the FBP-on-$S^2$ test bed
\cite{halperin2026FDM} reports $R$ from $\sim 2$ at
small $|\lambda|$ to $\sim 200$--$280$ at $|\lambda|
\sim 100$ ($N = 400$, $d = 2$), the below-$N$
saturation attributed there to spherical-harmonic
multiplet block structure.
Figure~\ref{fig:participation_ratio} extends to the
trained-network setting: $R(K)$ on $M(t)$ at four
phases for sparse parity and grokking, $5$ seeds,
$N = 1000$. The two-regime picture is qualitatively
confirmed ($R \approx N$ at $K = 1$, $R$ of order
$2$--$8$ at $K = N$, smooth crossover). The delocalised
band contracts post-transition (pre-grokking
$R(K{=}20) \approx 338$, $R(K{=}100) \approx 319$;
post-grokking $212$ and $91$; sparse parity $337, 322 \to
248, 244$). Strict $R \sim N$ for $K \lesssim \sqrt N$ is
not observed pre-transition ($R \approx 0.35\,N$, the
same multiplet-block saturation as on FBP-on-$S^2$). The
crossover is broad, spanning $K \in [\sqrt N, N/4]$
pre-transition and contracting to $K \lesssim \sqrt N$
post-transition. The same $\sqrt N$ boundary controls
per-eigenvalue BBS accuracy in
Appendix~\ref{app:beta_window}.

\subsection{Smooth-manifold positive controls for
Algorithm 1}
\label{app:calibration}

\begin{figure}[htbp]
  \centering
  \includegraphics[width=0.85\textwidth]{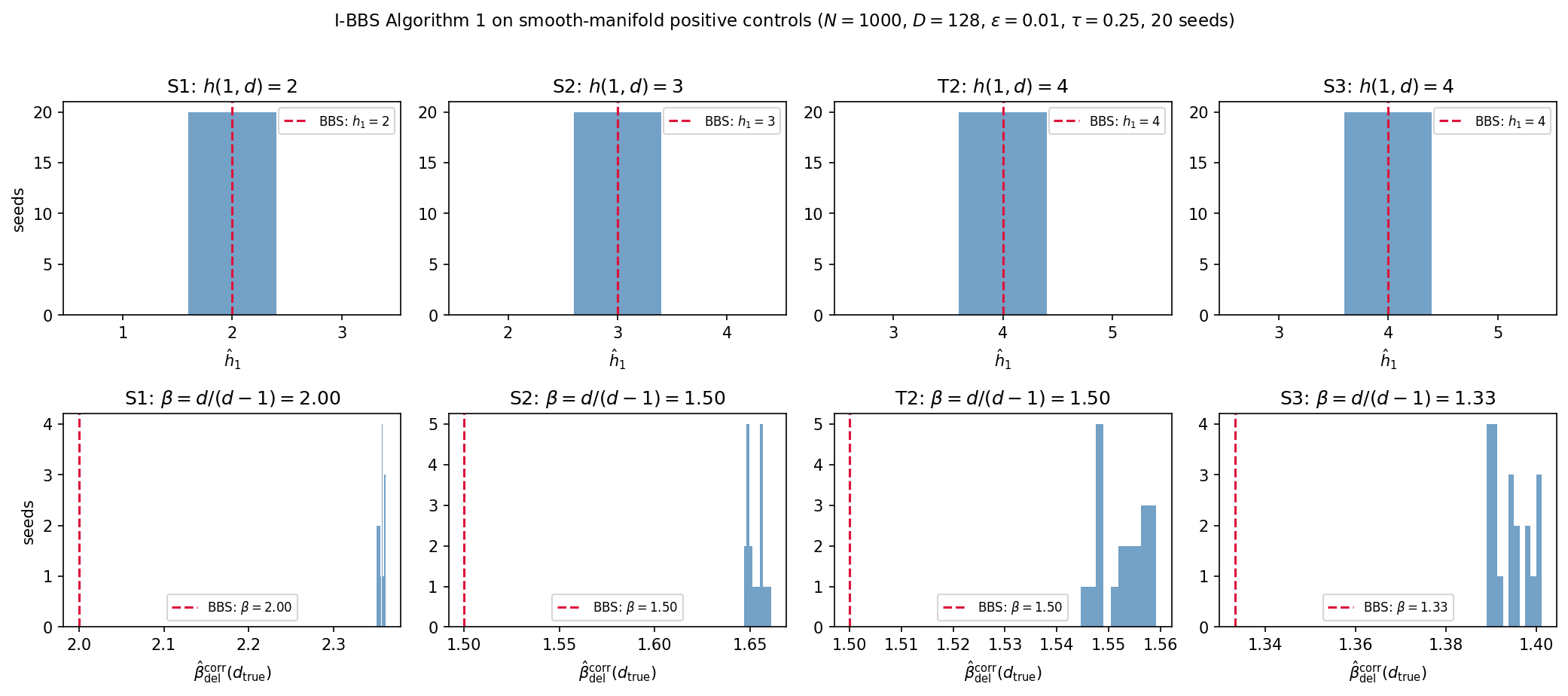}
  \caption{Per-seed I-BBS readouts on the
  smooth-manifold positive controls ($N = 1000$,
  $D = 128$, $\epsilon = 0.01$, $\tau = 0.25$, $20$
  seeds). Top: $\hat h_1$ histograms, every seed on the
  predicted integer (red dashed). Bottom: corrected slope
  $\hat\beta_{\rm del}^{\rm corr}(d_{\rm true})$ vs target
  $d/(d-1)$ (red dashed), concentrated within
  $\sim 5$--$18\,\%$ of target.}
  \label{fig:synthetic_calib}
\end{figure}

We test the I-BBS toolkit on synthetic samples from
smooth manifolds of known geometry at $N = 1000$,
$D = 128$, RSM noise $\epsilon = 0.01$ (so the
cosine-Frobenius noise level $\eta_{\cos} \approx
\epsilon^2$ sits deep in the perturbative regime), $20$
seeds. For each
of $S^1, S^2, T^2, S^3$ we generate uniform samples,
embed in $\mathbb R^D$, add $\epsilon\xi$ noise,
$L^2$-normalise, form the arccos distance matrix, and
run Algorithm~1 at $\tau = 0.25$.
Table~\ref{tab:positive-controls} and
Figure~\ref{fig:synthetic_calib} summarise.

\begin{table}[htbp]
\centering
\footnotesize
\begin{tabular}{l c c c c}
\toprule
Manifold & BBS $h_1$ & Recovered $\hat h_1$ (20 seeds)
& BBS $\beta_{\rm del}$ & $\hat\beta_{\rm del}^{\rm
corr}(d_{\rm true})$ (mean $\pm$ std) \\
\midrule
$S^1$ ($d = 2$) & $2$ & $2$ (20/20) & $2.000$ & $2.36 \pm 0.003$ \\
$S^2$ ($d = 3$) & $3$ & $3$ (20/20) & $1.500$ & $1.65 \pm 0.004$ \\
$T^2$ ($S^1{\times}S^1$) & $4$ & $4$ (20/20) & $1.500$ & $1.55 \pm 0.004$ \\
$S^3$ ($d = 4$) & $4$ & $4$ (20/20) & $1.333$ & $1.39 \pm 0.004$ \\
\bottomrule
\end{tabular}
\caption{I-BBS Algorithm~1 on smooth-manifold positive
controls. The multiplet diagnostic recovers $h(1, d)$
exactly in $20/20$ seeds, with $h(1, d) = d$ on spheres
(embedding dimension $d$, intrinsic $d-1$) and $h_1 = 4$
on $T^2$ (two $S^1$ Fourier pairs, I-BBS Table~4),
distinguishing $T^2$ from $S^2$ at the same intrinsic
dimension $2$. The corrected
delocalised slope sits $\sim 4$--$18\,\%$ above target at
$\epsilon = 0.01$ (a residual noise-induced offset on the
$\Delta\beta$ correction of \cite{halperin2026IBBS}).
Residual-RMT is RSM-like in $20/20$ for each manifold
(Wigner-KL $1.3$--$1.9$).}
\label{tab:positive-controls}
\end{table}

The integer $\hat h_1$ is thus the primary readout, with
the corrected slope retaining a small noise-induced shift.
$d_\beta$ is therefore a spectral observable that ranks
manifold complexity rather than a literal dimension
estimator under ambient noise.

\subsection{Finite-cluster controls for Algorithm 1}
\label{app:finite-cluster-controls}

Each finite vertex set should produce a clean integer
$\hat h_1$ counting the between-vertex contrasts of the
lowest non-trivial irrep, and a delocalised slope not
matching any smooth-$S^{d-1}$ target. The simulation
recipe is uniform across configurations. We choose a
finite vertex set
$\{v_1, \dots, v_K\} \subset S^{D-1}$ in the chosen
geometry (placement rules below). We take $\lfloor N/K
\rfloor$ copies at each vertex, embed into
$\mathbb R^D$ by writing the vertex coordinates into
the first few axes, add RSM noise
$\epsilon\,\xi_i$ with $\xi_i \sim \mathcal N(0, I_D)$, and
re-project onto $S^{D-1}$ by $L^2$-normalisation. Finally we form
$M = \arccos(\hat h \hat h^\top)$ and its rank-ordered
absolute spectrum. Repeat over $20$ seeds and report
mean$\,\pm\,1\sigma$ (or median + IQR). The seven
configurations of
Table~\ref{tab:finite-cluster-controls} are constructed as:
\begin{itemize}
\item \emph{$\mathbb Z_2$ antipodal}: $K = 2$ vertices
$\pm \hat e_1 \in S^{D-1}$; closed-form derivation in
Appendix~\ref{app:antipodal_blobs}.
\item \emph{$\mathbb Z_k$ on $S^1$} ($k \in \{4, 8\}$):
$K = k$ vertices at $v_q = (\cos\theta_q, \sin\theta_q,
0, \dots, 0)$ with $\theta_q = 2\pi q/k$, $q = 0,
\dots, k - 1$, embedded into $\mathbb R^D$ via the
first two coordinates. The arc-distance matrix on the
$k$ vertices is circulant. Its DFT eigenvalues, scaled
by $N/k$, give the closed-form $\sigma \to 0$
spectrum \cite{gray2006circulant} listed in
Table~\ref{tab:closed-form-vs-sim}.
\item \emph{$\mathbb Z_8$ cube on $(\mathbb Z_2)^3$}:
$K = 8$ vertices $v_b \in \{\pm 1/\sqrt 3\}^3 \subset
\mathbb R^3 \subset S^{D-1}$ for $b \in \{0, 1\}^3$;
the lowest non-trivial irrep is the $3$-dim sign-flip
representation, giving $\hat h_1 = 3$.
\item \emph{$\mathbb Z_8$ random vertices}: $K = 8$
i.i.d.\ uniform points on $S^{D-1}$, drawn once and
fixed; no symmetry, all $C - 1 = 7$ between-vertex
contrasts emerge as separate eigenvalues. The only
configuration in the table that admits no closed form
(see Section~\ref{sec:discussion}).
\item \emph{Simplex-ETF, $C = 10$}: $K = C$ vertices
$v_c = \tfrac{1}{\sqrt{1 - 1/C}}(e_c - \mathbf 1/C)
\in \mathbb R^C \subset S^{D-1}$, the equiangular
tight frame of \cite{sustik2007etf}; pairwise Gram
$-1/(C-1)$.
\item \emph{$\mathbb Z_p$ Fourier soliton}: $K = p =
113$ tokens at $\theta_t = 2\pi t/p$, $t = 0, \dots,
p - 1$, with embedding $v_t \propto
\bigl(\cos(m\theta_t), \sin(m\theta_t)\bigr)_{m=1}^{k_{\rm active}}
\in \mathbb R^{2 k_{\rm active}}$ on the first
$2 k_{\rm active}$ axes ($k_{\rm active} = 6$ active
Fourier-mode pairs, matching the grokking readout of
\S\ref{subsec:grokking_results}). Closed-form
spectrum from the band-limited circulant in
Table~\ref{tab:closed-form-vs-sim}.
\end{itemize}
Settings throughout this appendix: $N = 1000$, $D = 128$,
$\epsilon = 0.05$, $\tau = 0.25$, $20$ seeds. The contrast between the $\mathbb Z_8$ cube
($\hat h_1 = 3$ from the $(\mathbb Z_2)^3$ symmetry's
$3$-dim sign-flip irrep) and the $\mathbb Z_8$
random-vertex configuration ($\hat h_1 = 7$ from all
$C-1 = 7$ contrasts) is the matrix fingerprint of
the sparse-parity $\hat h_1 = 7$ reading
(Section~\ref{sec:results-sparseparity}): the trained
network represents the $2^k = 8$ bit patterns as $8$
inequivalent vertices rather than a permutation-symmetric
orbit. The slope diagnostic is ambiguous on these
configurations (minimum distance of
$\hat\beta_{\rm del}^{\rm corr}$ to the closest
smooth-$S^{d-1}$ BBS target spans $0.04$--$1.69$, often
landing on $d_{\rm guess} = 4$ rather than at
$d_{\rm true}$). The multiplet combined with the
band-gap fingerprint is what identifies the regime.

\begin{table}[htbp]
\centering
\footnotesize
\setlength{\tabcolsep}{4pt}
\begin{tabularx}{\linewidth}{@{}l c c c X@{}}
\toprule
Configuration & predicted $h_1$ & recovered $\hat h_1$
& $\log$ band gap & Notes \\
\midrule
$\mathbb Z_2$ antipodal &
1 & 1 (20/20) & $2.19 \pm 0.004$ &
$S^0$; sparse-parity 2-cluster reduction, input-topology
boundary case \\
$\mathbb Z_4$ on $S^1$ &
2 & 2 (20/20) & $1.96 \pm 0.004$ &
task-switch ground truth (4 vertices, $\hat h_1 = 2$
doublet) \\
$\mathbb Z_8$ on $S^1$ &
2 & 2 (20/20) & $1.41 \pm 0.003$ &
GAN octagon (8 vertices on $S^1$, $\hat h_1 = 2$ from
the lowest Fourier mode) \\
$\mathbb Z_8$ cube ((Z$_2)^3$ symmetric) &
3 & 3 (20/20) & $1.06 \pm 0.002$ &
8 cube vertices in $\mathbb R^3$; lowest irrep is the
3-dim sign-flip representation, $\hat h_1 = 3$ \\
$\mathbb Z_8$ random vertices (no symmetry) &
7 & 7 (20/20) & $1.31 \pm 0.031$ &
8 random vertices on $S^{127}$ (the appendix-wide
$D = 128$); all 7 between-vertex contrasts emerge as
separate eigenvalues (\textbf{sparse-parity reduction}) \\
Simplex-ETF, $C = 10$ &
9 & 9 (20/20) & $1.37 \pm 0.005$ &
10-vertex simplex-ETF, neural-collapse asymptotic limit
\\
Fourier-soliton on $\mathbb Z_p$, $p = 113$, $k = 6$ &
12 & 12 (20/20) & $1.15 \pm 0.006$ &
grokking ground truth ($6$ active Fourier-mode pairs on
$S^1$, $\hat h_1 = 2k$) \\
\bottomrule
\end{tabularx}
\caption{Algorithm~1 on finite-cluster controls. Each
configuration recovers its predicted $\hat h_1$ in
$20/20$ seeds with a clean inter-multiplet gap
(log gap $\sim 1$--$2$). Residual-RMT is
RSM-like in $20/20$ on every configuration
(consistent with the injected RSM noise). See main text
for the $\mathbb Z_8$ cube vs random-vertex contrast
and the slope-diagnostic caveat.}
\label{tab:finite-cluster-controls}
\end{table}

\begin{figure}[!htbp]
  \centering
  \includegraphics[width=0.75\textwidth]{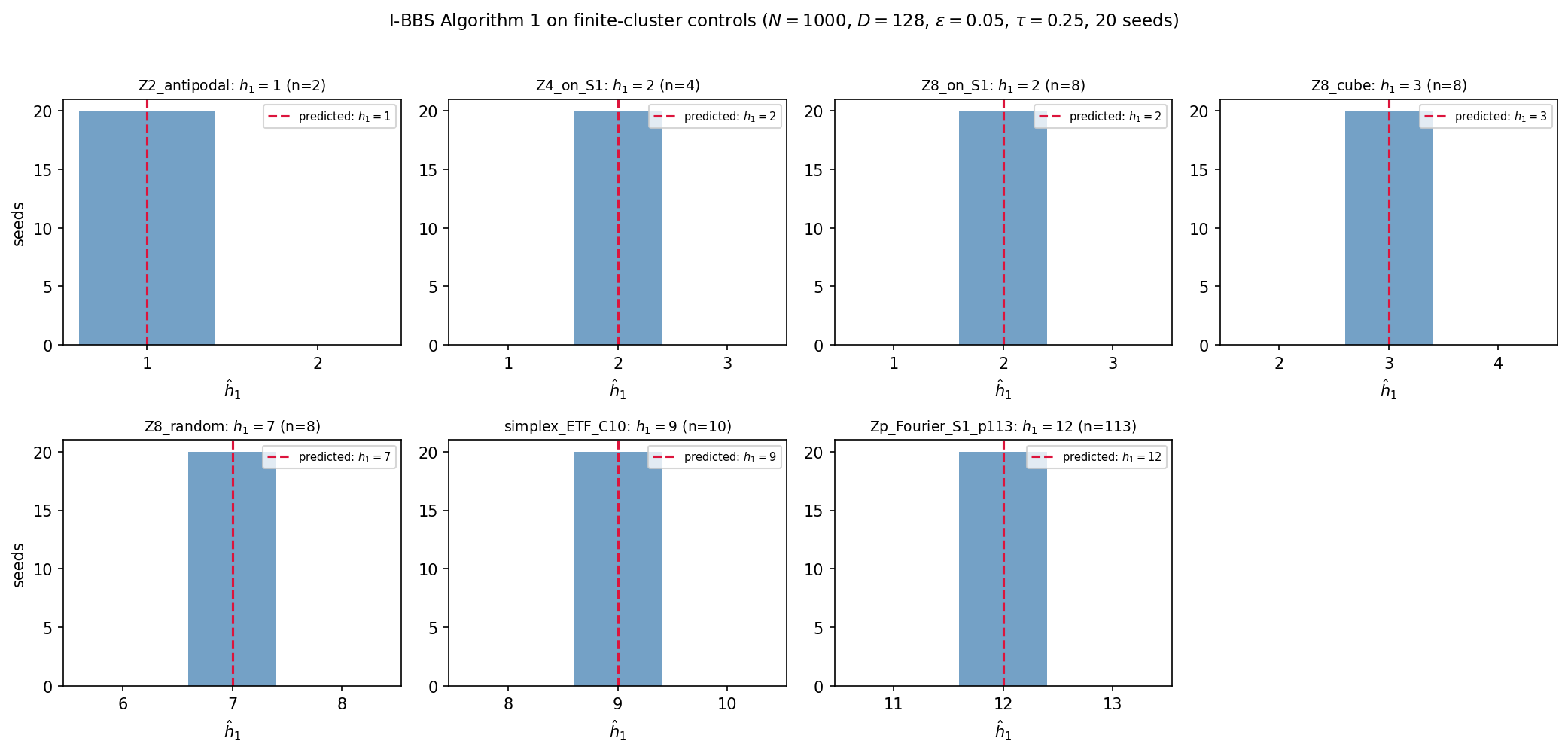}
  \caption{Per-seed $\hat h_1$ histograms for the
  finite-cluster controls of
  Table~\ref{tab:finite-cluster-controls}; red dashed:
  predicted $h_1$. Every seed lands on the prediction
  for every configuration.}
  \label{fig:synth-finite-cluster}
\end{figure}

\subsection{Noise-model separability null
(RSM vs.\ FSM)}
\label{app:noise-null}

\begin{figure}[htbp]
  \centering
  \includegraphics[width=0.9\textwidth]{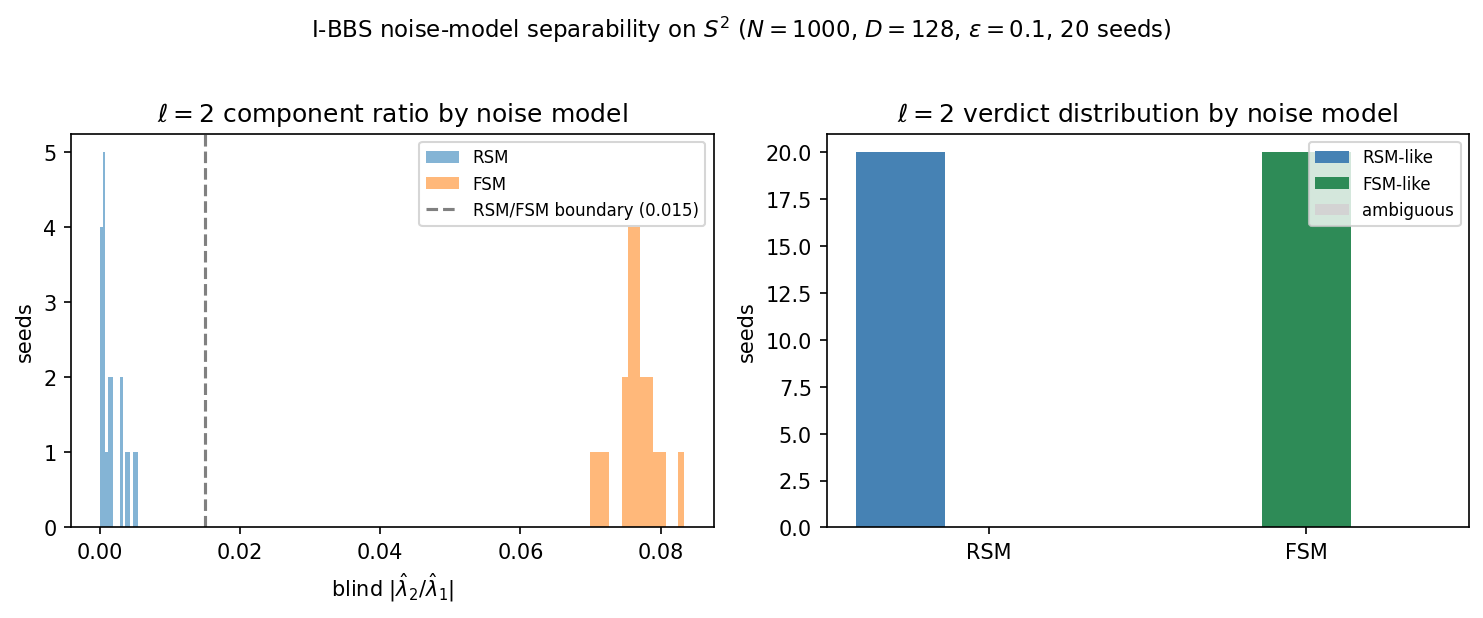}
  \caption{Noise-model separability on $S^2$.
  Left: per-model blind $\ell = 2$ component ratio
  $|\hat\lambda_2/\hat\lambda_1|$ (RSM at the parity
  floor near $0.002$, FSM near $0.08$), with the
  geometric-mean boundary $0.015$. Right: per-model
  $\ell = 2$ verdict counts ($20/20$ each).}
  \label{fig:synth-noise-null}
\end{figure}

We test the noise-model diagnostic on a controlled $S^2$
null at $N = 1000$, $D = 128$, $\epsilon = 0.10$,
$\tau = 0.25$, $20$ seeds. RSM is the convex-combination
forward $\cos M^{(D)} = (1 - \epsilon^2)\cos M^{(d)} +
\epsilon^2 \cos M^{(D-d)}$ with the residual Gram drawn
uniformly on $S^{D-d-1}$. FSM injects the degree-two
Gegenbauer profile $\mathbb E[\cos M^{(D)}] =
(1 - \epsilon)\cos M^{(d)} + \epsilon\,P_2(\cos M^{(d)})$
with the distance noise $\sqrt 2\,\epsilon\sin^2\theta\,
\xi$. The heat-kernel embedding is the isotropic limit of
RSM and is no longer a separate channel.
Table~\ref{tab:noise-null} and
Figure~\ref{fig:synth-noise-null} show the blind
$\ell = 2$ component separating the two classes with zero
overlap in $20 + 20$ seeds: RSM stays at the parity
floor, $|\hat\lambda_2/\hat\lambda_1| = 0.0018 \pm
0.0017$, well below the geometric-mean boundary $0.015$,
while FSM is populated at $0.0764 \pm 0.0030$. The
residual-RMT bulk is the secondary consistency check: it
is Wigner-consistent in $20/20$ FSM seeds and peaked with
curvature outliers in $20/20$ RSM seeds.

\begin{table}[htbp]
\centering
\small
\begin{tabular}{lccc}
\toprule
Channel & $\ell = 2$ verdict (20 seeds) & $|\hat\lambda_2/\hat\lambda_1|$ & Residual Wigner-consistent \\
\midrule
RSM & 20/20 RSM-like & $0.0018 \pm 0.0017$ & $0/20$ \\
FSM & 20/20 FSM-like & $0.0764 \pm 0.0030$ & $20/20$ \\
\bottomrule
\end{tabular}
\caption{Noise-model separability null on $S^2$
($N = 1000$, $D = 128$, $\epsilon = 0.10$, $20$ seeds).
The blind $\ell = 2$ component is the primary classifier;
the residual-RMT bulk is the secondary check.}
\label{tab:noise-null}
\end{table}

\subsection{Theoretical spectrum of two antipodal Gaussian
blobs on $S^{d-1}$}
\label{app:antipodal_blobs}

The Group~B atomic-cluster regime (sparse parity, input
topology, GAN $\mathbb Z_8$ leading band) is
approximated by $N$ points on $S^{d-1}$ in two tight
Gaussian blobs of width $\sigma$ at antipodal locations
$\pm \hat n$. We derive the spectrum of $M$ in closed
form to leading order in $\sigma$ as an analytical
reference complementing the simulation-based one of
Section~\ref{sec:discussion}: it confirms
$\hat h_1 = 1$, $\hat d = 0$ and predicts the
next-order structure visible in the simulation plots.

Parameterise the cloud as
\begin{equation}
\hat h_i \;=\; \frac{\epsilon_i\,\hat n + \sigma\,\xi_i}
                    {\|\epsilon_i\,\hat n + \sigma\,\xi_i\|},
\qquad
\xi_i \sim \mathcal N(0, I_{d-1}) \text{ in } T_{\hat n} S^{d-1},
\qquad
\epsilon_i \in \{+1, -1\},
\label{eq:blob_param}
\end{equation}
with the first $N/2$ indices in blob A ($\epsilon = +1$)
and the last $N/2$ in blob B ($\epsilon = -1$).

For small $\sigma$, expand the inner product
$c_{ij} = \hat h_i \cdot \hat h_j$ in powers of $\sigma$:
$\hat h_i \cdot \hat h_j = \epsilon_i \epsilon_j -
\tfrac{1}{2} \sigma^2 \|\epsilon_i \xi_i - \epsilon_j \xi_j\|^2
+ O(\sigma^4)$. Substituting into
$M_{ij} = \arccos c_{ij}$,
\begin{align}
M_{ij}^{\rm same} &\;\approx\; \sigma\,\|\xi_i - \xi_j\|,
&& (\text{same blob}),
\label{eq:M_same}\\
M_{ij}^{\rm cross} &\;\approx\; \pi - \sigma\,\|\xi_i + \xi_j\|,
&& (\text{cross blob}).
\label{eq:M_cross}
\end{align}

Decompose $M = \pi M^{(0)} + \sigma\,\delta M$ with
$M^{(0)} = \bigl(\begin{smallmatrix} 0 & J \\ J & 0 \end{smallmatrix}\bigr)$,
$J = \mathbf 1\mathbf 1^{\!\top}$ the
$(N/2)\times(N/2)$ all-ones matrix, and
$(\delta M)_{ij} = -\|\xi_i^{\epsilon_i} + \epsilon_i \epsilon_j
\xi_j^{\epsilon_j}\|$ off-diagonal, $+\|\xi_i - \xi_j\|$
diagonal. $\pi M^{(0)}$ has rank $2$ with eigenvectors
$u_\pm = N^{-1/2}(\mathbf 1_A,\,\pm\mathbf 1_B)$ and
$\pi M^{(0)} u_\pm = \pm(\pi N/2)\,u_\pm$, so the two
non-trivial leading-order eigenvalues are
$\lambda_{1,2}^{(0)} = \pm \pi N/2$.

The first-order correction from $\sigma\,\delta M$ follows
from the entry statistics. With $\xi_i$ i.i.d.\
$\mathcal N(0, I_{d-1})$, the norms
$\|\xi_i \pm \xi_j\|$ both follow $\sqrt 2\,\chi_{d-1}$, so
the mean entry on either block is
\begin{equation}
\langle\|\xi_i \pm \xi_j\|\rangle
\;=\; \sqrt 2 \cdot \mathbb E[\chi_{d-1}]
\;=\; 2\,\frac{\Gamma(d/2)}{\Gamma((d-1)/2)}
\;=\; \sqrt{4(d-1)/\pi}\,(1 + O(1/d)),
\label{eq:mean_distance}
\end{equation}
exact at $2/\sqrt\pi \approx 1.128$ for $d = 2$ and
$\sqrt\pi \approx 1.772$ for $d = 3$. The mean-piece of
$\delta M$ is rank-$1$ aligned with $u_-$,
\begin{equation}
\delta M^{\rm mean} \;=\;
-\sqrt{4(d-1)/\pi}\,
\bigl(\mathbf 1_A - \mathbf 1_B\bigr)
\bigl(\mathbf 1_A - \mathbf 1_B\bigr)^{\!\top}
+ \text{(cross-term sign convention)},
\label{eq:dM_rank1}
\end{equation}
with eigenvalue $\sigma N\sqrt{4(d-1)/\pi}$ on $u_-$ and
zero on $u_+$ (signs cancel for the symmetric mode).
First-order shifts:
\begin{equation}
\lambda_1 \;\approx\; +\frac{\pi N}{2},
\qquad
\lambda_2 \;\approx\; -\frac{\pi N}{2}
\;+\; \sigma N \sqrt{\frac{4(d-1)}{\pi}}.
\label{eq:lambda12_pred}
\end{equation}
For $N = 400$, $\sigma = 0.05$, $d = 2$:
$\lambda_2 = -605.75$ vs simulation
$-606.19 \pm 0.77$; $d = 3$: $-596.40$ vs
$-593.38 \pm 0.82$ ($20$ seeds).

The bulk eigenvalues and the I-BBS gap complete the
picture. The remaining $N - 2$ eigenvalues come from
fluctuations of $\sigma\,\delta M$ orthogonal to
$u_\pm$. The two within-blob ``Perron-of-an-ERM'' modes
at scale $(N/2)\sigma\sqrt{4(d-1)/\pi}$ are absorbed into
$\lambda_1, \lambda_2$, the bulk modes are
$O(\sigma\sqrt{N(d-1)})$. Numerically $\lambda_3 \approx
12$ for $d = 2$ and $\approx 10$ for $d = 3$ at
$N = 400$, $\sigma = 0.05$. The inter-multiplet gap is
\begin{equation}
\frac{\lambda_2}{\lambda_3} \;\sim\;
\frac{\pi}{\sigma}\,\sqrt{\frac{\pi}{4(d-1)}}\,(1 + O(\sigma)),
\label{eq:gap_prediction}
\end{equation}
$\approx 52$ for $d = 2$ and $61$ for $d = 3$
(simulation: $51.6$, $61.1$; exact agreement). The
log-gap $\approx 1.7$ sits far above the
gap-walk threshold $\tau = 0.25$, so the multiplet
diagnostic robustly returns $\hat h_1 = 1$ (the
between-cluster contrast below the Perron) with
$\hat d = 1$, matching the sparse-parity, input-topology,
and GAN-doublet rows of Table~\ref{tab:ibbs_summary}.

\subsection{Seed and subset robustness of the principal numbers}
\label{app:seed_robustness}

Each seed controls both the training trajectory
(initialisation, mini-batch order, AdamW state) and
the $N$-subset of the test split, so each row of
Table~\ref{tab:robustness} is an independent joint
draw of (training run, evaluation subset).

\begin{table}[htbp]
\centering
\small
\begin{tabular}{l c c c c}
\toprule
& final test acc & $\beta$ plateau (mem) & $\beta$ final &
transition step \\
\midrule
Grokking ($p = 113$) &
$0.994 \pm 0.018$ & $0.479 \pm 0.026$ &
$1.774 \pm 0.107$ & $3779 \pm 482$ \\
Sparse parity ($k = 3$, $d = 30$) &
$1.000 \pm 0.000$ & $0.897 \pm 0.042$ &
$2.323 \pm 0.040$ & $4264 \pm 2140$ \\
\bottomrule
\end{tabular}
\caption{Robustness of the principal OMD quantities at
the standardised $N = 1000$, $10$-seed setup.
Grokking: $p = 113$, $\tau = 0.25$. Sparse parity:
$d = 30$, $k = 3$, $\tau = 0.25$. ``Transition step'' is
the first checkpoint at test accuracy $> 0.99$.
``$\beta$ plateau (mem)'' is the mean $\beta_{\rm raw}$
at $d_{\rm guess} = 2$ on the pre-transition window
($\mathrm{step} > 100$, test accuracy $< 0.6$, and step
$<$ transition step). ``$\beta$ final'' is the
final-checkpoint $\beta_{\rm raw}$ at $d_{\rm guess} =
1$. Cross-seed CV on $\beta$ final: $\sim 6\,\%$
grokking, $\sim 2\,\%$ sparse parity;
transition-step CV $\sim 13\,\%$ and $\sim 50\,\%$
respectively (sparse-parity's wider window reflects
optimiser-discovery dependence on the random orientation
of the relevant input subspace).}
\label{tab:robustness}
\end{table}

\subsection{Additional robustness checks}
\label{app:additional_robustness}

Three short ablations supporting the main-paper
diagnostics.

\label{app:N_sensitivity}\emph{$N$-sensitivity of
$\beta(t)$ on sparse parity.}
The main paper uses $N = 1000$. For this $N$-sensitivity
check we anchor on the earlier $N = 400$ baseline of
\cite{halperin2026FDM}. Rerunning sparse parity with
$N \in \{100, 200, 400, 800\}$ leaves the transition
step $N$-invariant. The pre- and post-transition $\beta$
plateaus converge as $N$ grows (small-$N$ bias pushes
$\beta_{N=100}$ slightly below the $N \geq 200$ value).

\label{app:early_warning}\emph{Early-warning detection of
the grokking transition.}
The $\beta_{\rm train}(t)$ trajectory exits the
memorisation plateau before test accuracy lifts off. On
the canonical grokking run (seed $0$,
Section~\ref{subsec:grokking_results}) the plateau is
$\beta_{\rm plateau} = 0.574 \pm 0.020$ over steps
$500$--$2000$. The $3\sigma$ exit threshold $\beta = 0.632$
is crossed at step $1832$, while test accuracy first
reaches $0.5$ at step $3686$ and $0.7$ at step $4389$,
giving an early-warning lead time of $1854$--$2557$ steps
(roughly $40$--$50\,\%$ of the way to the transition). The
matrix spectrum lifts off the memorisation basin well
before the scalar test loss moves.

\label{app:rotating_subset}\emph{Rotating-subset
stability.}
Recomputing $\beta(t)$ at every grokking checkpoint with
a freshly random $N' = 200$ subset drawn from the stored
$N = 400$ cloud (eight draws per checkpoint) tracks the
fixed-subset trajectory throughout, with rms difference
$0.057$ and intra-draw s.d.\ $0.008$. The spectral signal
is therefore a property of the representation geometry
rather than of the specific evaluation subset, and the
diagnostic can in principle be deployed online with a
rotating evaluation window.

\begin{figure}[t]
\centering
\includegraphics[width=\linewidth]{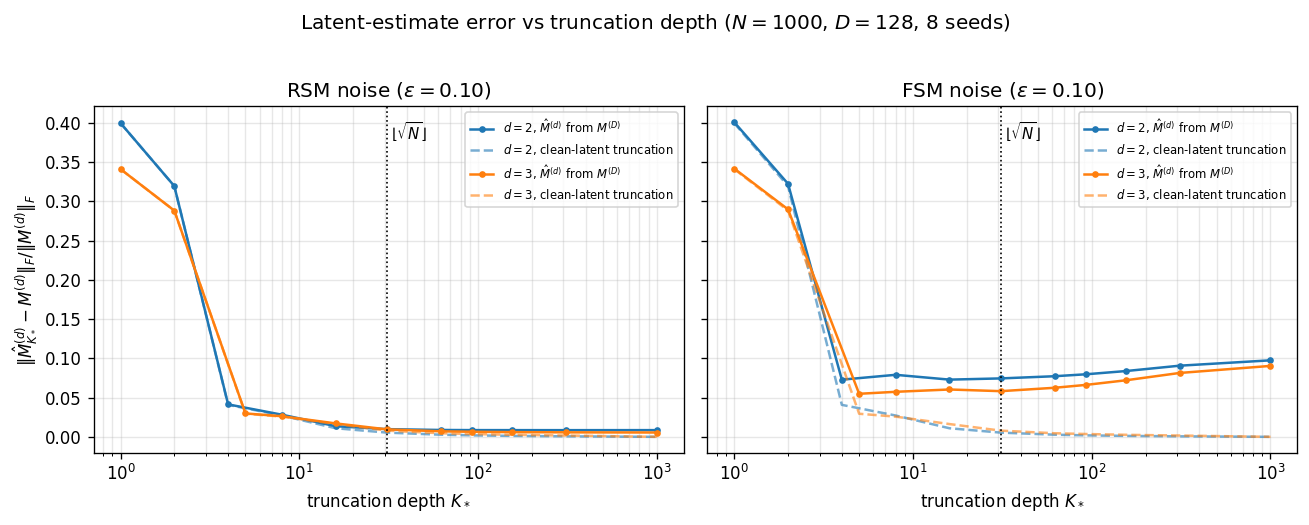}
\caption{Latent-estimate error
$\|\hat M^{(d)}_{K_*} - M^{(d)}\|_F / \|M^{(d)}\|_F$
versus truncation depth $K_*$ on synthetic BBS data
($N = 1000$, $D = 128$, eight seeds, $\epsilon = 0.10$).
Solid: estimate formed from the noisy ambient $M^{(D)}$.
Dashed: irreducible clean-latent truncation. Dotted
vertical line: $K_* = \lfloor\sqrt N\rfloor$. Left, RSM
(low-amplitude noise), where deeper cuts recover the
latent tower. Right, FSM (louder noise), where the noise
bulk enters past $\lfloor\sqrt N\rfloor$.}
\label{fig:reconstruction_depth}
\end{figure}

\label{app:reconstruction_depth}\emph{Reconstruction-depth
sensitivity of $\hat M^{(d)}$.}
The latent estimate $\hat M^{(d)}$ truncates the ambient
spectrum at the latent-versus-noise crossover
$K_* = \lfloor\sqrt N\rfloor$. On synthetic BBS data with
a known clean latent matrix $M^{(d)}$ ($N = 1000$,
$D = 128$, eight seeds, $S^2$ and $S^3$), the latent
weight discarded beyond $\lfloor\sqrt N\rfloor$ is
$0.5$--$0.8\%$ of $\|M^{(d)}\|_F$, and beyond the coarse
$1 + \hat h_1$ band it is $3$--$4\%$.
Figure~\ref{fig:reconstruction_depth} traces the
estimation error
$\|\hat M^{(d)}_{K_*} - M^{(d)}\|_F / \|M^{(d)}\|_F$
against the cut $K_*$. For low-amplitude noise (RSM, here
$\lesssim 1\%$) the error falls from about $4\%$ at the
band to about $1\%$ at $\lfloor\sqrt N\rfloor$ as the
higher latent multiplets are recovered, then flattens.
For louder noise (FSM, here near $10\%$) the band is
already close to optimal, and pushing the cut past
$\lfloor\sqrt N\rfloor$ raises the error as the Wigner
bulk enters the reconstruction. In every case the cut at
$\lfloor\sqrt N\rfloor$ is at or within a sub-percent of
the error floor, a robust noise-agnostic choice. The
residual-RMT verdict is identical for the band and the
$\lfloor\sqrt N\rfloor$ reconstructions across both noise
models and both dimensions ($48$ of $48$ synthetic
cases), so the per-experiment verdicts do not depend on
the reconstruction depth.

\subsection{Window and threshold robustness sweep
$(\tau, K_{\max}, N)$ for Algorithm 1}
\label{app:robustness-sweep}

We sweep
$\tau \in \{0.10, 0.15, 0.25, 0.35, 0.50\}$,
$K_{\max} \in \{30, 50\}$, $N \in \{400, 1000\}$
(the $N = 400$ cell drawn by random subsampling from
the stored $N = 1000$ matrix, fixed across seeds) on
four representatives, $10$ seeds per cell. Task switch
gives $\hat h_1 = 2$ in $90$--$100\%$ across every
cell. Input topology gives $\hat h_1 = 1$ and
$\hat d_\beta = 4$ in $10/10$ across the entire grid.
Regression is similarly $10/10$ stable except at
$\tau = 0.50$, where the gap walk fails to close in any
seed, the conservative ``not identified'' outcome.
Sparse parity bifurcates at $\tau \approx 0.20$: for
$\tau \le 0.15$ the gap walk closes early on a singlet
($\hat h_1 = 1$ in $30$--$70\%$, missing the
$8$-vertex band), while $\tau \ge 0.25$ traverses to
the full $\hat h_1 = 7$ band ($40$--$90\%$), matching
the body-text reading. The production setting
$(\tau, K_{\max}, N) = (0.25, 50, 1000)$ therefore
reflects properties of the spectrum rather than
artefacts of the window, with the operational caveat
that atomic-cluster regimes require $\tau \ge 0.25$ for
the multiplet to register.

\begin{figure}[!htbp]
  \centering
  \includegraphics[width=0.7\textwidth]{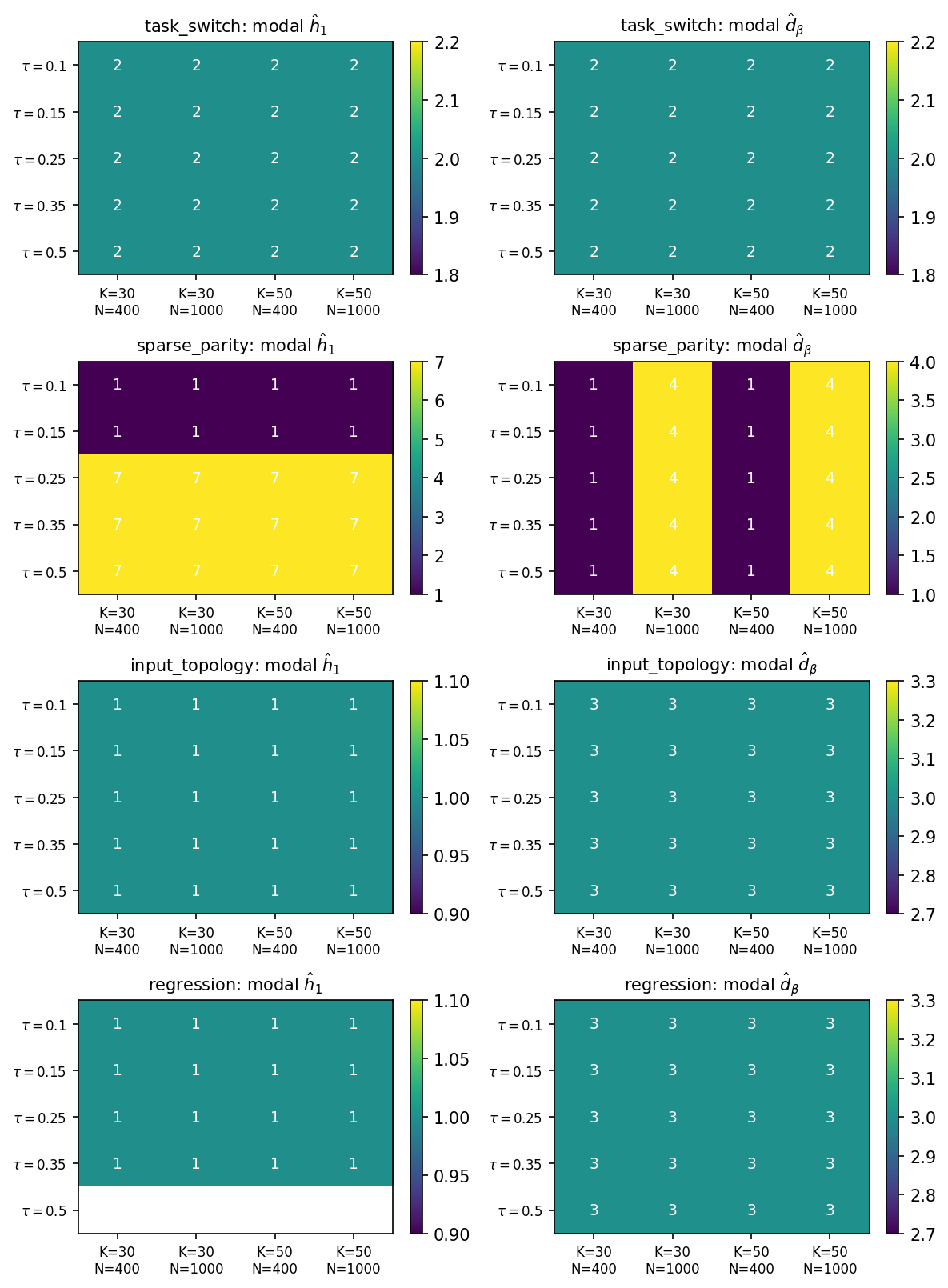}
  \caption{Algorithm~1 robustness sweep over
  $(\tau, K_{\max}, N)$ for four representatives, one
  per row, with two heatmaps each: modal $\hat h_1$
  (left) and modal $\hat d_\beta$ (right). In every
  heatmap $\tau$ runs down the vertical and the
  $(K_{\max}, N)$ pairs along the horizontal. The
  integer printed in each cell is the mode of that
  estimate over the $10$ seeds of the cell (the
  most frequent value), and the cell colour encodes
  the same integer. A blank cell marks the conservative
  ``not identified'' outcome, where the gap walk closes
  in no seed and no mode exists (regression at
  $\tau = 0.50$).}
  \label{fig:robustness-sweep}
\end{figure}

\subsection{Reversibility and attractor probes via
$\delta_G$}
\label{app:delta-G-probes}

\begin{figure}[htbp]
  \centering
  \includegraphics[width=0.85\textwidth]{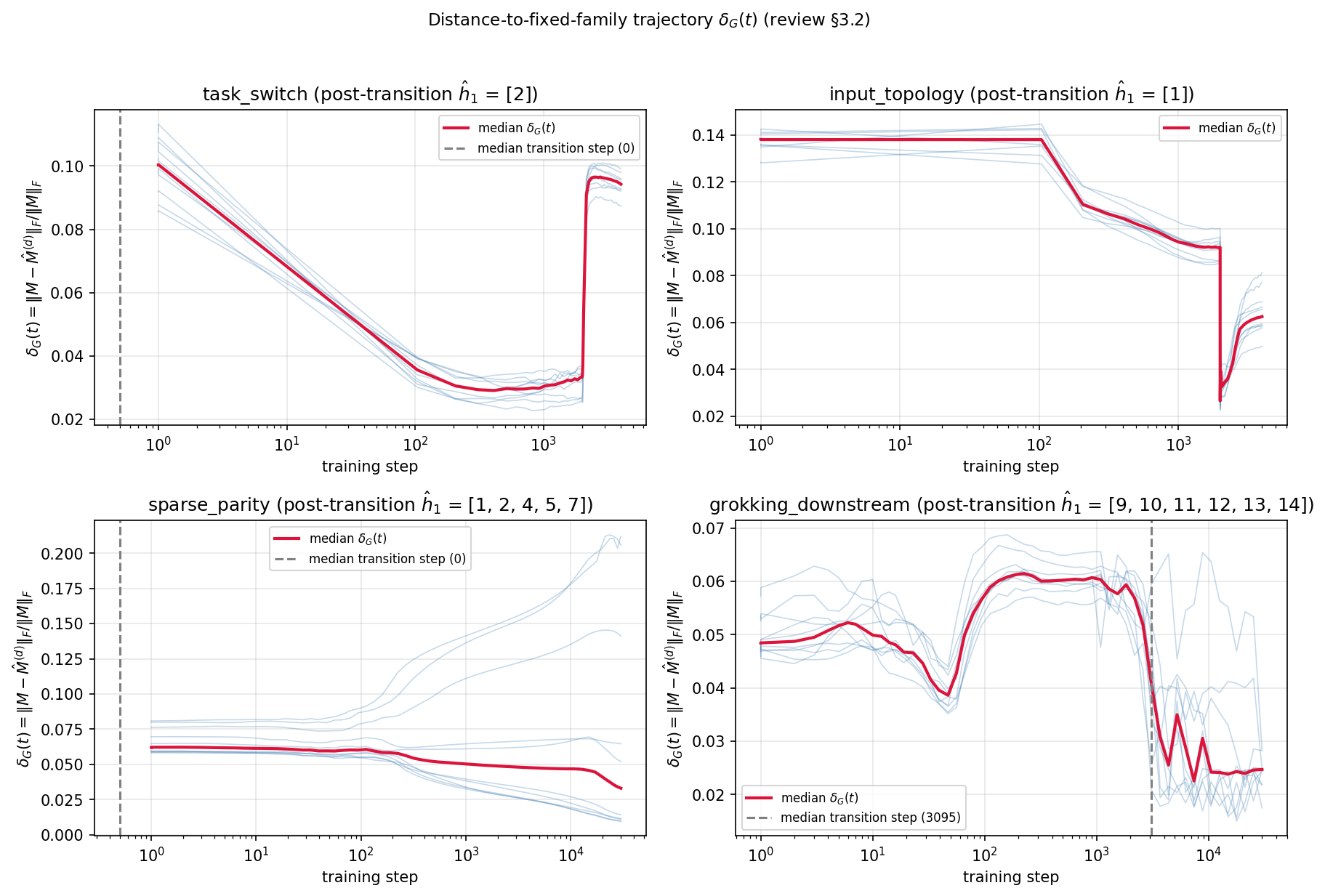}
  \caption{$\delta_G(t)$ per seed (thin blue) and median
  (thick red) across the four Group~B experiments.
  Dashed grey: median transition step (first step at
  which test accuracy $\ge 0.5$).}
  \label{fig:delta-G-trajectory}
\end{figure}

A direct probe of $\mathcal F_G$ as an attractor
(Appendix~\ref{app:fixed-points}) is the distance
trajectory $\delta_G(t) = \|M(t) -
\mathrm{rec}_{1+\hat h_1^{\rm post}}(M(t))\|_F /
\|M(t)\|_F$, the fraction of $\|M(t)\|_F$ outside the
rank-$(1+\hat h_1^{\rm post})$ image, with
$\hat h_1^{\rm post}$ the final-checkpoint multiplet
for the seed. Attracting $\mathcal F_G$ predicts
$\delta_G$ large pre-transition and small
post-transition. Table~\ref{tab:delta-G} reports
init/final values across the four Group~B experiments.
The trajectories are in
Figure~\ref{fig:delta-G-trajectory}. Grokking
(downstream) and input topology show the predicted
$\sim 2\times$ contraction with unanimous
$\hat h_1^{\rm post}$. Task switch is flat at $\sim
0.10$ throughout (a doublet is present already at
init in any rank-decaying random matrix, so the OMD
transition refines the doublet content, not its rank.
$\delta_G$ is not the right metric here, $\hat h_1$
identity is). Sparse parity averages flat because
$\hat h_1^{\rm post}$ is heterogeneous across seeds
($\{7{:}5, 1{:}2, 2{:}1, 4{:}1, 5{:}1\}$), though the five
modal-$\hat h_1=7$ seeds do contract (median ratio
$\approx 1.6$).

\begin{table}[htbp]
\centering
\footnotesize
\begin{tabular}{lcccc}
\toprule
Experiment & $\hat h_1^{\rm post}$ &
$\delta_G^{\rm init}$ & $\delta_G^{\rm final}$ &
ratio \\
\midrule
Task switch & $\{2{:}10\}$ &
$0.100 \pm 0.008$ & $0.094 \pm 0.003$ & $1.06$ \\
Input topology & $\{1{:}10\}$ &
$0.137 \pm 0.004$ & $0.064 \pm 0.009$ & $2.13$ \\
Sparse parity & $\{7{:}5, 1{:}2, 2{:}1, 4{:}1, 5{:}1\}$ &
$0.067 \pm 0.009$ & $0.073 \pm 0.078$ & $0.91$ \\
Grokking (downstream) &
$\{12{:}5, 9{:}1, 10{:}1, 11{:}1, 13{:}1, 14{:}1\}$ &
$0.050 \pm 0.003$ & $0.024 \pm 0.003$ & $2.06$ \\
\bottomrule
\end{tabular}
\caption{$\delta_G$ at first and last stored ckpt
across the four Group~B experiments ($10$ seeds,
$N = 1000$, $\tau = 0.25$). Grokking and input
topology show the expected $\sim 2\times$ contraction;
task switch and sparse parity flagged in text.}
\label{tab:delta-G}
\end{table}

After the transition $\delta_G(t)$ should stay small
for the remainder of training. We measure the relative
late-tail drift $(\langle\delta_G\rangle_{\rm last\,10\%} -
\langle\delta_G\rangle_{\rm last\,25\%}) /
\langle\delta_G\rangle_{\rm last\,25\%}$ across $10$
seeds. The four Group~B experiments give: grokking
(downstream) $-9.6\% \pm 10.2\%$ (late-tail $\delta_G
\approx 0.027$, mild contraction within optimisation
noise); input topology $+2.4\% \pm 0.9\%$
($\delta_G \approx 0.064$, tightly stationary); task
switch $-0.7\% \pm 0.2\%$ ($\delta_G \approx 0.095$,
essentially exact stationarity); sparse parity
$-11.1\% \pm 12.7\%$ ($\delta_G \approx 0.075$,
heterogeneous between the stationary modal-$\hat h_1=7$
seeds and the drifting off-modal seeds, the same split as
above). Three of four experiments show $<10\%$ late-tail
drift, consistent with $\mathcal F_G$ being a stationary
fixed family within the $30{,}000$-step training horizon.

A sharper test is reversibility. We modify the task-switch
and input-topology drivers to insert a second switch back
to phase A at step $4000$ (phases A: $0$--$2000$, B:
$2000$--$4000$, A$_2$: $4000$--$6000$), $5$ seeds each at
$N = 1000$, and track $\delta_G^A(t)$ and $\delta_G^B(t)$,
the same metric referred to the multiplets $\hat h_1^A$,
$\hat h_1^B$ of the phase-A and phase-B segments.

For task switch, $\hat h_1^A = 1$ and $\hat h_1^B = 2$
in $5/5$ seeds, matching the body-text one-shot reading.
Median $\delta_G^A$ goes $0.047$ (end-A) $\to 0.283$
(end-B) $\to 0.177$ (end-A$_2$), a $6\times$ rise during
phase B and a $1.6\times$ fall during A$_2$, recovering
$\sim 63\%$ of the contraction toward $\mathcal F_A$.
Median $\delta_G^B$ goes $0.034 \to 0.093 \to 0.079$,
so the doublet representation is not fully unwound in
the time available. The partial reversibility is
consistent with $\mathcal F_A$ as an attractor, the residual
offset reflecting optimiser history and the finite
phase-A$_2$ length.

For input topology, $\hat h_1 = 1$ in both phases, so
$\delta_G^A \equiv \delta_G^B$ and the reversibility
metric does not discriminate. The single $\delta_G$ trace
is $0.092 \to 0.066 \to 0.144$, the post-A$_2$ expansion
consistent with the network's single-cluster representation
absorbing two-cluster input variance.

\begin{figure}[htbp]
  \centering
  \includegraphics[width=\textwidth]{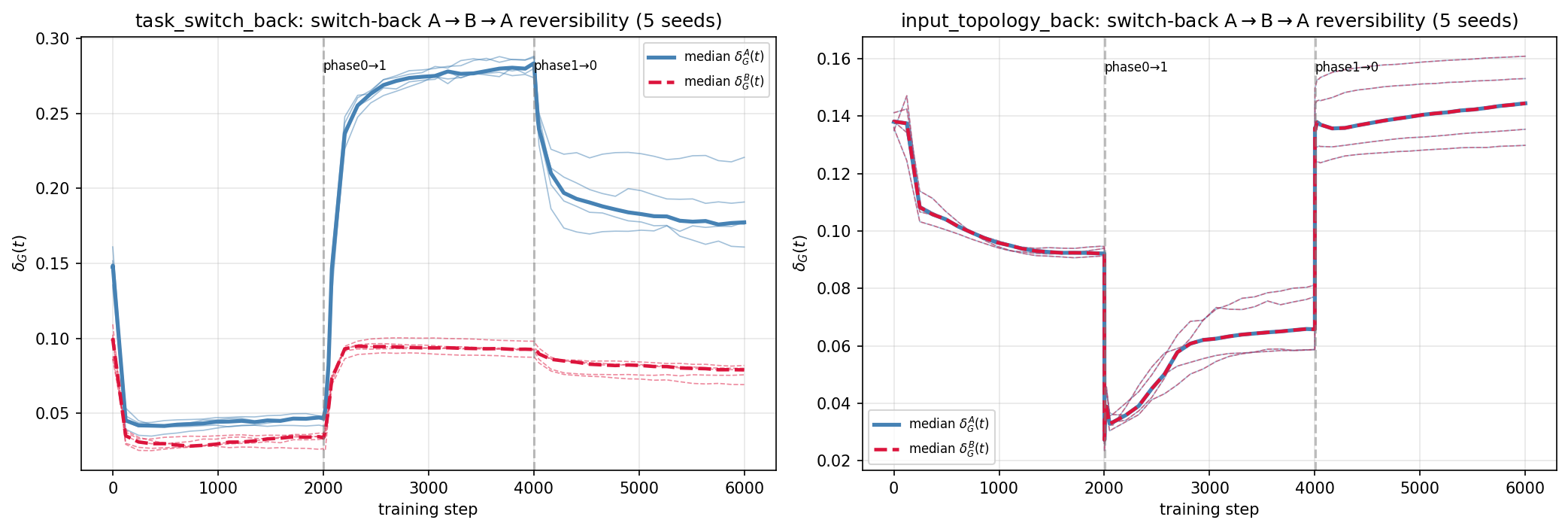}
  \caption{Switch-back A$\to$B$\to$A on task switch
  (left) and input topology (right); $5$ seeds, $N = 1000$.
  Blue/red: median $\delta_G^A(t)$/$\delta_G^B(t)$, thin
  lines per-seed. Dashed grey verticals: phase boundaries
  (steps $2000$, $4000$).}
  \label{fig:switchback-T5p2}
\end{figure}

The sharpest attractor test perturbs and returns. At step
$20{,}000$ (well post-transition, $100\%$ accuracy) we add
Gaussian noise of std $\sigma_W$ to every model weight,
reset the optimiser, and continue to step $30{,}000$,
sweeping $\sigma_W \in \{0.05, 0.10, 0.50, 1.00\}$, $3$
seeds each at $N = 1000$, with $\hat h_1^{\rm pre}$ from
the immediately pre-perturb ckpt and $\delta_G^{\rm pre}$
measured as above.
Table~\ref{tab:perturb-T5p3-sigma-sweep} reports the
three phase points (pre, immediately post-perturb, end
of $10{,}000$-step recovery) and the recovery fraction
$(\delta_G^{\rm post} - \delta_G^{\rm final}) /
(\delta_G^{\rm post} - \delta_G^{\rm pre})$.

\begin{table}[htbp]
\centering
\small
\begin{tabular}{ccccccc}
\toprule
$\sigma_W$ &
$\delta_G^{\rm pre}$ &
$\delta_G^{\rm post}$ &
$\delta_G^{\rm final}$ &
recovery &
acc post &
acc final \\
\midrule
$0.05$ & $0.015$ & $0.022$ & $0.030$ & $-129\%$ & $1.00$ & $1.00$ \\
$0.10$ & $0.015$ & $0.040$ & $0.031$ & $+35\%$  & $0.88$ & $1.00$ \\
$0.50$ & $0.015$ & $0.059$ & $0.059$ & $+1\%$   & $0.50$ & $0.54$ \\
$1.00$ & $0.015$ & $0.059$ & $0.058$ & $+1\%$   & $0.49$ & $0.52$ \\
\bottomrule
\end{tabular}
\caption{Perturb-and-return median values across $3$
sparse-parity seeds at each $\sigma_W$. Cleanest
single demonstration of $\mathcal F_G$ as an attractor:
$\sigma_W = 0.10$ is the basin-boundary cell.}
\label{tab:perturb-T5p3-sigma-sweep}
\end{table}

Three regimes appear. At $\sigma_W = 0.05$ the
perturbation is in-basin: test accuracy unaffected,
$\delta_G^{\rm pre}$ jump within the late-tail drift
envelope above. At
$\sigma_W = 0.10$ the perturbation reaches the basin
boundary: test accuracy transiently drops to $0.88$ and
recovers fully, $\delta_G^{\rm pre}$ recovers $+35\%$ of
the perturb jump within the $10{,}000$-step window. This
is the cleanest demonstration in the paper of
$\mathcal F_G$ as a true attractor with finite basin:
the network leaves and returns both functionally and
geometrically. At $\sigma_W \in \{0.50, 1.00\}$ the
perturbation kicks the network out of the
parity-solution manifold: test accuracy collapses to
random chance and does not recover, $\delta_G^{\rm pre}$
saturates at $\sim 0.06$ (the matrix signature of
a random-feature state, not a different sparse-parity
fixed family). The basin width is therefore on the
order of $\sigma_W \sim 0.1$ for this experiment. The
off-modal $\hat h_1^{\rm pre} = 1$ seed above starts at
$\delta_G^{\rm pre} = 0.20$ and \emph{decreases} on
perturbation, the same sign issue carried over.

\begin{figure}[htbp]
  \centering
  \includegraphics[width=0.85\textwidth]{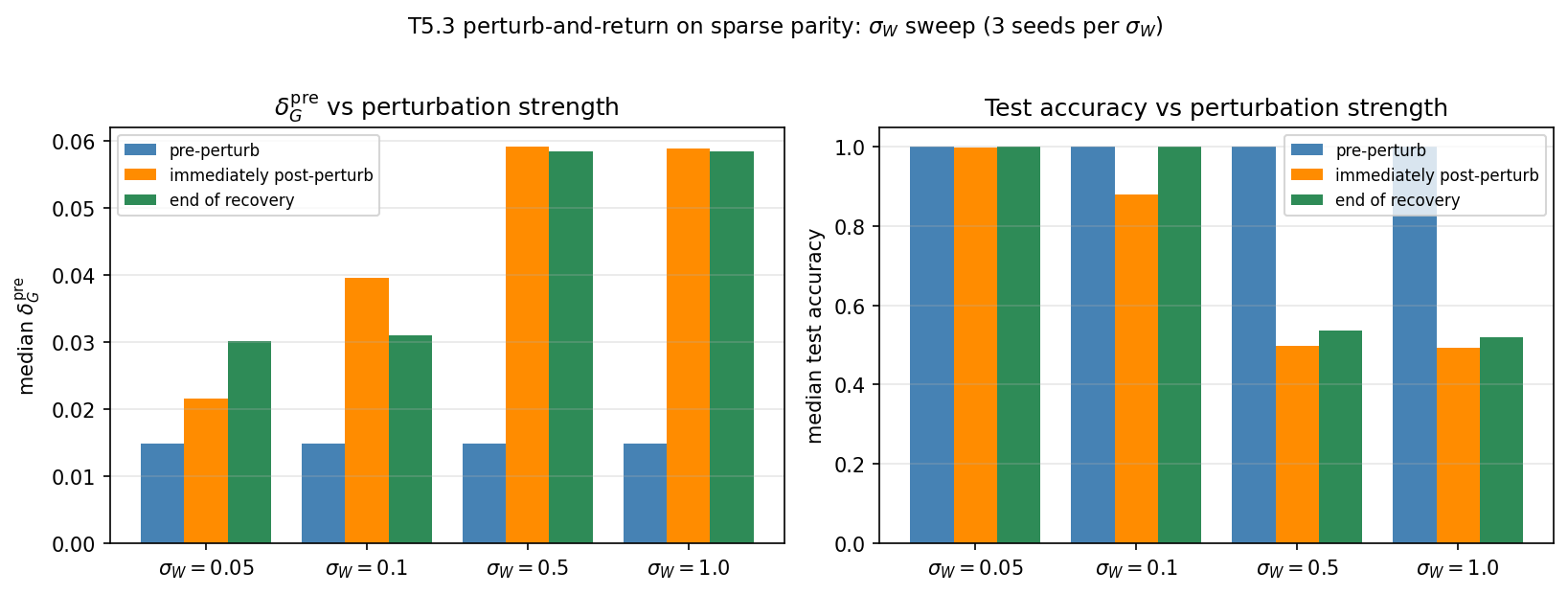}
  \caption{$\sigma_W$ sweep on sparse parity ($3$ seeds
  per cell). Left: median $\delta_G^{\rm pre}$ at pre,
  immediately post-perturb, end of recovery. Right:
  median test accuracy at the same three points.
  $\sigma_W = 0.10$ is the basin-boundary cell where
  both observables briefly leave and return; larger
  $\sigma_W$ destroys the parity solution.}
  \label{fig:perturb-T5p3-sigma-sweep}
\end{figure}

\begin{figure}[htbp]
  \centering
  \includegraphics[width=0.9\textwidth]{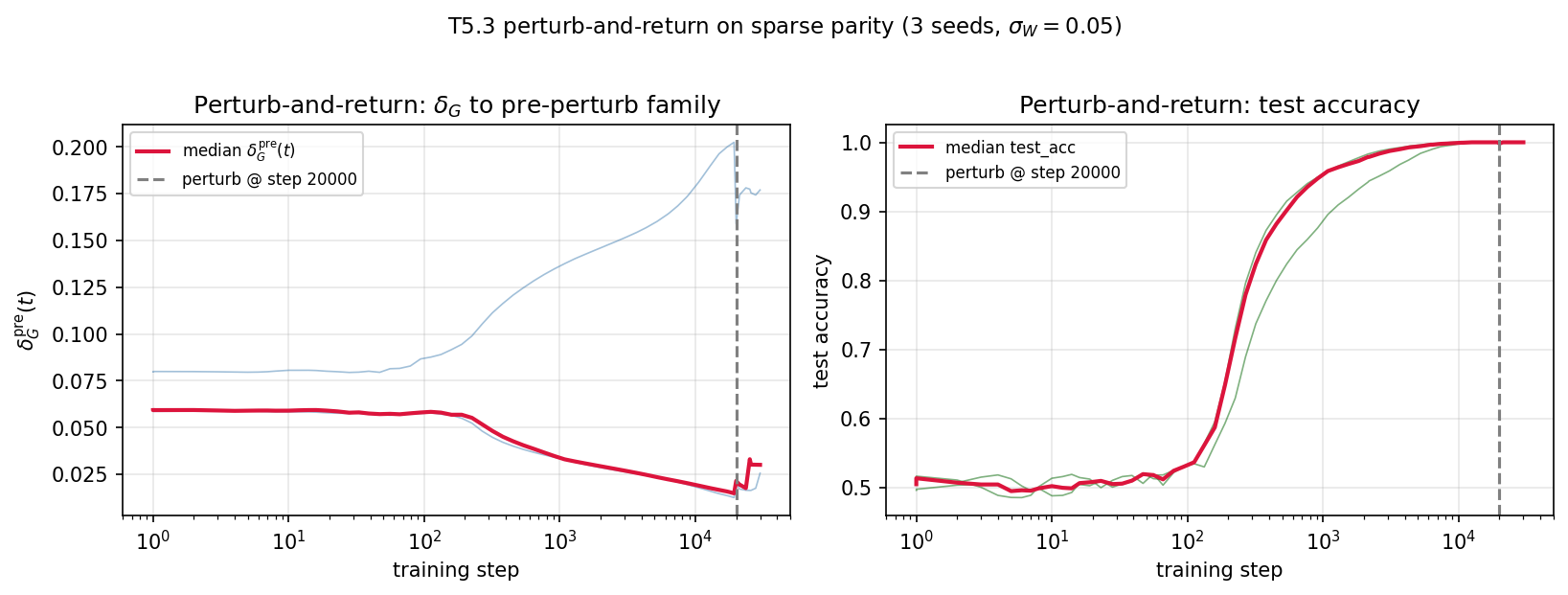}
  \caption{In-basin trajectories at $\sigma_W = 0.05$:
  $\delta_G^{\rm pre}(t)$ (left) and test accuracy
  (right) per seed and median. Dashed grey: perturbation
  step.}
  \label{fig:perturb-T5p3}
\end{figure}

\subsection{Token-deduplicated upstream reading on the
modular-arithmetic transformer}
\label{app:token-dedup}

\begin{figure}[htbp]
  \centering
  \includegraphics[width=0.9\textwidth]{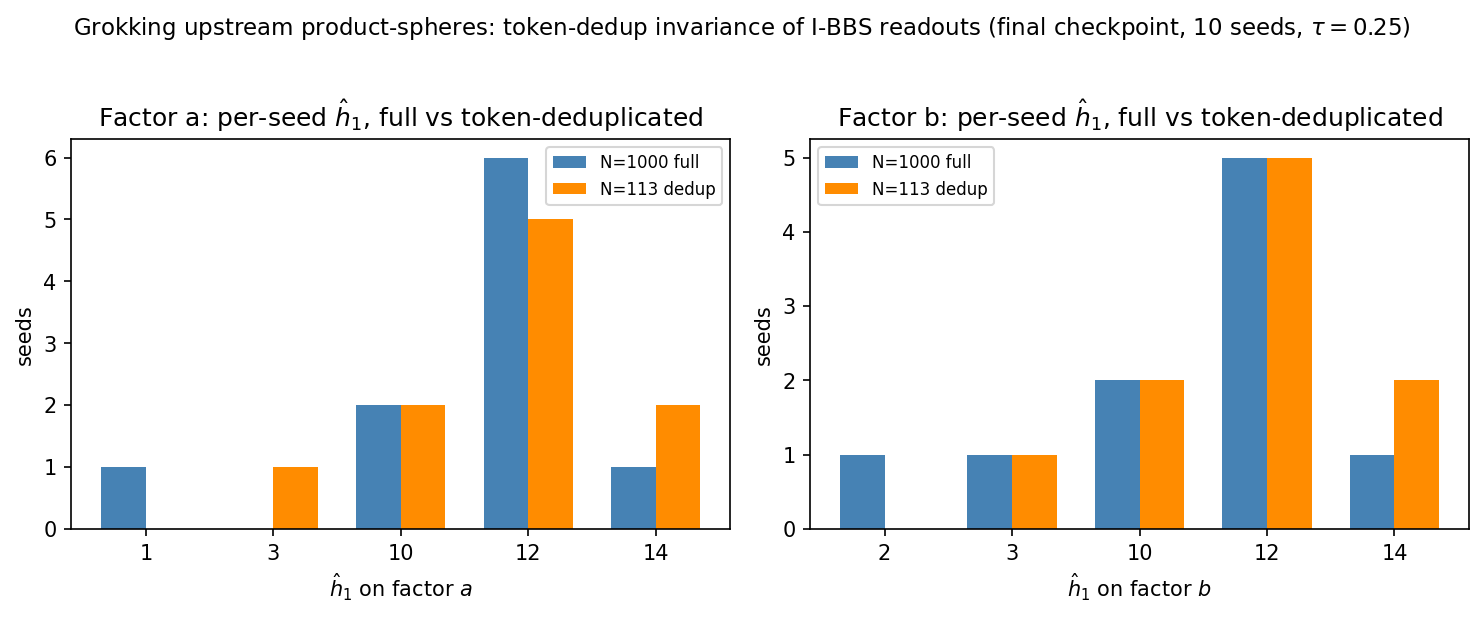}
  \caption{Per-seed $\hat h_1$ on upstream factors $a$
  and $b$, $\tau = 0.25$. Blue: full $N = 1000$;
  orange: dedup $N = 113$. Dedup removes the
  multiplicity smearing at the
  $\mathbb Z_p$-equivariant band edge.}
  \label{fig:token-dedup}
\end{figure}

Because the transformer has a single embedding table,
per-position rows $H[:, \mathrm{pos}, :]$ for the same
token id are identical, so the random $N = 1000$
sample is effectively $\le p = 113$ distinct tokens
repeated with the multiplicity of the random
$(a, b)$ draw. Deduplicating to the unique rows
($N = 113$ each) gives the sampling-multiplicity-free
counterpart and the $p^2$-grid-equivalent for the
per-factor I-BBS readout. Table~\ref{tab:token-dedup}
and Figure~\ref{fig:token-dedup} compare diagnostics at
the final checkpoint, $10$ seeds, $\tau = 0.25$:
$\hat d_\beta$ and the RSM verdict are perfectly
invariant, the $\hat h_1 = 12$ modal value is preserved
on both factors, and two outlier seeds (factor $a$
seed $04$: $\hat h_1 = 1 \to 14$; factor $b$ seed $09$:
$2 \to 14$) move from a degenerate to the expected
high-multiplet band, consistent with multiplicity
weighting smearing the band edge. The body-text
$T^2 = S^1(a) \times S^1(b)$ verdict is therefore a
property of the per-factor token-embedding spectrum.

\begin{table}[htbp]
\centering
\small
\begin{tabular}{lll}
\toprule
Quantity & Full ($N = 1000$) & Dedup ($N = 113$) \\
\midrule
factor $a$ $\hat h_1$ dist &
$\{12{:}6, 10{:}2, 1{:}1, 14{:}1\}$ &
$\{12{:}5, 10{:}2, 14{:}2, 3{:}1\}$ \\
factor $b$ $\hat h_1$ dist &
$\{12{:}5, 10{:}2, 14{:}1, 3{:}1, 2{:}1\}$ &
$\{12{:}5, 10{:}2, 14{:}2, 3{:}1\}$ \\
$\hat d_\beta$ both factors & $\{4{:}10\}$ & $\{4{:}10\}$ \\
verdict both factors &
RSM-like ($10/10$) &
RSM-like ($10/10$) \\
\bottomrule
\end{tabular}
\caption{Per-factor I-BBS readout, final ckpt, $10$
seeds, $\tau = 0.25$. $\hat d_\beta$ and verdict are
invariant under deduplication; the $\hat h_1$ modal is
preserved.}
\label{tab:token-dedup}
\end{table}

%==============================================================================


\begin{thebibliography}{99}
\setlength{\itemsep}{0pt}
\setlength{\parskip}{0pt}

\bibitem{mezard1999}
M.~M\'ezard, G.~Parisi, and A.~Zee.
\emph{Spectra of Euclidean random matrices.}
\textit{Nuclear Physics B} \textbf{559}, 689--701 (1999).

\bibitem{goetschy2013}
A.~Goetschy and S.~E.~Skipetrov.
\emph{Euclidean random matrices and their applications in physics.}
arXiv:1303.2880, 2013.

\bibitem{bogomolny2003}
E.~Bogomolny, O.~Bohigas, and C.~Schmit.
\emph{Spectral properties of distance matrices.}
\textit{Journal of Physics A: Mathematical and General}
\textbf{36}, 3595--3616 (2003). arXiv:nlin/0301044.

\bibitem{bogomolny2007}
E.~Bogomolny, O.~Bohigas, and C.~Schmit.
\emph{Distance matrices and isometric embeddings.}
arXiv:0710.2063, 2007.

\bibitem{halperin2026IBBS}
I.~Halperin.
\emph{I-BBS: Inference of Latent Sub-Manifolds in Representation Spaces
Using Random Distance Matrices.} 2026.

\bibitem{halperin2026FDM}
I.~Halperin.
\emph{Frustrated Dynamics of Distance Matrices.}
arXiv:2605.05376, 2026.

\bibitem{power2022grokking}
A.~Power, Y.~Burda, H.~Edwards, I.~Babuschkin, V.~Misra.
\emph{Grokking: generalization beyond overfitting on small
algorithmic datasets.}
arXiv:2201.02177, 2022.

\bibitem{halperin2026grokking}
I.~Halperin.
\emph{Grokking as Bagel Formation in Activation Space:
Spectral Evidence for a Phase Transition.}
2026.

\bibitem{mezard1987beyond}
M.~M\'ezard, G.~Parisi, and M.~A.~Virasoro.
\emph{Spin Glasses and Beyond: An Introduction to the
Replica Method and Its Applications.}
World Scientific, Singapore (1987).

\bibitem{kriegeskorte2008rsa}
N.~Kriegeskorte, M.~Mur, P.~A.~Bandettini.
\emph{Representational similarity analysis: connecting the
branches of systems neuroscience.}
\textit{Frontiers in Systems Neuroscience} \textbf{2}, 4
(2008).

\bibitem{kornblith2019cka}
S.~Kornblith, M.~Norouzi, H.~Lee, G.~Hinton.
\emph{Similarity of neural network representations revisited.}
In \textit{ICML} (2019).

\bibitem{papyan2020neural}
V.~Papyan, X.~Y.~Han, D.~L.~Donoho.
\emph{Prevalence of neural collapse during the terminal phase of
deep learning training.}
\textit{Proceedings of the National Academy of Sciences}
\textbf{117}(40), 24652--24663 (2020).

\bibitem{fang2021layerpeeled}
C.~Fang, H.~He, Q.~Long, W.~J.~Su.
\emph{Exploring deep neural networks via layer-peeled model:
minority collapse in imbalanced training.}
\textit{Proceedings of the National Academy of Sciences}
\textbf{118}, e2103091118 (2021).

\bibitem{zhu2021geometric}
Z.~Zhu \emph{et al.}.
\emph{A geometric analysis of neural collapse with unconstrained
features.}
In \textit{NeurIPS} (2021).

\bibitem{mixon2020neural}
D.~G.~Mixon, H.~Parshall, J.~Pi.
\emph{Neural collapse with unconstrained features.}
arXiv:2011.11619, 2020.

\bibitem{he2023law}
H.~He, W.~J.~Su.
\emph{A law of data separation in deep learning.}
\textit{Proceedings of the National Academy of Sciences}
\textbf{120}, e2221704120 (2023).

\bibitem{rangamani2023intermediate}
A.~Rangamani, M.~Lindegaard, T.~Galanti, T.~A.~Poggio.
\emph{Feature learning in deep classifiers through
intermediate neural collapse.}
In \textit{ICML} (2023).

\bibitem{facco2017twonn}
E.~Facco, M.~d'Errico, A.~Rodriguez, A.~Laio.
\emph{Estimating the intrinsic dimension of datasets by a minimal
neighborhood information.}
\textit{Scientific Reports} \textbf{7}, 12140 (2017).

\bibitem{ansuini2019intrinsic}
A.~Ansuini, A.~Laio, J.~H.~Macke, D.~Zoccolan.
\emph{Intrinsic dimension of data representations in deep
neural networks.}
In \textit{NeurIPS} (2019).

\bibitem{li2018measuring}
C.~Li, H.~Farkhoor, R.~Liu, J.~Yosinski.
\emph{Measuring the intrinsic dimension of objective landscapes.}
In \textit{ICLR} (2018).

\bibitem{thilak2022slingshot}
V.~Thilak, E.~Littwin, S.~Zhai, O.~Saremi, R.~Paiss, J.~Susskind.
\emph{The slingshot mechanism: an empirical study of adaptive
optimizers and the grokking phenomenon.}
arXiv:2206.04817, 2022.

\bibitem{nanda2023progress}
N.~Nanda, L.~Chan, T.~Lieberum, J.~Smith, J.~Steinhardt.
\emph{Progress measures for grokking via mechanistic
interpretability.}
arXiv:2301.05217, 2023.

\bibitem{bronstein2021geometric}
M.~M.~Bronstein, J.~Bruna, T.~Cohen, P.~Veli\v{c}kovi\'c.
\emph{Geometric deep learning: grids, groups, graphs,
geodesics, and gauges.}
arXiv:2104.13478, 2021.

\bibitem{cohen2016group}
T.~Cohen, M.~Welling.
\emph{Group equivariant convolutional networks.}
In \textit{ICML} (2016).

\bibitem{esteves2018learning}
C.~Esteves, C.~Allen-Blanchette, A.~Makadia, K.~Daniilidis.
\emph{Learning SO(3) equivariant representations with
spherical CNNs.}
In \textit{ECCV} (2018).

\bibitem{halperin2026frustrated}
I.~Halperin.
\emph{Order Out of Noise and Disorder: Fate of the Frustrated
Manifold.}
arXiv:2601.18653, 2026.

\bibitem{halperin2026fields}
I.~Halperin.
\emph{Frustrated Fields: Statistical Field Theory for Frustrated
Brownian Particles on 2D Manifolds.}
arXiv:2605.05366, 2026.

\bibitem{DavisKahan1970}
C.~Davis and W.~M.~Kahan.
\emph{The rotation of eigenvectors by a perturbation. III.}
\textit{SIAM Journal on Numerical Analysis} \textbf{7}(1), 1--46 (1970).

\bibitem{gray2006circulant}
R.~M.~Gray.
\emph{Toeplitz and Circulant Matrices: A Review.}
\textit{Foundations and Trends in Communications and
Information Theory} \textbf{2}(3), 155--239 (2006).

\bibitem{sustik2007etf}
M.~A.~Sustik, J.~A.~Tropp, I.~S.~Dhillon, R.~W.~Heath.
\emph{On the existence of equiangular tight frames.}
\textit{Linear Algebra and its Applications}
\textbf{426}(2--3), 619--635 (2007).

\bibitem{kumar2024grokking}
T.~Kumar, B.~Bordelon, S.~J.~Gershman, C.~Pehlevan.
\emph{Grokking as the transition from lazy to rich training
dynamics.}
arXiv:2310.06110, 2024.

\bibitem{landau1980statistical}
L.~D.~Landau and E.~M.~Lifshitz.
\emph{Statistical Physics, Part 1.}
Course of Theoretical Physics, Vol.~5, 3rd ed.,
Pergamon Press, Oxford, 1980.

\bibitem{gronwall1919}
T.~H.~Gr\"onwall.
\emph{Note on the derivatives with respect to a parameter of
the solutions of a system of differential equations.}
\textit{Ann. of Math.} \textbf{20}, 292--296 (1919).

\bibitem{alstott2014powerlaw}
J.~Alstott, E.~Bullmore, D.~Plenz.
\emph{powerlaw: A Python package for analysis of heavy-tailed
distributions.}
\textit{PLoS ONE} \textbf{9}, e85777 (2014).

\bibitem{martin2021weightwatcher}
C.~H.~Martin, T.~S.~Peng, M.~W.~Mahoney.
\emph{Predicting trends in the quality of state-of-the-art
neural networks without access to training or testing data.}
\textit{Nature Communications} \textbf{12}, 4122 (2021).
WeightWatcher package: \texttt{https://github.com/CalculatedContent/WeightWatcher}.

\end{thebibliography}
\end{document}